\documentclass[journal]{IEEEtran}
\newtheorem{thm}{Theorem}[section]
\newtheorem{cor}[thm]{Corollary}
\newtheorem{lem}[thm]{Lemma}
\newtheorem{asmp}[thm]{Assumption}
\newtheorem{prop}[thm]{Proposition}
\newtheorem{problem}{Problem}
\newtheorem{defn}[thm]{Definition}
\newtheorem{rem}[thm]{Remark}

\newcommand{\ltl}{LTL$_{-\bigcirc}$}

\makeatletter
\renewcommand*{\@opargbegintheorem}[3]{\trivlist
  \item[\hskip \labelsep{\it\quad  #1\ #2:}] {\it(#3)}\ }
\makeatother

\IEEEoverridecommandlockouts
\usepackage{graphicx} 
\graphicspath{{}}
\usepackage{bbm}
\usepackage{hyperref}
\usepackage{epsfig} 
\usepackage{amsmath}
\usepackage{amssymb}
\usepackage{epstopdf}
\usepackage{cite}
\usepackage[noend,ruled,linesnumbered,resetcount]{algorithm2e}
\SetKwComment{Comment}{$\triangleright$\ } {}
\usepackage{multirow}
\usepackage{rotating}
\usepackage{subfigure}
\usepackage{color}
\usepackage{mysymbol}
\usepackage{url}
\usepackage{ltlfonts}
\usepackage{tabularx,ragged2e,booktabs}
\usepackage[normalem]{ulem}
\usepackage[utf8x]{inputenc}
\usepackage{multirow}
\newcommand{\Qb}{\mathcal{Q}_{B}}
\newcommand{\qb}{q_{{B}}}
\newcommand{\vect}[1]{\boldsymbol{\mathbf{#1}}}
\newcommand{\wf}{\mathcal{W}_{\text{free}}}

\newcommand{\x}[1]{\vect{x}^{\text{#1}}}
\newcommand{\q}[2]{{q}_{#1}^{\text{#2}}}

\newcommand{\RNum}[1]{\uppercase\expandafter{\romannumeral #1\relax}}

\newcommand*\diff{\mathop{}\!\mathrm{d}}

\newcommand{\xx}{\overline{\bbx_{i}\bbx_{i}'}}

\usepackage[hang,flushmargin]{footmisc}
\makeatletter
\newcommand{\algorithmfootnote}[2][\footnotesize]{%
  \let\old@algocf@finish\@algocf@finish
  \def\@algocf@finish{\old@algocf@finish
    \leavevmode\rlap{\begin{minipage}{\linewidth}
    #1#2
    \end{minipage}}%
  }%
}
\makeatother

\begin{document}

\title{An Abstraction-Free Method for Multi-Robot Temporal Logic Optimal Control Synthesis}

\author{Xusheng Luo, Yiannis~Kantaros,~\IEEEmembership{Member,~IEEE,} and \\Michael~M.~Zavlanos,~\IEEEmembership{Senior Member,~IEEE} \thanks{Xusheng Luo, Yiannis Kantaros and Michael M. Zavlanos are with the Department of Mechanical Engineering and Materials Science, Duke University, Durham, NC 27708, USA. $\left\{\text{xusheng.luo, yiannis.kantaros, michael.zavlanos}\right\}$@duke.edu. This work is supported in part by the ONR under agreement \#N00014-18-1-2374 and
the AFOSR under award \#FA9550-19-1-0169.}}

\maketitle
\thispagestyle{empty}
\pagestyle{empty}
\begin{abstract}
  The majority of existing Linear Temporal Logic (LTL) planning methods rely on the construction of a discrete product automaton, that combines a discrete abstraction of robot mobility and a B$\ddot{\text{u}}$chi automaton that captures the LTL specification. Representing this product automaton as a graph and using graph search techniques, optimal plans that satisfy the  LTL task can be synthesized. However, constructing expressive discrete abstractions makes the synthesis problem computationally intractable. In this paper, we propose a new sampling-based LTL planning algorithm that does not require any discrete abstraction of robot mobility. Instead, it {incrementally builds  trees} that explore the product state-space, until a maximum number of iterations is reached or a feasible plan is found. The use of trees makes {data storage} and graph search  tractable, which significantly increases the scalability of our algorithm. To accelerate the construction of feasible plans, we introduce bias in the sampling process which is guided by transitions in the B$\ddot{\text{u}}$chi automaton that belong to the shortest path to the accepting states. We show that our planning algorithm, {with and without bias,} is probabilistically complete and asymptotically optimal. Finally, we present numerical experiments showing that our method outperforms relevant temporal logic planning methods.
\end{abstract}


\section{Introduction}
\IEEEPARstart{M}{otion} planning traditionally consists of generating robot trajectories that reach a  goal region from a starting point while avoiding obstacles \cite{lavalle2006planning}. Methods for point-to-point navigation range from using potential fields and navigation functions \cite{choset2005principles,arvanitakis2016mobile} to sampling-based algorithms \cite{kuffner2000rrt,kavraki1996probabilistic,karaman2011sampling}. More recently, {new planning approaches have been proposed that} can handle a richer class of tasks, than the classical point-to-point navigation, and can capture temporal goals. Such tasks can be, e.g., sequencing or coverage \cite{fainekos2005temporal}, data gathering \cite{guo2017distributed}, intermittent communication \cite{kantaros2018distributed}, or persistent surveillance \cite{leahy2016persistent}, and can be captured using formal languages, such as Linear Temporal Logic (LTL) \cite{baier2008principles},  developed in concurrency theory.

Control synthesis for robots under complex tasks, captured by Linear Temporal Logic (LTL) formulas, builds upon either bottom-up approaches when independent LTL expressions are assigned to robots \cite{kress2009temporal, kress2007s} or top-down approaches when a global LTL formula describing a collaborative task is assigned to a team of robots \cite{chen2011synthesis, chen2012formal}, as in this work. Common in the above works is that they rely on model checking theory \cite{baier2008principles, clarke1999model}, to find plans that satisfy LTL-specified tasks, without optimizing task performance. Optimal control synthesis under local and global LTL specifications has been addressed in \cite{smith2011optimal, guo2015multi} and \cite{kloetzer2010automatic, ulusoy2013optimality, ulusoy2014optimal}, respectively. In the case of global LTL tasks \cite{kloetzer2010automatic, ulusoy2013optimality, ulusoy2014optimal}, optimal discrete plans are derived for every robot using the individual transition systems that satisfy a bisimulation property \cite{fainekos2005temporal} and are obtained through an abstraction process \cite{conner2003composition,belta2004constructing,belta2005discrete,kloetzer2006reachability,boskos2019decentralized}, and a Non-deterministic B$\ddot{\text{u}}$chi Automaton (NBA) that represents the global LTL specification. Specifically, by taking the synchronous product among the transition systems and the NBA, a synchronous product automaton can be constructed. Then,  using graph-search techniques, optimal motion plans can be derived. A major limitation is the high computational cost of constructing expressive discrete abstractions that result in very large transition systems. Combining these large transition systems with many robots and complex tasks increases dramatically the size of the product automaton so that graph-search techniques become intractable. {An additional limitation is that the resulting discrete plans are only optimal given the abstraction that was used to compute them. For a different abstraction, a different optimal plan will be computed.} Therefore, global optimality can not be ensured.

To improve on the scalability of optimal control synthesis methods such as those discussed above, we recently proposed new  more efficient sampling-based algorithms for discrete transition systems and global temporal logic tasks that avoid the construction of the product automaton altogether~\cite{kantaros15asilomar, kantaros2017sampling,kantaros2018sampling,kantaros2018temporal,kantaros2020stylus,luo2019transfer,kantaros2018distributedOpt}. {Specifically, in~\cite{kantaros15asilomar}, we transformed given transition systems into trace-included transition systems, meaning that all the behaviors they generate can also be generated by the original transition systems, that have smaller state spaces but are still expressive enough to construct feasible motion plans.}
We experimentally validated the approach in~\cite{kantaros15asilomar} in~\cite{kantaros2018control}. {In~\cite{kantaros2017sampling,kantaros2018sampling} we proposed a more tractable sampling-based method that builds trees incrementally, similar to the approach proposed here, that approximate the product automaton until a motion plan is constructed, while in~\cite{kantaros2018temporal,kantaros2020stylus} we extended the method in~\cite{kantaros2017sampling,kantaros2018sampling} by introducing bias in the sampling process in directions that make the most progress towards satisfying the assigned task. As a result, the method in~\cite{kantaros2018temporal,kantaros2020stylus} can solve planning problems that are orders of magnitude larger than the planning problems that can be solved using\cite{kantaros2017sampling,kantaros2018sampling}. In~\cite{luo2019transfer}, we improved scalability of the biased sampling-based algorithm in~\cite{kantaros2018temporal,kantaros2020stylus} by using experience from solving previous LTL planning problems.} Finally, in~\cite{kantaros2018distributedOpt}, we provided a distributed implementation of~\cite{kantaros2018sampling} whereby robots collaborate to build subtrees so that the computational time is decreased  significantly. {However,~\cite{kantaros15asilomar,kantaros2017sampling,kantaros2018sampling,kantaros2018distributedOpt,kantaros2018temporal,kantaros2020stylus,luo2019transfer} require a discrete abstraction of the robot dynamics and  the environment, which is computationally expensive to construct~\cite{belta2005discrete}.

Motivated by existing sampling-based algorithms for point-to-point navigation in continuous state spaces \cite{karaman2011sampling},  we propose a sampling-based temporal logic planning method that, unlike the methods in~\cite{kantaros15asilomar,kantaros2017sampling,kantaros2018sampling,kantaros2018distributedOpt,kantaros2018temporal,kantaros2020stylus,luo2019transfer},  does not require any discrete abstraction of robot mobility.} Our algorithm builds incrementally trees that explore both the continuous state space of robot positions and the discrete state space of the NBA, simultaneously. Specifically, we first build a tree that connects an initial state to an \textit{accepting} state of the product automaton. Tracing the nodes from the accepting state back to the root returns a plan that corresponds to the \textit{prefix} part of the plan and is executed once. Then, we construct a new tree rooted at an accepting state in a similar way and propose a cycle-detection method to discover a loop around that root, which corresponds to the \textit{suffix} part of the plan and is executed indefinitely. By construction, the continuous execution of the generated plans satisfies the  LTL-based task. Moreover, we show that our algorithm is probabilistically complete and asymptotically optimal.  Inspired by~\cite{kantaros2018temporal,kantaros2020stylus}, to accelerate the construction of a feasible solution, we also propose a biased sampling method guided by transitions in the NBA that trace shortest paths to the accepting states. The simulations show that our biased sampling method can significantly accelerate the detection of feasible but also low-cost plans. Our algorithm can be viewed as an extension of $\text{RRT}^*$ \cite{karaman2011sampling} for LTL-based tasks. {Nevertheless, completeness and optimality of our algorithm can not be trivially inherited from RRT$^*$ that is designed exclusively for continuous state spaces, or from~\cite{kantaros15asilomar,kantaros2017sampling,kantaros2018sampling,kantaros2018distributedOpt,kantaros2018temporal,kantaros2020stylus,luo2019transfer} that rely on discrete abstractions of robot mobility.}

Related are also the methods in \cite{sahin2017provably,karaman2011linear,wolff2014optimization,shoukry2017linear}.~{Particularly, \cite{sahin2017provably} formulates the planning problem as an Integer-Linear-Program  (ILP) using a discrete abstraction of the environment.} To avoid the construction of discrete abstractions and the corresponding product automata, the authors of~\cite{karaman2011linear,wolff2014optimization} encode the LTL specifications as  mixed-integer linear programming (MILP) over the continuous system variables, and employ off-the-shelf optimization solvers to obtain the optimal solution. However,~\cite{karaman2011linear} only considers a subclass of co-safe LTL formulas  that require all robots to return to their original places and stay there forever, which can be satisfied by a finite-length robot trajectory, whereas here we consider arbitrary LTL formulas. Moreover, the methods in~\cite{karaman2011linear,wolff2014optimization} require the user-specified length of the { prefix-suffix plan} that satisfies the LTL formula. On the other hand, the algorithm proposed in~\cite{shoukry2017linear} relies on  the satisfiability modulo convex (SMC) approach \cite{shoukry2018smc}. By formulating the LTL planning problem as a feasibility problem over Boolean and convex constraints, and leveraging state-of-the-art SAT solvers and convex optimization solvers, this method scales  better than the MILP-based method \cite{karaman2011linear,wolff2014optimization} by relying on a coarse abstraction of the workspace.
However, since~\cite{shoukry2017linear} formulates the  problem as a feasibility problem, this method does not have any optimality guarantees.

To the best of our knowledge, the most relevant abstraction-free LTL planning methods are presented in \cite{karaman2009sampling,karaman2012sampling,vasile2013sampling,bhatia2010sampling,DBLP:conf/icra/HeLKV15}. Specifically, in \cite{karaman2009sampling,karaman2012sampling}, sampling-based algorithms  employ an RRG-like algorithm to build incrementally a Kripke structure until it is expressive enough to generate a  plan that satisfies a task expressed in deterministic $\mu$-calculus. However, building arbitrary structures to represent transition systems compromises scalability since, as the number of samples increases, so does the density of the constructed graph which requires more resources to save and search for optimal plans using graph search methods.  Motivated by this limitation, \cite{vasile2013sampling} proposes a similar RRG algorithm, but constructs incrementally sparse graphs representing transition systems that are then used to construct a product automaton. Then correct-by-construction discrete plans are synthesized by applying graph search methods on the product automaton. However, similar to \cite{karaman2009sampling,karaman2012sampling}, as the number of samples increases, the sparsity of the constructed graph is lost. Therefore, the methods in \cite{karaman2009sampling,karaman2012sampling,vasile2013sampling } can only be used for simple planning problems. To the contrary, our sampling-based method builds trees, instead of graphs of arbitrary structure, so that optimal plans can be easily constructed by tracing the sequence of parent nodes starting from desired accepting states. Combined with the proposed biased sampling method, this allows us to handle larger problems compared to, e.g.,  \cite{vasile2013sampling}. {Our approach is also closely related to~\cite{bhatia2010sampling,DBLP:conf/icra/HeLKV15}, where high-level states  selected from a discrete product automaton state space are used to guide the low-level sampling-based planner in the continuous space. However, the biased sampling method in~\cite{bhatia2010sampling,DBLP:conf/icra/HeLKV15} relies on a discrete abstraction of robot mobility,  whereas the biased sampling method proposed here is abstraction-free and guided by the NBA, as in our previous work~\cite{kantaros2018temporal,kantaros2020stylus} for discrete state spaces. Compared to~\cite{bhatia2010sampling,DBLP:conf/icra/HeLKV15}, we show that our method is asymptotically optimal and more scalable.}


The main contribution of this paper can be summarized as follows. First, we propose a new abstraction-free temporal logic planning algorithm for multi-robot systems that is highly scalable. Scalability is due to the use of trees to represent the transition system and the proposed bias in the sampling process.  Second, we show that the proposed algorithm, with or without bias, is probabilistically complete and asymptotically optimal. Finally, we provide extensive simulation studies that show that the proposed method outperforms relevant state-of-the-art algorithms~\cite{bhatia2010sampling,vasile2013sampling,shoukry2017linear}.

The rest of the paper is organized as follows. 
In Section~\ref{sec:preliminaries} and~\ref{sec:problem} we present preliminaries and the problem formulation, respectively. In Section~\ref{sec:optimal} we describe the sampling-based planning algorithm. We introduce the biased sampling method in Section~\ref{sec:biased}. Furthermore, we examine their correctness and optimality  in Section~\ref{sec:corr}. Simulation results and conclusive remarks are presented in Sections~\ref{sec:sim} and~\ref{sec:concl}, respectively.

\section{Preliminaries}\label{sec:preliminaries}
In this section, we formally describe Linear Temporal Logic (LTL) by presenting its syntax and semantics. Also, we briefly review preliminaries of automata-based LTL model checking.

Linear temporal logic~\cite{baier2008principles} is a type of formal logic whose basic ingredients are a set of atomic propositions $\mathcal{AP}$, the boolean operators, conjunction $\wedge$ and negation $\neg$, and  temporal operators, next $\bigcirc$ and until $\mathcal{U}$. LTL formulas over $\mathcal{AP}$ abide by the grammar $\phi::=\text{true}~|~\pi~|~\phi_1\wedge\phi_2~|~\neg\phi~|~\bigcirc\phi~|~\phi_1~\mathcal{U}~\phi_2.$ {For brevity, we abstain from deriving  other Boolean and temporal operators, e.g., \textit{disjunction} $\vee$, \textit{implication} $\Rightarrow$, \textit{always} $\square$, \textit{eventually} $\lozenge$, which can be found in \cite{baier2008principles}.} 

An infinite \textit{word} $\sigma$ over the alphabet $2^{\mathcal{AP}}$ is defined as an infinite sequence  $\sigma=\pi_0\pi_1\ldots\in (2^{\mathcal{AP}})^{\omega}$, where $\omega$ denotes an infinite repetition and $\pi_k\in2^{\mathcal{AP}}$, $\forall k\in\mathbb{N}$. The language $\texttt{Words}(\phi)=\left\{\sigma|\sigma\models\phi\right\}$ is defined as the set of words that satisfy the LTL formula $\phi$, where $\models\subseteq (2^{\mathcal{AP}})^{\omega}\times\phi$ is the satisfaction relation. An LTL formula $\phi$ can be translated into a Nondeterministic B$\ddot{\text{u}}$chi Automaton  defined as follows \cite{vardi1986automata}:
\begin{defn}[NBA]
A \textit{Nondeterministic B$\ddot{\text{u}}$chi Automaton} (NBA) $B$ over $2^{\mathcal{AP}}$ is defined as a tuple $B=\left(\ccalQ_{B}, \ccalQ_{B}^0,\Sigma,\rightarrow_B,\mathcal{Q}_B^F\right)$, where $\ccalQ_{B}$ is the set of states, $\ccalQ_{B}^0\subseteq\ccalQ_{B}$ is a set of initial states, $\Sigma=2^{\mathcal{AP}}$ is an alphabet, $\rightarrow_{B}\subseteq\ccalQ_{B}\times \Sigma\times\ccalQ_{B}$ is the transition relation, and $\ccalQ_B^F\subseteq\ccalQ_{B}$ is a set of accepting/final states.
\end{defn}

An \textit{infinite run} $\rho_B$ of $B$ over an infinite word $\sigma=\pi_0\pi_1\pi_2\dots$, $\pi_k\in\Sigma=2^{\mathcal{AP}}$, $\forall k\in\mathbb{N}$, is a sequence $\rho_B=q_B^0q_B^1q_B^2\dots$ such that $q_B^0\in\ccalQ_B^0$ and $(q_B^{k},\pi_k,q_B^{k+1})\in\rightarrow_{B}$, $\forall k\in\mathbb{N}$. An infinite run $\rho_B$ is called \textit{accepting} if $\texttt{Inf}(\rho_B)\cap\ccalQ_B^F\neq\varnothing$, where $\texttt{Inf}(\rho_B)$ represents the set of states that appear in $\rho_B$ infinitely often. The words $\sigma$ that produce an accepting run of $B$ constitute the accepted language of $B$, denoted by $\ccalL_B$. Then~\cite{baier2008principles} proves that the accepted language of $B$ is equivalent to the words of $\phi$, i.e., $\ccalL_B=\texttt{Words}(\phi)$.

\section{Problem Formulation}\label{sec:problem}

Consider $N$  robots residing in a workspace, represented by a compact subset $\mathcal{W}\subset\mathbb{R}^d$, $d\in\{2,3\}$. 
 Let $\ccalO$, an open subset of $\ccalW$, denote the set of obstacles and $\ccalW_{\text{free}}=\ccalW\setminus\ccalO$ denote the free workspace. Let $\mathcal{R}=\{\ell_j\}_{j=1}^{W}$ be a set of $W\in\mathbb{N}^+$ closed labeled regions that can have any arbitrary shapes. 
  Given a labeled region of interest $\ell_j, j \in [W]$, where $[W]$ is the shorthand of $\{1,\ldots,W\}$, let $\partial\ell_j$ denote its boundary, i.e., the closure of $\ell_j$ without its interior. {The motion  of robot $i$ is captured by a Transition System (TS)  defined as follows:

\begin{defn}[TS]
  A {\it Transition System} for robot $i$, denoted by TS$_{i}$, is a tuple TS$_i = \{\ccalW, \bbx_i^0, \to_i, \ccalA\ccalP_i, L_i\}$ where: (a) $\ccalW = \{\bbx_i\}$ is the set of infinite position states of robot $i$; (b) $\bbx_i^0$ is the initial position of robot $i$; (c) $\to_i \subseteq \ccalW \times \ccalW$ is the transition relation.  We consider robot dynamics subject to holonomic constraints, as in \cite{karaman2011sampling}. Let $\overline{\bbx_{i}\bbx_{i}'}$ denote the straight line connecting the points $\bbx_{i}$ and $\bbx_{i}'$. A transition from $\bbx_{i}$ to $\bbx'_i$, denoted by $(\bbx_{i}, \bbx'_{i})\in\rightarrow_i$, exists if (i) $\xx$ is inside $\ccalW_{\text{free}}$ and (ii) it crosses any boundary $\partial \ell_j, j\in[W]$ at most once, {as in~\cite{vasile2013sampling}};\footnote{{Suppose $\bbx_i \in \ell_m$, $\bbx_i' \in \ell_n$ and $\overline{\bbx_i \bbx_i'} \subset \ccalW_{\text{free}}$ crosses region $\ell_k$ twice, further consider a  task that requires to always avoid $\ell_k$. If the transition depends only on condition (i) whether $\overline{\bbx_i \bbx_i'}$ is inside $\ccalW_{\text{free}}$, then we have $(\bbx_i, \bbx_i') \in \rightarrow_i$. However, the path corresponding to $\overline{\bbx_i \bbx_i'}$ crosses region $\ell_k$ thus violating the task. The transition condition (ii) excludes this case  where the discrete plan satisfies the task but the underlying continuous path does not, because it passes through an undesired region between two   endpoints.}}
    (d) $\mathcal{AP}_i=\bigcup_{j=1}^W\{\pi_{i}^{\ell_j}\} \cup \{\varnothing\}$ is the set of atomic propositions, where $\pi_{i}^{\ell_j}$ is true if robot $i$ is inside region $\ell_j$ {(including its boundary)} and false otherwise, and $\varnothing$ means robot $i$ is not inside any labeled region; (e) $L_{i}:\mathcal{W}\rightarrow {\mathcal{AP}_i}$  is the observation (labeling) function  that returns an atomic proposition satisfied at a location $\bbx_{i}$, i.e., $L_i(\bbx_{i})=\pi_i^{\ell_j} ~\text{or}~ \varnothing$.
\end{defn}

{We emphasize that a TS is defined over the continuous state space.} Given a TS,  we define the Product Transition System (PTS), which captures all possible combinations of robot behaviors captured by their respective TS$_{i}$, as follows:
\begin{defn}[PTS]
  Given $N$ transition systems TS$_i = \{\ccalW, \bbx_i^0, \to_i, \ccalA\ccalP_i, L_i\}$, the product transition system $\text{PTS} = \text{TS}_1 \times  \text{TS}_2 \times \cdots \times  \text{TS}_N$ is a tuple $\text{PTS} = (\ccalW^N, \bbx_0, \to_{\text{PTS}}, C, \ccalA\ccalP,L_{\text{PTS}})$ where (a) $\ccalW^N = \ccalW \times \cdots \times \ccalW$ is the set of states and $\bbx=[\bbx_{1}^{T},\bbx_{2}^{T},\dots,\bbx_{N}^{T}]^T \in\mathcal{W}^N$ is a vector that stacks the positions ${\bf{x}}_i$ of all robots $i\in[N]$; (b) $\bbx_0 = [\bbx_{1}^{0,T},\dots,\bbx_{N}^{0,T}]^T\in\mathcal{W}^N$ represents the initial positions of all robots; (c) $\longrightarrow_{\text{PTS}}\subseteq\mathcal{W}^N\times\mathcal{W}^N$ is the transition relation defined by the rule $\frac{\bigwedge _{\forall i}\left(\bbx_{i}\rightarrow_{i}\bbx_{i}'\right)}{\bbx\rightarrow_{\text{PTS}}\bbx'}$, $\forall\, \bbx, \bbx' \in \ccalW^N$. In words, there exists a transition from $\bbx$ to $\bbx'$ if there exists a transition from $\bbx_i$ to $\bbx_i'$ for {all} $i\in[N]$; (d) $C: \ccalW^N\times\ccalW^N\rightarrow\mathbb{R}^+$ is the cost of traveling from one point to another for all robots. {The cost function can be defined arbitrarily as long as it is monotonic. Examples include traveled distance or energy consumption.} For holonomic planning, we use the Euclidean distance between $\bbx, \bbx'\in\ccalW^N$  as the cost function, i.e.,
\begin{align}\label{eq:c}
  C(\bbx,\bbx') = \|\bbx - \bbx'\|_2;
\end{align}
(e) $\mathcal{AP}=\bigcup_{i=1}^N\mathcal{AP}_i$ is the set of atomic propositions; (f)  $L=\bigcup_{i}^N L_{i}: \mathcal{W}^N\rightarrow{2^{\mathcal{AP}}}$ is an observation function that returns the set of atomic propositions satisfied at a state $\bbx\in\mathcal{W}^N $.
\end{defn}}

Given the PTS and the NBA $B$, we can define the \textit{Product B$\ddot{\text{u}}$chi Automaton} (PBA) $P = \text{PTS} \times B$ as follows~\cite{baier2008principles}:

\begin{defn}[PBA]\label{defn:pba}
The \textit{Product B$\ddot{\text{u}}$chi Automaton} is defined by the tuple $P=(\mathcal{Q}_P, \mathcal{Q}_P^0,\longrightarrow_{P},\mathcal{Q}_P^F, C)$, where
(a) $\mathcal{Q}_P=\ccalW^N\times\mathcal{Q}_{B}$ is an infinite set of states; (b) $\mathcal{Q}_P^0=\bbx_0\times\mathcal{Q}_B^0$ is a set of initial states;
(c) $\longrightarrow_{P}\subseteq\mathcal{Q}_P\times 2^{\mathcal{AP}}\times\mathcal{Q}_P$ is the transition relation defined by the rule: $\frac{(\bbx\rightarrow_{\text{PTS}} \bbx')\wedge(q_{B}\xrightarrow{L\left(\bbx \right)}_{B}q_{B}')}{q_{P}=\left(\bbx,q_{B}\right)\longrightarrow_P q_{P}'=\left(\bbx',q_{B}'\right)}$. The transition from the state $q_P\in\mathcal{Q}_P$ to $q_P'\in\mathcal{Q}_P$, is denoted by $(q_P,q_P')\in\longrightarrow_P$, or $q_P\longrightarrow_P q_P'$;
(d) $\mathcal{Q}_P^F=\ccalW^N\times\mathcal{Q}_B^F$ is a set of accepting/final states;
(e) With a slight abuse of notation, $C$ is the cost function between two product states, defined as the cost between respective configurations in $\ccalW^N$.
\end{defn}

In this paper, we restrict our attention to LTL formulas that exclude the `next' temporal operator, denoted by LTL$_{-\bigcirc}$.\footnote{The syntax of LTL$_{-\bigcirc}$ is the same as the syntax of LTL excluding the ‘next’ operator. The choice of LTL$_{-\bigcirc}$ over LTL is motivated by the fact that we are interested in the continuous time execution of the synthesized plans, in which case the next operator is not meaningful. {For example, to satisfy the formula $\Diamond (\pi_1^{\ell_1} \wedge \bigcirc \pi_{1}^{\ell_2})$, robot 1 should visit  region $\ell_2$ right after visiting $\ell_1$, which is not possible in continuous time since this requires instantaneous motion.} This choice of LTL$_{-\bigcirc}$ is common in relevant works, see, e.g., \cite{kloetzer2008fully} and the references therein.} \ltl $\,$  formulas are satisfied by discrete plans $\tau$ that are infinite sequences of locations of $N$ robots in $\mathcal{W}^N_{\text{free}}$, i.e., $\tau=\tau(1)\tau(2)\dots\tau(k)\dots$, where $\tau(k)\in\mathcal{W}^N_{\text{free}}$. {If an LTL$_{-\bigcirc}$ formula is satisfiable, then there exists  a plan $\tau$ that can be written  in a finite representation, called {\em prefix-suffix} structure, i.e., $\tau=\tau^{\text{pre}}[\tau^{\text{suf}}]^\omega$ where the prefix part  $\tau^{\text{pre}}=\tau(1)\tau(2)\dots\tau(K)$ is executed only once followed by the indefinite execution of the suffix part $\tau^{\text{suf}}=\tau(K)\tau(K+1)\dots\tau(K+S)\tau(K+S+1)$,  where $\tau(K+S+1)=\tau(K)$~\cite{baier2008principles}.} A discrete plan $\tau$ satisfies $\phi$ if the trace generated by $\tau$, defined as $\texttt{trace}(\tau):=L(\tau(1))\dots L(\tau(K))[L(\tau(K))\dots L(\tau(K+S+1))]^{\omega}$, belongs to $\texttt{Words}(\phi)$. We denote $\texttt{trace}(\tau)\in\texttt{Words}(\phi)$ by $\tau\models\phi$. 

Given a discrete plan ${\tau}$, either finite or infinite, a continuous path $\tilde{\tau}$ can be generated by joining any two consecutive points in $\tau$ using a line segment. In this sense, the continuous path $\tilde{\tau}$ essentially captures the execution of the discrete plan $\tau$. Let $\tilde{\tau}:[0,1]\to\ccalW_{\text{free}}^N$ be a parameterized continuous path, where $\tilde{\tau}(0)=\bbx_0$. We denote its trace by $\texttt{trace}(\tilde{\tau}):=\{L(\tilde{\tau}(s_i))\}_{i=0}^n$, where
$s_{i+1} = \inf \{ s\in[s_i, 1] \,|\, \, L(\tilde{\tau}(s)) \not= L(\tilde{\tau}(s_i)) \}$ and $s_0 = 0$, $L(\tilde{\tau}(s_n)) = L(\tilde{\tau}(1))$. Like the prefix-suffix structure in the discrete plan $\tau$, a continuous path $\tilde{\tau}$ satisfying $\phi$ can be written as $\tilde{\tau} = \tilde{\tau}_1|\tilde{\tau}_2|\tilde{\tau}_2|\cdots$, where $\tilde{\tau}_1$ and $\tilde{\tau}_2$ are prefix and suffix paths, respectively, $\tilde{\tau}_1(1)=\tilde{\tau}_2(0)=\tilde{\tau}_2(1)$, and  $|$ stands for the concatenation of two paths. Hence, $\texttt{trace}(\tilde{\tau}) = \texttt{trace}(\tilde{\tau}_1) [\texttt{trace}(\tilde{\tau}_2)]^{\omega}$. The transition relation defined before between any two configurations $\mathbf{x}$ and $\mathbf{x}'$ ensures that {if $\tau$ satisfies $\phi$, so does $\tilde{\tau}$. Thus, we can focus on finding discrete feasible plans $\tau$ for control.}

Furthermore, we define the {\it cost} of a continuous path ${\tilde{\tau}}$ as~\cite{karaman2011sampling}
\begingroup\makeatletter\def\f@size{10}\check@mathfonts
\def\maketag@@@#1{\hbox{\m@th\normalsize\normalfont#1}}%
\begin{align}\label{eq:cost2}
  J({\tilde{\tau}}) = \sup_{\{ m\in\mathbb{N}, 0=s_0 < s_1 < \cdots < s_m = 1\} } \sum_{i=1}^m {C({\tilde{\tau}}(s_{i}), {\tilde{\tau}}(s_{i-1}))}.
\end{align}
\endgroup
In essence, this cost is equal to the Euclidean distance traversed by the continuous path $\tilde{\tau}$ in $\ccalW^N$. The cost of the discrete plan $\tau$ corresponding to $\tilde{\tau}$ can be defined equivalently, as the length of a piecewise linear path in $\ccalW^N$. Specifically, we define the cost of a prefix-suffix plan $\tau = \tau^{\text{pre}} [\tau^{\text{suf}}]^w$  as:
\begingroup\makeatletter\def\f@size{10}\check@mathfonts
\def\maketag@@@#1{\hbox{\m@th\normalsize\normalfont#1}}%
\begin{align}\label{eq:cost1}
  J({\tau})&= w J({\tau}^{\text{pre}}) +  (1 - w) J({\tau}^{\text{suf}}),
\end{align}
\endgroup
{where $w\in [0,1]$ is a user-specified parameter that assigns different weights to the cost associated with the prefix plan and recurrent cost of the suffix plan.}

The problem addressed in this paper is stated as follows.
\begin{problem}\label{pr:problem}
Given the product space $P$, composed of the transition system of robots  and the NBA derived from a global {LTL$_{-\bigcirc}$} specification $\phi$ defined over the set $\mathcal{AP}$, determine a plan $\tau$ that minimizes $J(\tau)$ and satisfies $\phi$.
\end{problem}

\section{Sampling-Based Optimal Control Synthesis}\label{sec:optimal}
In this section, we propose a sampling-based algorithm to solve Problem~\ref{pr:problem}, called TL-RRT$^*$ for Temporal Logic RRT$^*$, that incrementally constructs trees that explore the product state space $\ccalQ_P =\ccalW^N\times\ccalQ_B$. These trees are used to design a feasible plan $\tau=\tau^{\text{pre}}[\tau^{\text{suf}}]^{\omega}$, which satisfies the desired LTL formula $\phi$ and minimizes the cost function~\eqref{eq:cost1}. Our algorithm,  inspired by the RRT$^*$ method for point-to-point navigation~\cite{karaman2011sampling}, is described in Alg.~\ref{alg:plans}. In Alg.~\ref{alg:plans}, first the LTL formula is translated to an NBA $B=\{\ccalQ_B,\ccalQ_B^0,\rightarrow_B,\ccalQ_B^F\}$ [line~\ref{alg1:line1}, Alg.~\ref{alg:plans}]. Then, in lines~\ref{alg1:line2}-\ref{alg1:line6}, candidate prefix {plans} $\tau^{\text{pre},a}$ are constructed, followed by the construction of corresponding suffix {plans} $\tau^{\text{suf},a}$ in lines~\ref{alg1:line7}-\ref{alg1:line14}. Finally, the optimal plan $\tau=\tau^{\text{pre},a^{*}}[\tau^{\text{suf},a^{*}}]^{\omega}\models\phi$ is synthesized in lines~\ref{alg1:line15}-\ref{alg1:line16}.

\begin{algorithm}[t]
\caption{TL-RRT$^*$}
\LinesNumbered
\label{alg:plans}
\KwIn {{Product transition system $\text{PTS}$}, {LTL} formula $\phi$, maximum numbers of iterations $n_{\text{max}}^{\text{pre}}$, $n_{\text{max}}^{\text{suf}}$}
\KwOut {Optimal plans $\tau\models\phi$}
Convert $\phi$ to an NBA $B=\left(\ccalQ_B,\ccalQ_B^0,\rightarrow_B,\ccalQ_B^F\right)$\;\label{alg1:line1}
\Comment*[r]{Construction of Prefix Plans \ref{opt:pre}}
Define goal set: $\ccalQ_{\text{goal}}^{\text{pre}}$\;\label{alg1:line2}
Initial state: $q_P^0=(\bbx^0,q_B^0)$\;\label{alg1:line3}
$\left[\ccalT,\ccalP\right]=\texttt{ConstructTree}(\ccalQ_{\text{goal}}^{\text{pre}}, \textup{PTS},B,q_P^0,n_{\text{max}}^{\text{pre}})$\;\label{alg1:line4}
\For {$a=1:|\ccalP|$}{\label{alg1:line5}
$\tau^{\text{pre},a}=\texttt{FindPlan}(\ccalT,q_P^0,\mathcal{P}(a))$\;}\label{alg1:line6}
\Comment*[r]{Construction of Suffix Plans \ref{sec:suffix}}
\For {$a=1:|\ccalP|$ }{\label{alg1:line7}
  Initial state: $q_P^0=\ccalP(a)$\;\label{alg1:line8}
Define goal set: $\ccalQ_{\text{goal}}^{\text{suf}}(q_P^0)$\;\label{alg1:line9}
\If{\upshape $q_P^0\in\ccalQ_{\text{goal}}^{\text{suf}}$}{\label{alg1:line9a}
$\ccalT=(\{q_P^0\},\{q_P^0,q_P^0\},0)$, $\ccalS_{a}=\{q_P^0\}$\;\label{alg1:line9b}
}
\Else{\label{alg1:line9d}
$\left[\ccalT,\ccalS_{a}\right]=\texttt{ConstructTree}(\ccalQ_{\text{goal}}^{\text{suf}}, \textup{PTS},B,q_P^0,n_{\text{max}}^{\text{suf}})$\;}\label{alg1:line10}
\For {$e=1:|\ccalS_{a}|$}{\label{alg1:line11}
${\tau}_a^{\text{suf},e}=\texttt{FindPlan}(\ccalT,q_P^0,\mathcal{S}_{a}(e))$\;\label{alg1:line12} }
$e^{*}=\argmin_e(J(\tau_a^{\text{suf},e}))$\;\label{alg1:line13}
$\tau^{\text{suf},a}={\tau}_a^{\text{suf},e^{*}}$\;}\label{alg1:line14}
\Comment*[r]{Construction of optimal plans \ref{sec:plan}}
$a^*=\argmin_{a}(J(\tau^{\text{pre},a})+J(\tau^{\text{suf},a}))$\;\label{alg1:line15}
\Return {\upshape $\tau=\tau^{\text{pre},a^{*}}[\tau^{\text{suf},a^{*}}]^{\omega}$}\;\label{alg1:line16}
\end{algorithm}

\subsection{Construction of Prefix Plans}\label{opt:pre}
In this section, we describe how the prefix plan is constructed [lines~\ref{alg1:line2}-\ref{alg1:line6}, Alg.~\ref{alg:plans}] by building incrementally a tree, denoted by $\ccalT=\{\ccalV_{\ccalT},\ccalE_{\ccalT},\texttt{Cost}\}$, that explores the state space $\ccalQ_P$. The set $\ccalV_{\ccalT}$ consists of nodes $q_P=\left({\bf{x}},q_B\right)\in\mathcal{W}_{\text{free}}^N\times\mathcal{Q}_B$. The set of edges $\ccalE_{\ccalT}$ captures transitions among the nodes in $\ccalV_{\ccalT}$, i.e., $(q_P,q_P')\in\ccalE_{\ccalT}$, if $q_P \rightarrow_P q'_P $, for $q_P, q_P'\in\ccalV_{\ccalT}$.  Moreover, the function $\texttt{Cost}:\ccalV_{\ccalT} \rightarrow\mathbb{R}^+$ assigns the cost of reaching node $q_P$ from the root. Let ${\tau}$ denote the plan/path from the root to $q_P$ in the tree, then $\texttt{Cost}(q_P) = J({\tau})$.\footnote{We also refer a path  to a sequence of connecting nodes in a graph, which can  be  distinguished easily from the continuous path of a discrete plan.}

{The tree $\ccalT$ is rooted at the initial state $q_P^0=(\bbx^0,q_B^0)$ where ${\bf{x}}^0\in\mathcal{W}_{\text{free}}^N$ and $q_B^0\in\ccalQ_B^0$.\footnote{{In what follows, for the sake of simplicity we assume that the set of initial states $\ccalQ_B^0$ consists of only one state, i.e., $\ccalQ_B^0=\{q_B^0\}$. In case there are more than one initial states in $B$, then the lines~\ref{alg1:line2}-\ref{alg1:line6} in Alg.~\ref{alg:plans} should be executed for each possible initial state $q_B^0.$}} We define the goal region as
  \begingroup\makeatletter\def\f@size{10}\check@mathfonts
\def\maketag@@@#1{\hbox{\m@th\normalsize\normalfont#1}}%
\begin{equation}\label{eq:goalPre}
\ccalQ_{\text{goal}}^{\text{pre}}=\{q_P=(\bbx,~q_B)\in\ccalW^N_{\text{free}}\times\ccalQ_B~|~q_B\in\mathcal{Q}_B^{F}\},
\end{equation}
\endgroup
i.e., the goal region collects all final states of the PBA.
To construct a prefix plan, the tree $\ccalT$ attempts to find a sequence of states in $\ccalW_{\text{free}}^N\times\ccalQ_B$ that connects an initial state $q_P^0$ to any \textit{accepting} state in $\ccalQ_{\text{goal}}^{\text{pre}}$.  In Alg.~\ref{alg:tree}, the set $\ccalV_{\ccalT}$ initially contains only the initial state $q_P^0$. The set of edges is initialized as $\ccalE_{\ccalT}=\varnothing$. By convention, the cost of $q_P^0$ is zero [line~\ref{tree:line3}, Alg.~\ref{alg:tree}]. Below we describe the incremental construction of the tree $\ccalT=\{\ccalV_{\ccalT},\ccalE_{\ccalT},\texttt{Cost}\}$ [lines~\ref{tree:line5}-\ref{tree:line17}, Alg.~\ref{alg:tree}].

\subsubsection{Sampling a state \upshape$\bbx^{\text{rand}}$} \label{sec:sample}
The first step to construct the tree $\ccalT$ is to sample a state $\bbx^{\text{rand}}\in\ccalW_{\text{free}}^N$. We generate independently one sample from $\ccalW_{\text{free}}$ for each robot and stack these samples to construct $\x{rand}$. Any distribution can be used to get $\x{rand}$ as long as it is bounded away from zero on $\ccalW_{\text{free}}$.
\begin{algorithm}[t]
  \caption{{Function $\texttt{ConstructTree}(\ccalQ_{\text{goal}},$ $\textup{PTS}, B,~q_P^0,~n_{\max})$}}
\LinesNumbered
\label{alg:tree}
$\ccalV_{\ccalT}=\{q_P^0\}$; $\ccalE_{\ccalT}=\varnothing$; $\texttt{Cost}(q_P^0)=0$; {$\x{new} = \bbx_0$}\;\label{tree:line3}
\For {$n=1:n_{\max}$}{ \label{tree:line5}
  \While{\upshape $\x{new} \not\in \ccalW_{\text{free}}^N$}{\label{tree:line5.6}
    $\bbx^{\text{rand}}=\texttt{Sample}(\ccalW_{\text{free}})$  \label{tree:line6} \;
    {$\left\{(\bbx^{\text{nearest}}, q_B^{\text{nearest}} )\right\} =\texttt{Nearest}(\ccalT,\bbx^{\text{rand}})$  \label{tree:line7} \; }
$\bbx^{\text{new}}=\texttt{Steer}(\bbx^{\text{nearest}},\bbx^{\text{rand}})$  \; \label{tree:line8}}
  $\ccalQ_P^{\text{near}}=\texttt{Near}(\x{new}, \ccalT) \cup \left\{(\bbx^{\text{nearest}}, q_B^{\text{nearest}} )\right\}$ \;\label{tree:line8.5}
        \For {$b=1:|\mathcal{Q}_B|$}{\label{tree:line9}
                         $q_B^{\text{new}}=\mathcal{Q}_B(b)$, \quad
                         $q_{P}^{\text{new}}=(\bbx^{\text{new}},q_B^{\text{new}})$\;\label{tree:line11}
                          $\left[\ccalT,\texttt{added}\right]=\texttt{Extend}(q_{P}^{\text{new}},\ccalQ_P^{\text{near}}, \ccalT)$ \;  \label{tree:line12}
        \If{\upshape $\texttt{added}=1$}{\label{tree:line13}
        $\left[\ccalT,\texttt{Cost}\right]=\texttt{Rewire}(q_P^{\text{new}},\ccalQ_P^{\text{near}}, \ccalT)$ \; \label{tree:line17}
        }}
       } 
     {$\ccalP=\ccalV_{\ccalT}\cap\ccalQ_{\text{goal}}$\;}\label{tree:line15}
     \Return{$\ccalT, \ccalP$}\;
\end{algorithm}

\subsubsection{Constructing a state \upshape$\x{new}\in\ccalW_{\text{free}}^N$}\label{sec:new}
Given the state $\bbx^{\text{rand}}\in\ccalW_{\text{free}}^N$, we construct the state $\bbx^{\text{new}}\in\ccalW^N_{\text{free}}$. To do so, we first find the set $\ccalQ_P^{\text{nearest}}\subseteq\ccalV_{\ccalT}$, which consists of nodes $q_P^{\text{nearest}}=(\bbx^{\text{nearest}},q_B^{\text{nearest}})\in\ccalV_{\ccalT}$, that are the closest to $\bbx^{\text{rand}}$ in terms of a given distance function [line~\ref{tree:line7}, Alg.~\ref{alg:tree}]. Here we employ the Euclidean distance. Note that $\ccalQ_P^{\text{nearest}}$ may include multiple nodes with the same $\bbx^{\text{nearest}}$ but different $q_B^{\text{nearest}}$. We identify the nodes in $\ccalQ_P^{\text{nearest}}$ using the function $\texttt{Nearest}:\ccalW^N_{\text{free}}\rightarrow 2^{\ccalV_{\ccalT}}$ defined as
\begingroup\makeatletter\def\f@size{10}\check@mathfonts
\def\maketag@@@#1{\hbox{\m@th\normalsize\normalfont#1}}%
\begin{align}
  \texttt{Nearest}(\bbx^{\text{rand}})=\argmin_{q_P \in \ccalV_{\ccalT}}\Big\|\bbx^{\text{rand}}- q_P|_{\ccalW^N_{\text{free}}} \Big\|,
\end{align}
\endgroup
{where $(\cdot)|_{(\cdot)}$ stands for the projection, i.e., $q_P|_{\ccalW^N_{\text{free}}}\in\ccalW^N_{\text{free}}$.}

Next, given the state $\bbx^{\text{nearest}}\in\ccalW_{\text{free}}^N$, we construct the state $\bbx^{\text{new}}\in\ccalW^N_{\text{free}}$ using the function $\texttt{Steer}:\ccalW_{\text{free}}^N\times\ccalW^N_{\text{free}}\rightarrow\ccalW^N$ [line~\ref{tree:line8}, Alg.~\ref{alg:tree}]. Specifically, the state $\bbx^{\text{new}}$ is within distance at most $\eta>0$ from $\bbx^{\text{nearest}}$, i.e., $\left\|\bbx^{\text{nearest}}-\bbx^{\text{new}}\right\|\leq\eta$, where $\eta$ is user-specified, and should also satisfy  $\left\|\bbx^{\text{new}}-\bbx^{\text{rand}}\right\|\leq\left\|\bbx^{\text{nearest}}-\bbx^{\text{rand}}\right\|$. {The step size $\eta$ should consider  the clutter in  the environment, the maximum control input, and the duration of the time unit when tracking the  path using a robot controller.}   Furthermore, we check if $\x{new}$ lies in the obstacle-free space, i.e., $\x{new} \in \ccalW_{\text{free}}^N$. If not, we return to the \texttt{Sample} step.

\subsubsection{Constructing the node set \upshape$\ccalQ_P^{\text{near}}$}\label{sec:near}
Given  $\x{new} \in \ccalW_{\text{free}}^N$, we construct the set $\ccalQ_P^{\text{near}}\subseteq\ccalV_{\ccalT}$ [line~\ref{tree:line8.5}, Alg.~\ref{alg:tree}] that {is the union of the set $\ccalQ_P^{\text{nearest}}$ and the set}
\begingroup\makeatletter\def\f@size{10}\check@mathfonts
\def\maketag@@@#1{\hbox{\m@th\normalsize\normalfont#1}}%
\begin{align}\label{eq:near}
  \texttt{Near}(\x{new}, \ccalT) =  \Large\{{q}_P= &({\bbx}, {q}_B)\in\ccalV_{\ccalT}~| \nonumber \\ & \left\|{\bbx}
  -\bbx^{\text{new}}\right\|\leq r_n(\ccalV_{\ccalT})\Large\},
\end{align}
\endgroup
{which collects nodes that lie within a certain distance $r_n(\ccalV_T)$ from $\x{new}$, where the connection radius $r_n(\ccalV_{\ccalT})$ is selected as}
\begingroup\makeatletter\def\f@size{10}\check@mathfonts
\def\maketag@@@#1{\hbox{\m@th\normalsize\normalfont#1}}%
\begin{equation}\label{eq:r}
r_n(\ccalV_{\ccalT})=\min\Big\{\gamma_{\text{TL-RRT}^*}\Big(\frac{\log\left|[\ccalV_{\ccalT}]_{\sim}\right|}{\left|[\ccalV_{\ccalT}]_{\sim}\right|}\Big)^{\frac{1}{\texttt{dim}+1}}, \eta\Big\}.
\end{equation}
\endgroup
{In~\eqref{eq:r}, $\eta>0$, and $[\ccalV_{\ccalT}]_{\sim}$ is the equivalence class defined as
  $[\ccalV_{\ccalT}]_{\sim} = \{\x{} \,|\,(\x{} ,q_B)\in\ccalV_{\ccalT}, \forall q_B \in \ccalQ_B\}$, i.e., $[\ccalV_\ccalT]_{\sim}$ collects all $\x{}$'s appearing in the tree.} Also, $|\cdot|$ is the cardinality of a set, $\texttt{dim}$ is the dimension of the product workspace $\ccalW^N$, i.e., $\texttt{dim}=d\cdot N$,  and $\gamma_{\text{TL-RRT}^*}$ is a constant as
  \begingroup\makeatletter\def\f@size{10}\check@mathfonts
  \def\maketag@@@#1{\hbox{\m@th\normalsize\normalfont#1}}%
  \begin{align}\label{gamma}
  \gamma_{\text{TL-RRT}^*}\geq (2 + \theta) \Big( \frac{(1+\epsilon/4)J(\tau^*)}{(\texttt{dim}+1)\theta (1-\kappa)} \cdot   \frac{\mu(\ccalW_{\text{free}}^N)}{\zeta_{\texttt{dim}}}\Big)^{\frac{1}{\texttt{dim}+1}},
  \end{align}
  \endgroup
      {{where $\epsilon \in (0,1)$, $\theta\in (0, 1/4)$ and  $\kappa\in(0,1)$ are three ratio terms,} $\tau^*$ is the optimal path, $\tau^{*,\text{pre}}$ for the prefix part and $\tau^{*,\text{suf}}$ for the suffix part, $\mu(\ccalW_{\text{free}}^N)$ is the measure of $\ccalW_{\text{free}}^N$, and $\zeta_{\texttt{dim}}$ is the volume of the unit sphere in $\mathbb{R}^{\texttt{dim}}$. {Intuitively, as the size of free workspace or the cost of the optimal path become larger, $\gamma_{\textup{TL-RRT}^*}$ also becomes larger, which leads to a larger connection radius so that more nodes are considered when adding the new state to the tree and further improving the tree, thus covering more space; see the following steps~\hyperref[sec:extend]{4)} and \hyperref[sec:rewire]{5)}.}
        {The lower bound of $\gamma_{\text{TL-RRT}^*}$ is the same as that in~\cite{solovey2020revisiting} and will be discussed in Theorem~\ref{prop:opt}}.
}}

\begin{figure}[t]
  \centering
  \includegraphics[width=0.5\linewidth]{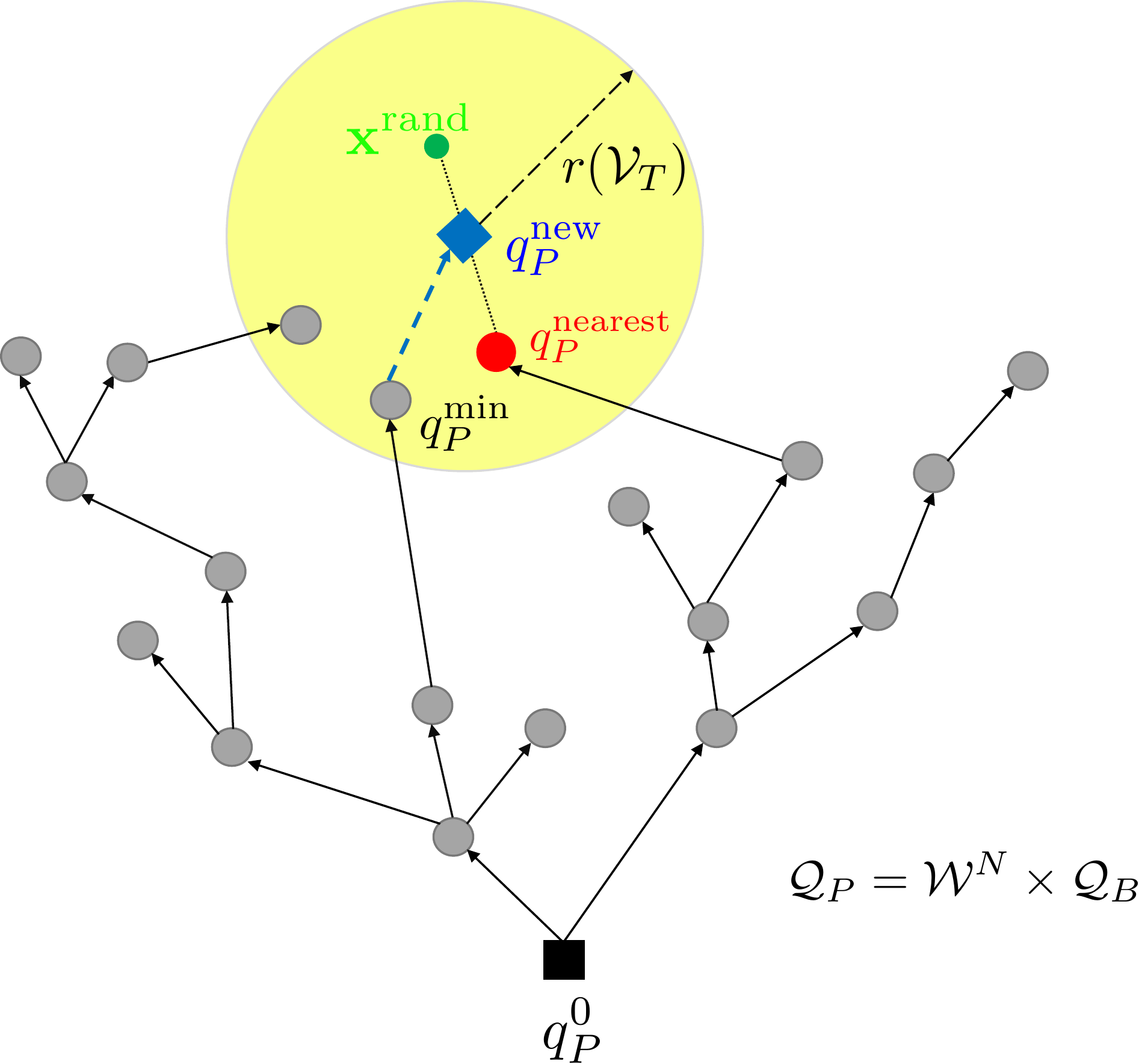}
  \caption{Graphical depiction of Alg.~\ref{alg:extend}. The black square is the root of the tree and the gray and red disks represent nodes in the set $\ccalV_{\ccalT}$.  The  green dot is the state $\bbx^{\text{rand}}$ generated by function $\texttt{Sample}$. The blue diamond and the red dot  stand for states $q_P^{\text{new}}$ and $q_P^{\text{nearest}}$, respectively. {$\q{P}{new}$ is connected to the tree $\ccalT$ through the node $\q{P}{min}$ that incurs the smallest cost from the root (see the blue dashed edge), which may not be the node $\q{P}{nearest}$.}}
  \label{fig:ext}
\end{figure}

After obtaining $\x{new}$ and $\ccalQ_{P}^{\text{near}}$, we pair a B$\ddot{\text{u}}$chi state $\q{B}{new}$ with $\x{new}$ to construct a state $q_P^{\text{new}}=({\bf{x}}^{\text{new}},q_B^{\text{new}})$ which will be examined if it can be added to the tree $\ccalT$ through nodes in $\ccalQ_{P}^{\text{near}}$. The addition is accomplished by the function $\texttt{Extend}$ described in Alg.~\ref{alg:extend}. {This procedure  is repeated for all states $(\bbx^{\text{new}},\ccalQ_B(b))$} [line~\ref{tree:line9}, Alg.~\ref{alg:tree}], where $\ccalQ_B(b)$ denotes the $b$-th state in the set $\ccalQ_B$ for $b\in[|\ccalQ_B|]$, assuming an arbitrary enumeration of the elements in $\Qb$ [line~\ref{tree:line11}, Alg.~\ref{alg:tree}]. Appending $q_B^{\text{new}}=\ccalQ_B(b)$ to $\bbx^{\text{new}}$, we construct the state $q_P^{\text{new}}=({\bf{x}}^{\text{new}},q_B^{\text{new}})$.  In what follows, we describe the function \texttt{Extend} for a given state $q_P^{\text{new}}$, which is illustrated in Fig.~\ref{fig:ext}.

\subsubsection{Extending $\ccalT$ towards \upshape$q_P^{\text{new}}$}\label{sec:extend}
\begin{algorithm}[t]
\caption{Function $\texttt{Extend}(q_{P}^{\text{new}},\ccalQ_P^{\text{near}}, \ccalT)$}
\label{alg:extend}
                         $\texttt{added}=0$; $c=\infty$\label{ext:line3}
\For { \upshape \textit{all}$~$ $q_P^{\text{near}}=(\bbx^{\text{near}},q_B^{\text{near}})\in \ccalQ_P^{\text{near}}$ }{\label{ext:line4}
                    \If{\upshape $\bbx^{\text{near}} \rightarrow \bbx^{\text{new}} \wedge (q_B^{\text{near}},L(\bbx^{\text{near}}),q_B^{\text{new}})\in\rightarrow_B$}{\label{ext:line5}
                    $c'=\texttt{Cost}(q_P^{\text{near}})+C(\bbx^{\text{near}},\bbx^{\text{new}})$\;\label{ext:line6}
                    \If{$c'\leq c$}{\label{ext:line7}
                    $\texttt{added}=1$; $q_P^{\text{min}}=q_P^{\text{near}}$\;\label{ext:line9}
                    $c=c'$; $\texttt{Cost}(q_P^{\text{new}})=c'$\;\label{ext:line11}}}}
                    \If{\upshape $\texttt{added}=1$}{\label{ext:line12}
                    $\ccalV_{\ccalT}=\ccalV_{\ccalT}\cup \{q_P^{\text{new}}\}$,
                    $\ccalE_{\ccalT}=\ccalE_{\ccalT}\cup\{(q_P^{\text{min}},q_P^{\text{new}})\}$ \;\label{ext:line13}
                    }
                    \Return {$\ccalT$, \upshape $\texttt{added}$}\;
\end{algorithm}

In lines~\ref{ext:line4}-\ref{ext:line11} of Alg.~\ref{alg:extend}, we select the parent of the state $q_P^{\text{new}}$ among $ \ccalQ_P^{\text{near}}$. Specifically, for each state $q_P^{\text{near}}=(\bbx^{\text{near}},q_B^{\text{near}})\in \ccalQ_P^{\text{near}}$, we check (i) if $\bbx^{\text{near}} \rightarrow \bbx^{\text{new}}$, and (ii) if $(q_B^{\text{near}},L(\bbx^{\text{near}}),q_B^{\text{new}})\in\rightarrow_B$ [line~\ref{ext:line5}, Alg.~\ref{alg:extend}]. In words, we check whether $q_P^{\text{near}}$ is a candidate parent of $q_P^{\text{new}}$. If so, the cost of the state $q_P^{\text{new}}$ is $\texttt{Cost}(q_P^{\text{near}}) + C(\bbx^{\text{near}},\bbx^{\text{new}})$ [line~\ref{ext:line6}, Alg.~\ref{alg:extend}],
where $\texttt{Cost}(q_P^{\text{near}})$ is the cost of reaching $q_P^{\text{near}}$ from  $q_P^0$, and $C(\bbx^{\text{near}},\bbx^{\text{new}})$ from~\eqref{eq:c} is the cost of reaching $q_P^{\text{new}}$ from $q_P^{\text{near}}$.
Among all candidate parents $q_P^{\text{near}}$ of $q_P^{\text{new}}$, we pick the one that incurs the minimum cost $\texttt{Cost}(q_P^{\text{new}})$, denoted by $q_P^{\text{min}}$ [lines~\ref{ext:line7}-\ref{ext:line11}, Alg.~\ref{alg:extend}]. If the set of candidate parents is not empty, then the sets $\ccalV_{\ccalT}$, $\ccalE_{\ccalT}$ are updated as: $\ccalV_{\ccalT}=\ccalV_{\ccalT}\cup \{q_P^{\text{new}}\}$ and $\ccalE_{\ccalT}=\ccalE_{\ccalT}\cup\{(q_P^{\text{min}},q_P^{\text{new}})\}$ [lines~\ref{ext:line13}, Alg.~\ref{alg:extend}]. If  $q_P^{\text{new}}$ is added to $\ccalV_{\ccalT}$, the \textit{rewiring} process follows.

\subsubsection{Rewiring through \upshape$q_P^{\text{new}}$}\label{sec:rewire}

\begin{algorithm}[t]
\caption{Function $\texttt{Rewire}(q_P^{\text{new}},\ccalQ_P^{\text{near}},\ccalT)$}
\label{alg:rewire}
        \For {\upshape $\textit{all}$~$q_P^{\text{near}}=(\bbx_P^{\text{near}},q_B^{\text{near}})\in \ccalQ_P^{\text{near}}$}{\label{rew:line1}
                \If{\upshape $\bbx^{\text{new}} \rightarrow \bbx^{\text{near}} \wedge (q_B^{\text{new}},L(\bbx^{\text{new}}),q_B^{\text{near}})\in\rightarrow_B$}{\label{rew:line2}
                 \If{{\upshape \texttt{Cost}$(q_P^{\text{near}})>\texttt{Cost}(q_P^{\text{new}})+C(\bbx^{\text{new}},\bbx^{\text{near}})$}}{\label{rew:line3}
                  $\texttt{Cost}(q_P^{\text{near}})=\texttt{Cost}(q_P^{\text{new}})+C(\bbx^{\text{new}},\bbx^{\text{near}})$\;\label{rew:line4}
                        $\ccalE_{\ccalT}=\ccalE_{\ccalT}\setminus \{(\texttt{Parent}(q_P^{\text{near}}),q_P^{\text{near}})\}$\;\label{rew:line5}
                        $\ccalE_{\ccalT}=\ccalE_{\ccalT}\cup\{(q_P^{\text{new}},q_P^{\text{near}})\}$\;}}}\label{rew:line6}
\Return {$\ccalT$, \upshape $\texttt{Cost}$}\;
\end{algorithm}

\begin{figure}[t]
  \centering
  \includegraphics[width=0.5\linewidth]{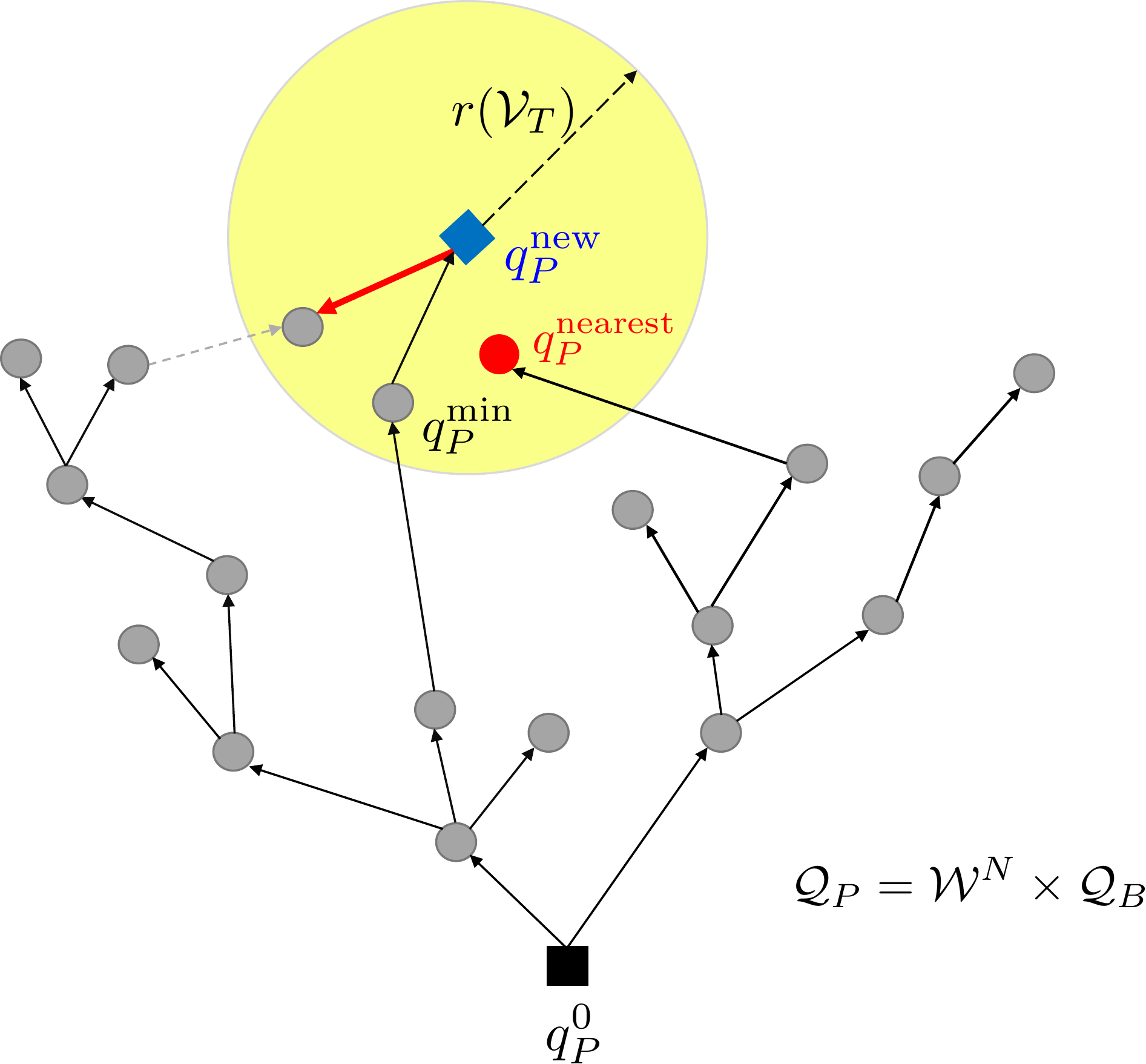}
  \caption{\footnotesize{Graphical depiction of Alg.~\ref{alg:rewire}.  The dashed gray arrow stands for the edge that will be deleted from the set $\ccalE_{\ccalT}$ and the {thick} red arrow stands for the new edge that will be added to $\ccalE_{\ccalT}$ during the execution of Alg.~\ref{alg:rewire}.}}
  \label{fig:rew}
\end{figure}
Once a new state $q_{P}^{\text{new}}=(\bbx^{\text{new}},q_B^{\text{new}})$ has been added to the tree, we \textit{rewire} the nodes in $q_P^{\text{near}}\in\ccalQ_P^{\text{near}}$ through the node $q_{P}^{\text{new}}$ [line~\ref{tree:line17}, Alg.~\ref{alg:tree}] if this will decrease the cost $\texttt{Cost}(q_P^{\text{near}})$, i.e., if this will decrease the cost of reaching the node $q_P^{\text{near}}$ starting from the root $q_P^0$. The rewiring process is described in Alg.~\ref{alg:rewire}; see also  Fig.~\ref{fig:rew}.

Specifically, for each $q_P^{\text{near}}\in\ccalQ_P^{\text{near}}$, we first check if a transition from $q_P^{\text{new}}$ to $q_P^{\text{near}}$ is \textit{feasible}, i.e., if $\bbx^{\text{new}} \rightarrow \bbx^{\text{near}}$ and if $(q_B^{\text{new}},L(\bbx^{\text{new}}),q_B^{\text{near}})\in\rightarrow_B$. If so, then we check if the cost of $q_P^{\text{near}}$ will decrease due to rewiring through $q_P^{\text{new}}$, i.e.,  if $\texttt{Cost}(q_P^{\text{near}})>\texttt{Cost}(q_P^{\text{new}})+C(\bbx^{\text{new}}, \bbx^{\text{near}})$ [line~\ref{rew:line3}, Alg.~\ref{alg:rewire}]. If so, then the cost of $q_P^{\text{near}}$ is updated as $\texttt{Cost}(q_P^{\text{new}})+C(\bbx^{\text{new}},\bbx^{\text{near}})$ [line~\ref{rew:line4}, Alg.~\ref{alg:rewire}],
and the parent of $q_P^{\text{near}}$ becomes the node $q_P^{\text{new}}$ [line~\ref{rew:line5}-\ref{rew:line6}, Alg.~\ref{alg:rewire}], {where the function $\texttt{Parent}:\ccalV_{\ccalT}\rightarrow\ccalV_{\ccalT}$ maps a node to the unique parent node. By convention, we assume that $\texttt{Parent}(q_P^0)=q_P^0$, where $q_P^0$ is the root of the tree $\ccalT$.}

\subsubsection{Computing paths}

The construction of the tree $\ccalT$ ends after $n_{\text{max}}^{\text{pre}}$ iterations, where $n_{\text{max}}^{\text{pre}}$ is user-specified [line~\ref{tree:line5}, Alg.~\ref{alg:tree}]. Then we construct the set $\ccalP=\ccalV_{\ccalT}\cap\ccalQ_{\text{goal}}^{\text{pre}}$ [line~\ref{tree:line15}, Alg.~\ref{alg:tree}] that collects all states $q_P\in\ccalV_{\ccalT}$ that belong to the goal region $\ccalQ_{\text{goal}}^{\text{pre}}$. Given the tree $\ccalT$ and the set $\ccalP$ [line~\ref{alg1:line4}, Alg.~\ref{alg:plans}], we can compute the prefix plans [lines~\ref{alg1:line5}-\ref{alg1:line6}, Alg.~\ref{alg:plans}]. In particular, the path that connects the $a$-th state in the set $\mathcal{P}$, denoted by $\ccalP(a)$, to the root $q_P^0$ constitutes the $a$-th prefix plan and is denoted by $\tau^{\text{pre},a}$ [{function $\texttt{FindPlan}$} in line~\ref{alg1:line6}, Alg.~\ref{alg:plans}]. To compute $\tau^{\text{pre},a}$,  only the parent of each node in  $\ccalT$ is required, due to the tree structure of $\ccalT$. 
Notice that the computational complexity of computing the prefix plan $\tau^{\text{pre},a}$ in the tree $\ccalT$ is $O(|\ccalV_{\ccalT}|)$. On the other hand, if the product state-space $\ccalW^N\times\ccalQ_B$ was searched by a graph $\mathcal{G}=\{\mathcal{V},\mathcal{E}\}$ of arbitrary structure, as in~\cite{karaman2009sampling,karaman2012sampling,vasile2013sampling}, then the computational complexity of the Dijkstra algorithm to find the optimal plan $\tau^{\text{pre},a}$ that connects the state $\mathcal{P}(a)$ to the root would be $O(|\mathcal{E}|+|\mathcal{V}|\log(|\mathcal{V}|))$, where $|\mathcal{E}|+|\mathcal{V}|\log(|\mathcal{V}|)>|\mathcal{V}|$.


\subsection{Construction of Suffix Plans}\label{sec:suffix}

Once the prefix plans $\tau^{\text{pre},a}$ are constructed, the construction of the respective suffix plans $\tau^{\text{suf},a}$ follows [lines~\ref{alg1:line7}-\ref{alg1:line14}, Alg.~\ref{alg:plans}]. The suffix plan $\tau^{\text{suf},a}$ is a sequence of states in $\mathcal{W}_{\text{free}}^N\times\ccalQ_B$ that starts from the state $\mathcal{P}(a)$, $a \in [|\ccalP|]$, and ends at the same state $\mathcal{P}(a)$, i.e., a cycle around state $\mathcal{P}(a)$. Projection of this sequence on $\ccalW_{\text{free}}^N$ gives the suffix plan $\tau^{\text{suf},a}$.

For this purpose, we build a tree $\ccalT=\{\ccalV_{\ccalT}, \ccalE_{\ccalT}, \texttt{Cost}\}$ similarly as in Section~\ref{opt:pre}. The only differences are that: (i) the root of the tree is now $q_P^{0}=\ccalP(a)$, i.e.,  the \textit{accepting} state [line~\ref{alg1:line8}, Alg.~\ref{alg:plans}] detected during the construction of the prefix plans, (ii) the goal region, given the root $q_P^0$, is defined as
\begingroup\makeatletter\def\f@size{10}\check@mathfonts
  \def\maketag@@@#1{\hbox{\m@th\normalsize\normalfont#1}}%
{\begin{align}\label{eq:goalSuf}
\ccalQ_{\text{goal}}^{\text{suf}}& (q_P^0)  =\{q_P=(\bbx,~q_B)\in\ccalW^N_{\text{free}}\times\ccalQ_B~|\nonumber\\&(q_B,L(\bbx),q_P^0|_{\ccalQ_B})\in\rightarrow_B\wedge~\bbx \rightarrow q_P^0|_{\ccalW^N_{\text{free}}}\},
\end{align}}
\endgroup
i.e., it collects all states $q_P$ from which a transition to the root $q_P^0$ is feasible, but this transition will not be included in $\ccalE_{\ccalT}$. Note that before constructing trees to compute suffix plans, we first check if $q_P^0\in\ccalQ_{\text{goal}}^{\text{suf}}$, i.e., {if $(q_P^0|_{\ccalQ_B},L(q_P^0|_{\ccalW_{\text{free}}^N}),q_P^0|_{\ccalQ_B}) \in \rightarrow_B$} [line~\ref{alg1:line9a}, Alg.~\ref{alg:plans}]. If so, the tree $\ccalT$ is trivial, as it consists of only the root, and a loop around it with zero cost [line~\ref{alg1:line9b}, Alg.~\ref{alg:plans}].
If $q_P^0\notin\ccalQ_{\text{goal}}^{\text{suf}}$, then the tree $\ccalT$ is constructed by Alg.~\ref{alg:tree} [line~\ref{alg1:line10}, Alg.~\ref{alg:plans}].

Once a tree rooted at $q_P^0=\ccalP(a)$ is constructed, a set $\ccalS_{a}\subseteq\ccalV_{\ccalT}$ is constructed that collects all states $q_P\in\ccalV_{\ccalT}\cap\ccalQ_{\text{goal}}^{\text{suf}}(q_P^0)$ [lines~\ref{alg1:line11}-\ref{alg1:line12}, Alg.~\ref{alg:plans}].
Then for each state $q_P\in\ccalS_{a}$, there exists a suffix plan, denoted by $\tau^{\text{suf},e}_{a}$, $\forall e\in[|\ccalS_{a}|]$, and  we compute the cost $J(\tau^{\text{suf},e}_{a})$ using $J(\tau^{\text{suf},e}_{a})=\texttt{Cost}(\ccalS_{a}(e))+C(\ccalS_{a}(e)|_{\ccalW^N_{\text{free}}}\,q_P^0|_{\ccalW^N_{\text{free}}})$, where $\ccalS_{a}(e)$ denotes the $e$-th state in $\ccalS_{a}$. Among all detected suffix plans $\tau^{\text{suf},e}_{a}$ associated with the \textit{accepting} state $\mathcal{P}(a)$, we pick the one with the minimum cost, which constitutes the suffix plan $\tau^{\text{suf},a}$ [lines~\ref{alg1:line13}-\ref{alg1:line14}, Alg.~\ref{alg:plans}]. This process is repeated for $a\in[|\ccalP|]$ [line~\ref{alg1:line7}, Alg.~\ref{alg:plans}].

\begin{rem}
{The execution of the suffix paths requires that the robots visit the suffix waypoints infinitely often. However, in practice, it may be hard to visit the exact locations of these waypoints infinitely often due to possible uncertainties or disturbances in the dynamics or environment. We note that our algorithm TL-RRT$^*$ is not sensitive to such disturbances as the robots can visit a neighborhood of each waypoint wherein the same observation is made.}
\end{rem}
\subsection{Construction of the Optimal Discrete Plan}\label{sec:plan}

By construction, any  plan $\tau^{a}=\tau^{\text{pre},a}[\tau^{\text{suf},a}]^{\omega}$, with $\mathcal{S}_{a}\neq\varnothing$, $a\in\{1,\dots,|\mathcal{P}|\}$ satisfies the global LTL specification $\phi$. The cost $J(\tau^{a})$ of each plan $\tau^{a}$  is defined in {\eqref{eq:cost1}. Among} all plans $\tau^{a}\models\phi$, we pick the one with the smallest cost $J(\tau^{a})$ denoted by $\tau=\tau^{a^*}$, where $a^*=\text{argmin}_{a}{J(\tau^{a})}$ [lines~\ref{alg1:line15}-\ref{alg1:line16}, Alg.~\ref{alg:plans}].
\begin{rem}
  {The plan $\tau$ satisfies the LTL formula if all robots move synchronously by reaching next waypoints at the
same time. Future research includes extension of  TL-RRT$^*$ so that the plans can be executed asynchronously  while satisfying the LTL specification, as in~\cite{ulusoy2013optimality}.
  }
\end{rem}
\section{Biased Sampling Method} \label{sec:biased}
In this section,  we propose a biased sampling method that biases the construction of the tree towards shortest paths, {in terms of the number of transitions}, to the final states in the NBA to accelerate the construction of low-cost feasible plans. We call this method biased TL-RRT$^*$. {Intuitively, the biased TL-RRT$^*$ extends more often nodes in the tree  that are closer to accepting states and the new nodes $(\x{new}, \q{B}{new})$ that are added to the tree are such that any subsequent nodes added to the tree via them are even  closer to the accepting states.}

To this end, similar to~\cite{kantaros2018temporal,kantaros2020stylus}, we first prune the NBA by removing infeasible transitions. Specifically, a transition from a state $q_B$ to $q_B'$ is infeasible if it is enabled by a {propositional formula}, e.g., $\pi_i^{\ell_j}  \wedge \pi_i^{\ell_k}$, that requires a robot to be present at more than one disjoint regions simultaneously.

Moreover, we define a distance function $\rho: \Qb \times \Qb \rightarrow \mathbb{N}$ between any two B$\ddot{\text{u}}$chi states in the NBA, which captures the minimum number of transitions in the pruned NBA
\begingroup\makeatletter\def\f@size{10}\check@mathfonts
  \def\maketag@@@#1{\hbox{\m@th\normalsize\normalfont#1}}%
\begin{equation}
\rho(q_B,q_B')=\left\{
                \begin{array}{ll}
                  |SP_{q_B,q_B'}|, \mbox{if $SP_{q_B,q_B'}$ exists,}\\
                  \infty, ~~~~~~~~~\mbox{otherwise},
                \end{array}
              \right.
\end{equation}
\endgroup
where $SP_{q_B,q_B'}$ denotes the shortest path in the pruned NBA from $q_B$ to $q_B'$ and $|SP_{q_B,q_B'}|$ is the number of hops/transitions in this shortest path. Using this metric, we can define a feasible accepting B$\ddot{\text{u}}$chi state $q_B^F \in\ccalQ_B^F$ as (i) $\rho(q_B^0,q_B^{F})\neq \infty$ and (ii) $\rho(q_B^{F},q_B^{F})\neq\infty$. If there are multiple feasible accepting states, we randomly select one, denoted by $q_B^{F,\text{feas}}$, and use it throughout Alg.~\ref{alg:tree}. Moreover, given the tree $\ccalT$, we define the set $\ccalD_{\text{min}} \subseteq \ccalV_{\ccalT} $ that collects the nodes $q_P\in\ccalV_{\ccalT}$ which have the minimum distance $\rho(q_B,q_B^{F,\text{feas}})$. {Intuitively, these nodes are the ``closest'' to the accepting states that have B$\ddot{\text{u}}$chi state $q_B^{F,\text{feas}}$.} More details can be found in~\cite{kantaros2018temporal,kantaros2020stylus}. {Alternatively, a distance metric over automata can also be defined in terms of the number of sets of atomic propositions that still need to be satisfied until reaching an accepting state~\cite{lacerda2019probabilistic}. Note that any distance metric over automata can be used with our biased sampling method.}
\begin{algorithm}[t]
  \caption{Changes made to Alg.~\ref{alg:tree}}
  \label{alg:biased}
\SetAlgoLined
  \While{\upshape $\x{new} \not\in \ccalW_{\text{free}}^N$}{\label{biastree:line5.6}
      $[\x{rand}, \q{P}{closest}] = \texttt{BiasedSample}(\ccalW, \ccalT)$\; \label{biastree:line4}
$\bbx^{\text{new}}=\texttt{Steer}(\bbx^{\text{closest}},\bbx^{\text{rand}})$\;\label{biastree:line5}
  }
  $\ccalQ_P^{\text{near}}=\texttt{Near}(\x{new}, \ccalT) \cup \{q_P^{\text{closest}}\}$\; \label{biastree:line6}
\end{algorithm}

\subsection{Construction of Prefix Plans}\label{bias:pre}
The biased sampling-based algorithm for the prefix plans is similar to Alg.~\ref{alg:tree} except that the selection of $\x{rand}$ and the construction of the set $\ccalQ_P^{\text{near}}$ [lines~\ref{tree:line5.6}-\ref{tree:line8.5}, Alg.~\ref{alg:tree}] are now determined by Alg.~\ref{alg:biased}. To sample a state $\x{rand}$, we employ the function $\texttt{BiasedSample}$ in Alg.~\ref{alg:sample1}.

\subsubsection{Selection of \upshape{$\q{P}{closest} = (\x{closest}, \q{B}{closest})\in \ccalV_{\ccalT}$} [line~\ref{s:line2}, Alg.~\ref{alg:sample1}]}\label{pa:closest} {We select a node $\q{P}{closest}$ from the tree to expand. {Specifically, selection of $\q{P}{closest}$ is biased towards nodes that are closer to the accepting states, i.e., nodes in the set $\ccalD_{\text{min}}$.}} First, we sort {the nodes  in the sets  $\ccalD_{\text{min}}$ and $\ccalV_{\ccalT} \setminus\ccalD_{\text{min}}$}  in the opposite order that they were added to the tree. Then the point $\q{P}{closest}$ is sampled from a discrete probability function $f_{\text{closest}}(q_P|\ccalV_{\ccalT}, \ccalD_{\text{min}}):\ccalV_{\ccalT}\rightarrow[0,1]$,  defined as:
\begingroup\makeatletter\def\f@size{10}\check@mathfonts
  \def\maketag@@@#1{\hbox{\m@th\normalsize\normalfont#1}}%
\begin{align}\label{eq:fcl}
  & f_{\text{closest}}(q_P = i\,|\,\ccalV_{\ccalT},\ccalD_{\text{min}})   \\ \nonumber
  &= \left\{
                \begin{array}{ll}
                  p_{\text{closest}} \mathbb{P}_{\text{UG}}(q_P = i; \frac{1}{|\ccalD_{\text{min}}|}), &\mbox{if}~i\in\ccalD_{\text{min}}\\
                  (1-p_{\text{closest}}) \mathbb{P}_{\text{UG}}(q_P = i; \frac{1}{|\ccalV_{\ccalT} \setminus\ccalD_{\text{min}}|}), &\mbox{if}~i\in\ccalV_\ccalT\setminus \ccalD_{\text{min}}.
                \end{array}
              \right.
\end{align}
\endgroup
where $p_{\text{closest}} \in (0.5, 1)$ is a user-specified parameter denoting the probability of selecting any node in $\ccalD_{\text{min}}$ to be $q_P^{\text{closest}}$. $\mathbb{P}_{\text{UG}}$ denotes the UG distribution~\cite{akdougan2016uniform}, which compounds the uniform and geometric distributions. Given a countably infinite set $\mathbb{N}^+$ and a parameter $p$ such that $0<p<1$, the probability mass function of a UG distribution is defined as
\begingroup\makeatletter\def\f@size{10}\check@mathfonts
  \def\maketag@@@#1{\hbox{\m@th\normalsize\normalfont#1}}%
\begin{align}\label{equ:ug}
  \mathbb{P}_{\text{UG}}(q_P = i;p) = \sum_{n=i}^{\infty} \frac{1}{n} p (1-p)^{n-1},\quad i\in\mathbb{N}^+.
\end{align}
\endgroup

{To apply the UG distribution to $\ccalD_{\text{min}}$, we compute the probability for $q_P \!\!\in\!\! [|\ccalD_{\text{min}}|]$ and then normalize so that the probabilities add to 1.} The UG distribution has the property:
\begingroup\makeatletter\def\f@size{10}\check@mathfonts
  \def\maketag@@@#1{\hbox{\m@th\normalsize\normalfont#1}}%
\begin{align}
  \mathbb{P}_{\text{UG}}(q_P = i+1) = \mathbb{P}_{\text{UG}}(q_P=i)  - \frac{p(1-p)^{i-1}}{i},
\end{align}
\endgroup
where $\mathbb{P}_{\text{UG}}(q_P=1) = -\frac{p\log{p}}{1-p}$. We have $\mathbb{P}_{\text{UG}}(q_P=i+1) < \mathbb{P}_{\text{UG}}(q_P=i)$ and $\mathbb{P}_{\text{UG}}(q_P=i+1)$ approaches $\mathbb{P}_{\text{UG}}(q_P=i)$, $\forall i\in\mathbb{N}^+$, as $p$ goes to 0. Thus, the UG distribution in~\eqref{equ:ug} approximates a uniform distribution in the limit.  Since the UG distribution is unimodal with a mode of 1, sorting the nodes in the opposite order makes the probability that a newly-added node is selected as $\q{P}{closest}$ slightly larger compared to nodes that were added before, accelerating the expansion of the tree. {{The reason we prefer UG distribution to the uniform distribution is that the latter is ill-defined over an infinite set, which occurs when the node size goes to infinity.}} {Note that mathematically, the UG distribution is defined over a countably infinite set. To sample from a finite node set of the tree with size $n$, we first compute the cumulative probability function of the event $\{i\leq n\}$ assuming the node set is infinite, then we renormalize, so that the probability that a node is selected as $\q{P}{closest}$ is proportional to that when the node set is infinite.}

\subsubsection{Selection of $q_B^{\textup{succ,1}}, q_B^{\textup{succ,2}} \in \ccalQ_B$}\label{pa:succ} {Next, given the node $\q{P}{closest} = (\x{closest}, \q{B}{closest})$, we select two other succesive B$\text{\"{u}}$chi states $\q{B}{succ,1}$ and $\q{B}{succ,2}$ in the prune NBA, such that $\q{B}{closest}\xrightarrow{L(\x{closest})}_B \q{B}{succ,1} \to_B \q{B}{succ,2} $ and this two-hop transition proceeds towards the accepting state $q_B^{F,\text{feas}}$. That is, $\q{B}{succ,1}$ is closer to $q_B^{F,\text{feas}}$ than  $\q{B}{closest}$ and $\q{B}{succ,2}$ is closer than $\q{B}{succ,1}$ in terms of the distance function $\rho$. {If the new state  $\q{P}{new}$ is added to the tree through the selected node $\q{P}{closest}$, the first hop implies that  the B$\text{\"{u}}$chi state $\q{B}{succ,1}$ in $\q{P}{new}$ can be  $\q{B}{new}$  no matter what $\x{new}$ is.} The resulting $\q{P}{new}$ is more likely to be selected as $\q{P}{closest}$ in the next iterations. If so, the selection of $\x{new}$ should enable the second hop, i.e., $\q{B}{succ,1}\xrightarrow{L(\x{new})} \q{B}{succ,2}$, so that another product state including $\q{B}{succ,2}$ can be added to the tree, growing the tree further towards the accepting state $q_B^{F,\text{feas}}$. We refer the reader to~\cite{kantaros2020stylus} for more details.}

\begin{figure}[]
  \begin{center}
    \includegraphics[width=0.95\linewidth]{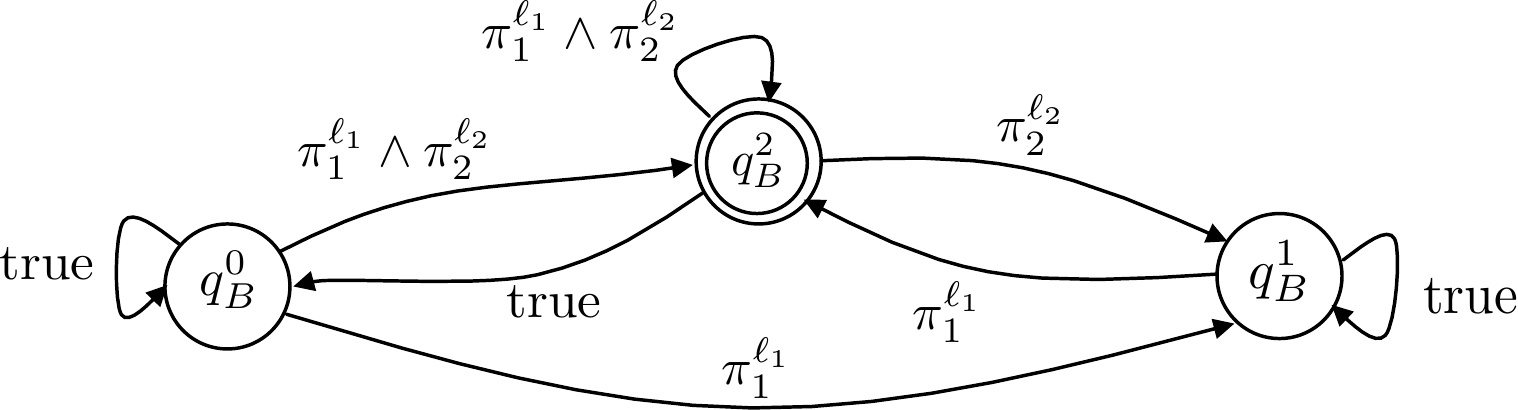}
  \end{center}
  \caption{{The NBA for the specification $\square \Diamond \pi_{1}^{\ell_1} \wedge \square \Diamond \pi_{2}^{\ell_2}$.}}
  \label{fig:example}
\end{figure}

\begin{algorithm}[t]
\caption{Function $\texttt{BiasedSample}(\ccalW, \ccalT)$}
\label{alg:sample1}
Sample $q_{P}^{\text{closest}}=(\x{closest},q_B^{\text{closest}})\in\ccalV_{\ccalT}$ by~\eqref{eq:fcl}\;\label{s:line2}
Sample $q_B^{\text{succ,1 }}, \q{B}{succ,2} \in q_B^{\textup{closest}}$\;\label{s:qBmin}
\If{$q_B^{\textup{succ,1}}$ \textup{and} $q_B^{\textup{succ,2}}$ \textup{exist}}
   {
     Select $\sigma^*$ enabling $\q{B}{succ,1} \to_B \q{B}{succ,2}$\;\label{s:sigmamin}
     Sample $\bbx^{\text{rand}}=[\bbx_{1}^{\text{rand},T},\dots,\bbx_{N}^{\text{rand},T}]^T$ by~\eqref{eq:fnew}\; \label{s:line5}
   }
   \Else{\Return{\upshape $\x{rand}=\varnothing, \q{P}{closest}=\varnothing$}\;}\label{s:empty1}

   \Return{\upshape$\x{rand},  \q{P}{closest}$}\;
\end{algorithm}
{\begin{example}
    Consider the specification $\square \Diamond \pi_{1}^{\ell_1} \wedge \square \Diamond \pi_{2}^{\ell_2}$, whose NBA is shown in Fig.~\ref{fig:example}. When finding the prefix plan, we have $q_B^2 = q_B^{F,\text{feas}}$ and the distance  $\rho(q_B^0, q_B^2)=1, \rho(q_B^1, q_B^2)=1, \rho(q_B^2, q_B^2)=0$. Initially, the tree $\ccalT$ only contains  the root $q_P^0=(\bbx_0, q_B^0)$. Assume also that the atomic propositions $\pi_1^{\ell_1}, \pi_2^{\ell_2} \not\in L(\bbx_0)$. {According to steps~\hyperref[pa:closest]{1)} and~\hyperref[pa:succ]{2)}, a possible case of selected vertices is} $\q{B}{closest}=q_B^0$ (because the tree only has the node $(\x{0}, q_B^0)$), $\q{B}{succ,1}=q_B^0$ (because $q_B^0 \xrightarrow{\text{true}} q_B^0$ and $q_B^0 \not\xrightarrow{L(\bbx_0)}q_B^1, q_B^2$) and $\q{B}{succ,2}=q_B^2$ (because $q_B^0 \to q_B^2$ and $\rho(q_B^2, q_B^2) < \rho(q_B^0, q_B^2)$). That is, when the NBA is at  $q_B^0$, it should more frequently follow  the shortest path $q_B^0, q_B^2$.
\end{example}}
\subsubsection{Sampling of $\x{rand} \in \ccalW$}\label{pa:rand} 
\begin{figure}[t]
  \centering
  \includegraphics[width=0.5\linewidth]{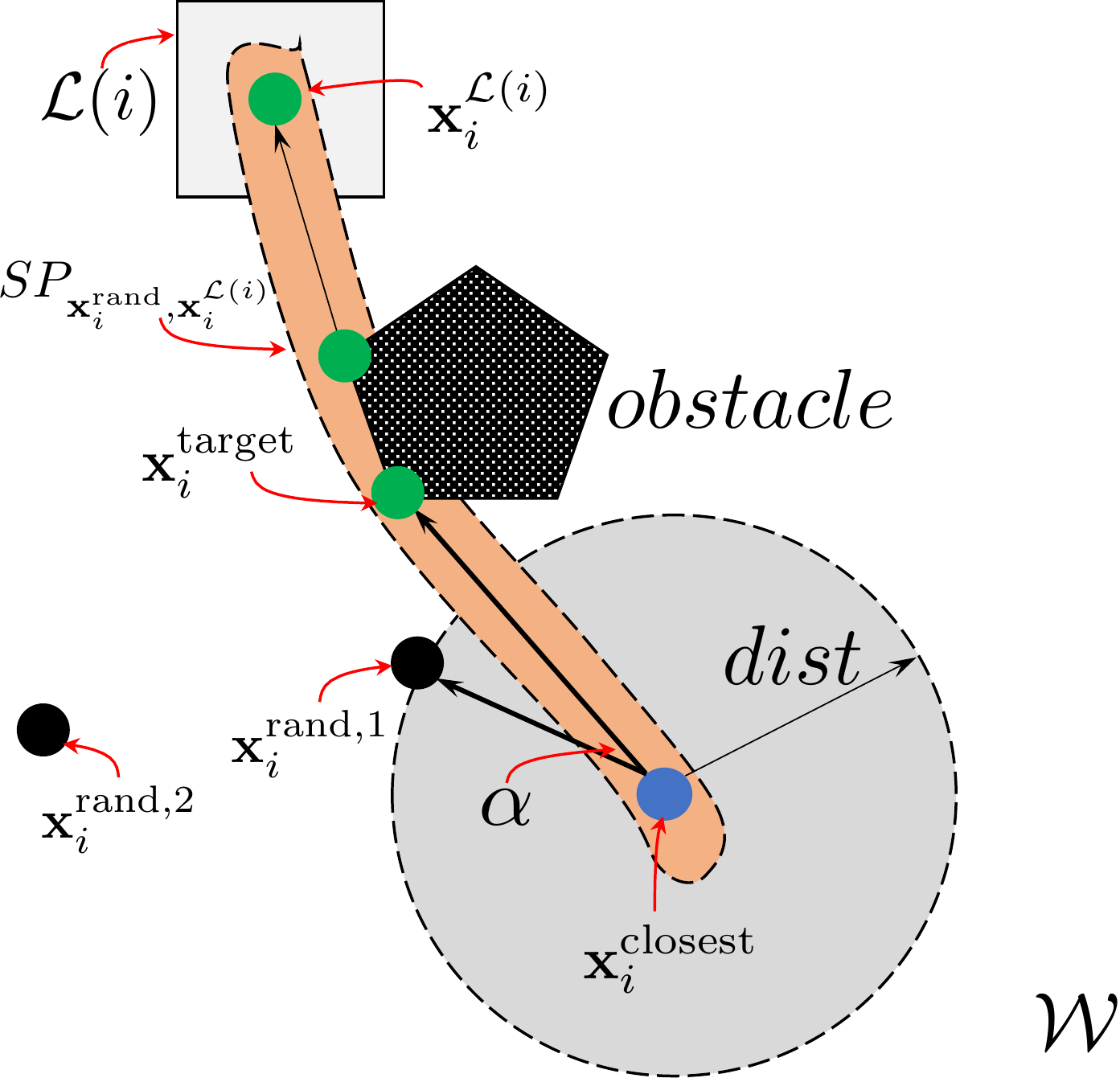}
  \caption{{Graphical depiction of the proposed biased sampling for robot $i$. 
        Given $\q{B}{succ,1}$ and $\q{B}{succ,2}$, the symbol ${\sigma^*}$ that enables this transition is used to sample $\x{rand}$. $\ccalL(i)$ is the labeled region where robot $i$ should be to satisfy ${\sigma^*}$. The path surrounded by light orange area is the shortest path from $\bbx_i^{\text{closest}}$ to $\bbx_i^{\mathcal{L}(i)}$ generated by visibility graph. $\bbx_i^{\text{rand},1}$ is the point generated by a normal distribution, with distance $d$ from $\bbx_i^{\text{closest}}$ and angle $\alpha$ between two thick arrow lines. $\bbx_i^{\text{rand},2}$ is the point generated by a uniform distribution.}}
  \label{fig:sampling}
\end{figure}

Given states $\q{B}{succ,1}$ and $\q{B}{succ,2}$, we select a {propositional formula} ${\sigma^*}$ such that $\q{B}{succ,1} \xrightarrow{{\sigma^*}}_B \q{B}{succ,2}$ in the NBA [line~\ref{s:sigmamin}, Alg.~\ref{alg:sample1}]; see Fig.~\ref{fig:sampling}. {For this, we convert the condition enabling the transition from $\q{B}{succ,1}$ to $\q{B}{succ,2}$ in the disjunctive normal form {of $\bigvee \bigwedge \pi_i^{\ell_j}$,} that is, $\q{B}{succ,1} \xrightarrow{\bigvee \bigwedge \pi_i^{\ell_j}} \q{B}{succ,2}$,  and we select one clause {$\bigwedge \pi_i^{\ell_j}$} randomly or the one with the minimum length to be ${\sigma^*}$.} Next, we construct a set $\ccalL$ whose $i$-th element $\ccalL(i)$ denotes the labeled region that robot $i$ should visit {according to the symbol $\pi_{i}^{\ell_j}$ in ${\sigma^*}$.  Note that $\ccalL(i)$ may not appear in ${\sigma^*}$.} If $\ccalL(i)$ is an empty symbol, denoted by $\epsilon^*$, then robot $i$ can be anywhere. In this case, with probability $p_{\text{idle}}$, we allow robot $i$ to stay at its current location to incur zero cost; otherwise, robot $i$ moves to any other reachable location. {If $ \pi_{i}^{\ell_j}$ appears in ${\sigma^*}$, then with high probability we draw one sample from $\ccalW_{\text{free}}$ that is closer to the location $\bbx_i^{\ccalL(i)}$ than $\bbx_{i}^{\text{closest}}$ is, {where $\bbx_{i}^{\ccalL(i)}$ is a point inside region $\ccalL(i)=\ell_j$, such as its centroid.}}
{For polygonal environments  we employ the geodesic path defined as the shortest path between two points, that is, a sequence of line segments that connect two points and pass through reflex  vertices of the polygonal boundary~\cite{kantaros2016global}.} In order for robot $i$ to reach $\bbx_{i}^{\ccalL(i)}$ fast, it should head towards the second vertex in this shortest path, denoted by $\bbx_i^{\text{target}} = SP_{\bbx_{i}^{\text{closest}},\bbx_{i}^{\ccalL(i)}}(2)$, where $SP_{\bbx_{i}^{\text{closest}},\bbx_{i}^{\ccalL(i)}}$ denotes the shortest geodesic path from $\bbx_{i}^{\text{closest}}$ to $\bbx_{i}^{\ccalL(i)}$; see Fig.~\ref{fig:sampling}. Given $\bbx_{i}^{\text{target}}$,  we select $\bbx_i^{\text{rand}}$ as
\begin{align}\label{eq:fnew}
\bbx_i^{\text{rand}} & = \mathbbm{1}_{[Y\leq y_{\text{rand}}]} ~\bbx_i^{\text{rand},1} + \mathbbm{1}_{[Y>y_{\text{rand}}]} ~\bbx_i^{\text{rand},2},
\end{align}
where $Y \in [0,1]$ is a random variable drawn from a uniform distribution, $y_{\text{rand}} \in (0.5, 1)$ is a weighting factor, $\mathbbm{1}_{[Y\leq y_{\text{rand}}]}$ is an indicator variable which is 1 if the event $\{Y\leq y_{\text{rand}}\}$ occurs, otherwise 0. Also in~\eqref{eq:fnew}, $\bbx_i^{\text{rand},1}$ is a point following a normal distribution centered at $\bbx_i^{\text{target}}$ and $\bbx_i^{\text{rand},2}$ is a point following a uniform distribution that is bounded away from zero on $\ccalW_{\text{free}}$. {The fact that $y_{\text{rand}}$ is greater than 0.5 ensures that $\bbx_i^{\text{rand}}$ is closer to $\bbx_i^{\text{target}}$, i.e., robot $i$ moves to $\bbx_i^{\text{target}}$ with high probability.} Specifically, the relative position of $\bbx_i^{\text{rand},1}$ with respect to $\bbx_i^{\text{closest}}$ can be determined by two parameters in the 2D workspace.\footnote{If the dimension of the workspace is 3, we need three parameters, including one distance parameter and two angle parameters.} {One is the distance $dist$ between $\bbx_i^{\text{rand},1}$ and $\bbx_i^{\text{closest}}$, and the other one is the angle $\alpha$ formed by two line segments connecting $\bbx_i^{\text{closest}}$ with $\bbx_i^{\text{rand},1}$ and $\bbx_i^{\text{target}}$, respectively; see also Fig.~\ref{fig:sampling}.}  We use a 2-dimensional normal distribution to sample $dist$ and $\alpha$, with mean $\mu_d$, $\mu_{\alpha}$ and standard deviation $\sigma_d$, $\sigma_{\alpha}$, i.e.,
\begin{align*}
  f(dist,\alpha) = \frac{1}{2\pi\sigma_{dist}\sigma_{\alpha}}\exp{\Big( -\frac{1}{2} \Big[ \frac{dist^2}{\sigma_{dist}^2} +\frac{\alpha^2}{\sigma_{\alpha}^2} \Big] \Big)}.
\end{align*}
Since the distance is non-negative, we use the absolute value $|dist|$ and $\alpha$ to obtain $\bbx_i^{\text{rand},1}$. To obtain $\bbx_i^{\text{rand},2}$, we draw a uniform sample from $\ccalW_{\text{free}}$.  If the dimension of the workspace is 3, we need three parameters, including one distance parameter and two angle parameters.
After obtaining $\bbx_i^{\text{rand}}$ by~\eqref{eq:fnew},  we construct $\bbx^{\text{rand}}=[\bbx_{1}^{\text{rand},T},\dots,\bbx_{N}^{\text{rand},T}]^T$ [line~\ref{s:line5}, Alg.~\ref{alg:sample1}].

Finally, given $\x{rand}$ and $\q{P}{closest}$ returned by Alg.~\ref{alg:sample1}, we generate $\x{new}$ using the $\texttt{Steer}$ function that returns a location that is closer to $\x{rand}$ than $\x{closest}$ is. Then we add $\q{P}{closest}$ to set $\ccalQ_{P}^{\text{near}}$ [lines~\ref{biastree:line5}-\ref{biastree:line6}, Alg.~\ref{alg:biased}]. Next, we construct states $\q{P}{new}$ by pairing a B$\ddot{\text{u}}$chi state $q_B$ to $\x{new}$, exactly as in Alg.~\ref{alg:tree}.
\subsection{Construction of Suffix Plans}
The algorithm to design  the suffix part differs from the one proposed for the biased prefix part in Section~\ref{bias:pre} only in that a cycle around the root needs to be computed. Specifically, once a node $\q{P}{new}$ is constructed, we check whether its B$\ddot{\text{u}}$chi component is the same as that of the root, i.e., whether it is the same  accepting B$\ddot{\text{u}}$chi state; if so, we store this node in a set $\ccalP$, otherwise the tree is built exactly as for the biased prefix part. Together with the construction of the tree, for each node $\q{P}{new}=(\x{new}, \q{B}{new})$ in $\ccalP$, we find a path from $\x{new}=[\bbx^{\text{new},T}_i,\ldots,\bbx^{\text{new},T}_N]^T$ to $\x{0}$ that forms a cycle. To find this path, we apply RRT$^*$~\cite{karaman2011sampling} to find a path for each robot $i \in [N]$, that connects $\x{new}_i$ to $\x{0}_i$ while treating all labeled regions as obstacles. This ensures that during the execution of the plan no other observation will be generated and, therefore, the robots will maintain the desired accepting B$\ddot{\text{u}}$chi state.

{
  \begin{rem}
 {While in this paper we focus on the multirobot planning problem, the proposed method, similar to RRT$^*$, can address optimal control problems  subject to temporal logic specifications too, such as  manipulation tasks~\cite{li2019formal} or estimation tasks~\cite{khodayi2019distributed}.} For this, it is important that infeasible transitions in the NBA can be identified and pruned since, otherwise, it is possible that TL-RRT{$^*$} can bias search towards those infeasible transitions, which can significantly deteriorate performance. In motion planning applications, it is simple to identify such infeasible transitions as discussed in Section~\ref{sec:biased}; see also~\cite{kantaros2018temporal,kantaros2020stylus}. However, in general, identifying infeasible NBA transitions may be problem specific  and not easy to do.
  \end{rem}
}

\section{Correctness and Optimality} \label{sec:corr}
In this section we show that  TL-RRT$^*$ is probabilistically complete and asymptotically optimal. Note that TL-RRT$^*$ does not trivially inherit the completeness and optimality properties of RRT$^*$, since TL-RRT$^*$ explores a combined continuous and discrete state space while RRT$^*$ is designed to explore only continuous state spaces. The resulting technical differences with RRT$^*$ are discussed in the proof of TL-RRT$^*$ in~\hyperref[app:comp]{Appendices}. First, we make the following assumptions.
\begin{asmp}[Nonpoint regions]
  \label{asmp:nonr}
  Every atomic proposition in the LTL formula $\phi$ is satisfied over a nonpoint region. More precisely, $\mu(\ell_j) >0$ where $\mu$ is the Lebesgue measure.
\end{asmp}

Assumption~\ref{asmp:nonr} ensures that any point within a labeled region can be sampled with nonzero probability. Otherwise, it is impossible to generate a feasible plan. In what follows, we denote by $\ccalB_{r}((\bbx, q_B))$ a ball of radius of $r$ centered at $(\bbx, q_B)$ in the product state space,
\begin{align*}
  \ccalB_{r}((\bbx, q_B)) = \{ (\bbx', q'_B) \in \ccalQ_P \,|\, dist((\bbx, q_B),(\bbx', q'_B)) \leq r \},
\end{align*}
where, with slight notational abuse, $ dist((\bbx, q_B),(\bbx', q'_B)) = \| \bbx -\bbx'\|$ is the distance between two product states. In words, a product state $(\bbx',q'_B) \in \ccalQ_P$ lies in the ball $\ccalB_r((\bbx, q_B))$ if  $\bbx'$ is at distance less than $r$ from $\bbx$. We denote by $\texttt{int}(\ccalB_{r}((\bbx, q_B)))$  the interior of the ball $\ccalB_{r}((\bbx, q_B))$. By definition of the distance function $dist$, a point $\bbx'$ lies in the ball $\ccalB_r((\bbx, \cdot))$ regardless of its B\text{\"{u}}chi state component. This definition of a ball is necessary since the product space consists of the continuous space $\ccalW_{\text{free}}^N$ and the discrete space $\ccalQ_B$, and there is not a physical notion of distance in $\ccalQ_B$.

\begin{asmp}[Convergence space]
  \label{asmp:nhb}
Let Assumption~\ref{asmp:nonr} hold. For any reachable product state $(\vect{x}, \qb) \in \wf^N \times \Qb$ from the root, there exists a constant $\delta_{\vect{x}} >0$ that depends on $\vect{x}$, such that any point  $\bbx'$ in $\ccalW_{\text{free}}^N$ for which  $(\bbx', \cdot)$ lies in the interior of the ball $\mathcal{B}_{\delta_{\vect{x}}}((\vect{x}, q_B))$, can be paired with the same B$\ddot{\text{u}}$chi state $\qb$ as the  center $\bbx$. Therefore, the product state $(\bbx', q_B)$  can also be reached from the root.
\end{asmp}

Assumption~\ref{asmp:nhb} ensures that a homotopy class exists around any feasible path such that any path in this class can be continuously deformed into another. 

\subsection{Probabilistic Completeness}
In this section, we show the probabilistic completeness of TL-RRT$^*$ by skipping $\texttt{Extend}$ and $\texttt{Rewire}$, and instead connecting $\q{P}{new}$ only with nodes in $\ccalQ_P^{\text{nearest}}$, similar to RRT~\cite{kuffner2000rrt}. {The reason is that, if there exists a candidate parent  $\q{P}{nearest}$ of $\q{P}{new}$ in $\ccalQ_P^{\text{nearest}}$, then $\q{P}{new}$ will be added to the tree $\ccalT$ regardless of which node in $\ccalV_{\ccalT}$ is selected by $\texttt{Extend}$  to be its parent. {Furthermore, $\texttt{Rewire}$ updates the set of edges of the tree and does not play any role in adding new nodes.} Therefore, it can not affect the completeness property. Thus, we focus on finding a candidate parent in $\ccalQ_P^{\text{nearest}}$. If connecting $\q{P}{new}$ with nodes in $\ccalQ_P^{\text{nearest}}$ can ensure probabilistic completeness, based on the fact that $\ccalQ_P^{\text{nearest}}$ is a subset of  $\ccalQ_{P}^{\text{near}}$ among which TL-RRT$^*$ attempts to find the parent, the probabilistic completeness of TL-RRT$^*$ follows directly. Note that when the number of nodes goes to infinity, the distance between the sampled point and the nearest nodes is much smaller than $\eta$. {Therefore, $\x{new}$ will be identical to $\x{rand}$. This argument can be found in Lemma 1 in~\cite{solovey2020revisiting}. Our proofs are based upon the observation that $\x{new}=\x{rand}$.}  The following theorem shows that unbiased TL-RRT* is probabilistically complete.

\begin{thm}[Probabilistic Completeness of TL-RRT$^*$ with Unbiased Sampling]\label{prop:compl}
Let Assumptions~\ref{asmp:nonr} and~\ref{asmp:nhb} hold and further assume that sampling in the free workspace is unbiased. Then,  TL-RRT$^*$ is probabilistically complete, i.e., if there exists a feasible plan that satisfies a given LTL formula $\phi$, then TL-RRT$^*$ will find it with probability 1.
\end{thm}
{\begin{IEEEproof}
    We provide a proof sketch and the details can be found in Appendix~\ref{app:comp}. The proof proceeds in two steps. First, given any product state in $\ccalQ_P$ that is one-hop-reachable from the root, we prove that the tree  will have a node that is arbitrarily close to it in terms of Euclidean distance, and  based on Assumption~\ref{asmp:nhb}, this node will have the same B$\ddot{\text{u}}$chi state as the given product state. The key idea is to show that {the expected distance between the tree and the given product state converges to 0 as the tree grows}.  Next, we extend the above result to multi-hop reachable states from the root. Therefore, given an accepting state that is reachable from the root,  a node that is arbitrarily close to it and has the same accepting B$\ddot{\text{u}}$chi state will be added to the tree with probability 1 as the number of iterations of Alg~1 goes to infinity.
  \end{IEEEproof}
}

Theorem~\ref{prop:compl} can also be extended for the case of the biased sampling introduced in Section~\ref{sec:biased}. Specifically, we have the following result, whose proof can be found in Appendix~\ref{app:biased}.
\begin{cor}[Probabilistic Completeness of TL-RRT$^*$ with Biased Sampling]\label{prop:biased}
Let Assumptions~\ref{asmp:nonr} and~\ref{asmp:nhb} hold. Then, the biased TL-RRT$^*$ is probabilistically complete, i.e., if there exists a feasible plan that satisfies a given LTL formula $\phi$, then the biased TL-RRT$^*$ will find it with probability 1.
\end{cor}

\subsection{Asymptotic Optimality}\label{sec:opt}
  In this section, we show the asymptotic optimality of TL-RRT$^*$. We first define a product plan $p$ given a discrete plan $\tau=\tau(1),\ldots, \tau(k), \ldots$ satisfying $\phi$. Taking $\texttt{trace}(\tau)$ as the input to the  NBA, a run $q_B^1, \ldots, q_B^k, \ldots $ will be produced. Given the one-to-one correspondence between states in ${\tau}$ and states in this run, we can construct a product state plan $p$ by pairing each position component with a B$\ddot{\text{u}}$chi state, i.e., $p = (\tau(1),q_B^1), \ldots, (\tau(2), q_B^2), \ldots$. In this case, $\tau$ is the projection of $p$ onto $\ccalW_{\text{free}}^N$. Moreover, let $\tau^*=\tau^{*,\text{pre}}[\tau^{*,\text{suf}}]^\omega$ be the optimal plan that satisfies $\phi$ and incurs the optimal cost defined in~\eqref{eq:cost1}. We use $\tau^{*,\text{pre}}|\tau^{*,\text{suf}}$ to represent $\tau^*$ since it suffices to characterize  the optimal plan.

{\begin{thm}[Asymptotic Optimality of TL-RRT$^*$ with Unbiased Sampling]\label{prop:opt}
  Let Assumptions~\ref{asmp:nonr} and~\ref{asmp:nhb} hold and further assume that sampling in the free workspace is unbiased. Consider also the parameter $r_n(\ccalV_{\ccalT})$ defined in \eqref{eq:r}. Then, TL-RRT$^*$ is asymptotically optimal, i.e., the discrete plan $\tau_{n_{\text{max}}^{\text{pre}}}^{n_{\text{max}}^{\text{suf}}}$ that is generated by this algorithm satisfies
\begin{equation}\label{eq:opt1}
  \lim_{{n_{\text{max}}^{\text{pre}}\to\infty , n_{\text{max}}^{\text{suf}}\to\infty } } \mathbb{P}\left(\left\{J(\tau_{n_{\text{max}}^{\text{pre}}}^{n_{\text{max}}^{\text{suf}}}) \leq (1 + \epsilon) J(\tau^*)\right\}\right)=1.
\end{equation}
where $\epsilon\!\in\!(0,1)$, $n_{\text{max}}^{\text{pre}}$ and $n_{\text{max}}^{\text{suf}}$ are the maximum numbers of iterations used in Alg.~\ref{alg:tree},  $\tau_{n_{\text{max}}^{\text{pre}}}^{n_{\text{max}}^{\text{suf}}}=\tau^{\text{pre},n_{\text{max}}^{\text{pre}}}|\tau^{\text{suf},n_{\text{max}}^{\text{suf}}}$, $\tau^* = \tau^{*,\text{pre}}|\tau^{*,\text{suf}}$ and $J$ is the cost function defined in \eqref{eq:cost1}.
\end{thm}}
\begin{figure}[t]
  \centering
  \includegraphics[width=0.65\linewidth]{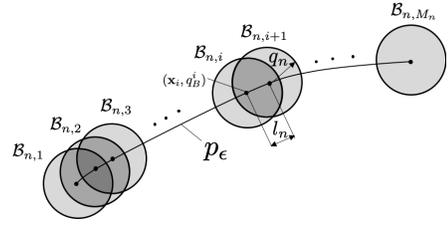}
    \caption{An illustration of the construction of balls covering the whole  product path $p_{\epsilon}$, modified from \cite{karaman2011sampling}. All balls have the same radius $q_n$. The spacing between  centers of two consecutive balls is $l_n$; see Appendix~\ref{sec:ball} for details.}
  \label{fig:balls}
\end{figure}
{\begin{IEEEproof}
    We provide a proof sketch based on~\cite{solovey2020revisiting} and the details can be found in Appendix~\ref{app:opt}.  
    { For any $\delta\in \mathbb{R}^+$, the $\delta$-interior of the space $\ccalW^N_{\text{free}}$, denoted by $\texttt{int}_{\delta}(\mathcal{W}^N_{\text{free}})$, is a subset of $\mathcal{W}^N_{\text{free}}$ containing states that are at least $\delta$ distance away from any obstacle. Then, we call a feasible path ${\tau}$ {\it robust} if there exists $\delta\in\mathbb{R}^+$ such that ${\tau} \subset \texttt{int}_{\delta}(\mathcal{W}^N_{\text{free}})$. By Assumption~\ref{asmp:nhb},  Problem~\ref{pr:problem} is robustly feasible. Following~\cite{solovey2020revisiting}, there exists a path $\tau_{\epsilon}$ such that $J(\tau_{\epsilon}) \leq (1 + \epsilon/4) J(\tau^*)$, where $\epsilon\in(0,1)$.} We  show that the cost of the plan returned by TL-RRT$^*$ is at most $(1 + \epsilon) J(\tau^*)$ when the number of iterations goes to infinity. To this end, we augment the path $\tau_{\epsilon}$ using B$\ddot{\text{u}}$chi states to obtain a product path $p_{\epsilon}$. Then,  at each iteration, we construct a sequence of equally-spaced balls of same radii centered along the path $p_{\epsilon}$ so that the sequence of balls covers it; see Fig.~\ref{fig:balls}. {Furthermore, the radii of these balls are proportional to the connection radius $r_n(\ccalV_\ccalT)$, which converges to 0 as the iteration $n$ grows, so the number of balls grows to infinity.} Consider the event that each ball contains a sample and, for any two adjacent balls, the sample inside the second ball is sampled after that inside the first ball. {Furthermore, assume that if a  ball is intersected  by a boundary of any region, this boundary can be locally approximated by a hyperplane when the ball is extremely small. Then, we require that samples should lie within those parts of the balls that contain their centers. This is because   the transition relation between two points requires that the straight line connecting them crosses any boundary at most once.} When this event occurs, TL-RRT$^*$ will connect together this sequence of samples, augmented by  B$\ddot{\text{u}}$chi states, in an ascending order.  {In this way, the tree contains  a path that arbitrarily approximates the path $p_{\epsilon}$.} We  prove that, when the connection radius $r_n(\ccalV_\ccalT)$ satisfies~\eqref{eq:r}, the probability of the above event converges to 1. Finally, the approximation of $p_{\epsilon}$   induces a path whose cost is no more than $(1 + \epsilon) J(\tau^*)$.
\end{IEEEproof}
}

{The next result shows the optimality of the biased TL-RRT$^*$ algorithm. The  proof can be found in Appendix~\ref{app:biasedopt}. }
  \begin{cor}[Asymptotic Optimality of TL-RRT$^*$ with Biased Sampling]\label{prop:opt1}
    Let Assumptions~\ref{asmp:nonr} and~\ref{asmp:nhb} hold. Assume also that the parameter $\gamma_{\text{TL-RRT}^*}$ in \eqref{eq:r} satisfies
  \begingroup\makeatletter\def\f@size{9}\check@mathfonts
  \def\maketag@@@#1{\hbox{\m@th\normalsize\normalfont#1}}%
  \begin{align}\label{gamma_biased}
  \gamma_{\text{TL-RRT}^*}\geq (2 + \theta) \Big( \frac{(1+\epsilon/4)J(\tau^*) {\mu(\ccalW_{\text{free}}^N)} }{(1-y_{\text{rand}})^N (\texttt{dim}+1)\theta (1-\kappa) {\zeta_{\texttt{dim}}}}\Big)^{\frac{1}{\texttt{dim}+1}}, \nonumber
  \end{align}
  \endgroup
where $(1-y_{\text{rand}})^N$ is the probability that  $\x{rand}$ is obtained following the uniform distribution at each iteration.  Then, the biased TL-RRT$^*$ is asymptotically optimal in the sense of~\eqref{eq:opt1}.
\end{cor}}

\section{Simulation Results}\label{sec:sim}
In this section, we present three case studies, implemented using Python 3.6.3 on a computer with 2.3 GHz Intel Core i5 and 8G RAM, that illustrate the efficiency and scalability of the proposed algorithm.  
We first show the correctness and optimality of the unbiased TL-RRT$^*$. Second, we compare our sampling-based TL-RRT$^*$ algorithm, with and without bias, with the synergistic  method in~\cite{bhatia2010sampling}, the RRG method in \cite{vasile2013sampling} and the SMC method in~\cite{shoukry2017linear}, with respect to the size of regions. Finally we test the scalability of biased TL-RRT$^*$ with respect to the complexity of tasks. It shows that biased TL-RRT$^*$ outperforms the synergistic method, the RRG method and the SMC method in terms of optimality and scalability. {The implementation is accessible from~\url{https://github.com/XushengLuo/TLRRT_star}.}

In all the following case studies, the LTL formula takes the general form of $\phi = \phi_{\text{spec}} \wedge \phi_{\text{col}}$, where $\phi_{\text{spec}}$ is the specified task and $\phi_{\text{col}}$ means collision avoidance among robots.  Specifically, $\phi_{\text{col}}$ requires that, at the same timestamp and in each dimension, the distance between any two robots is larger than $R$~\cite{shoukry2017linear,kantaros2018control}. {We assume that robots are equipped with motion primitives that allow them to move safely between consecutive waypoints in the discrete plan, as in~\cite{ulusoy2013optimality}.   This can be done during the execution phase of the plans using the methods proposed, e.g., in~\cite{soltero2011collision,zhou2017collision}. Specifically, given the possibly intersecting paths returned by TL-RRT$^*$, the methods in~\cite{soltero2011collision,zhou2017collision} can design policies that stop and resume robots to avoid collisions between them.}

We consider  planning problems for robots that lie in a $1\times 1$ workspace with $W=6$ isosceles right triangular regions of interest with side length $s$, and two rectangular obstacles; see Fig.~\ref{paths}. The parameters are set as follows: the step-size $\eta$ of the function $\texttt{Steer}$ is 0.25$N$, where $N$ is the number of robots, and $w$ in the cost function \eqref{eq:cost1} is 0.2. In the biased sampling method, $p_{\text{idle}}=1$, $p_{\text{closest}} = 0.9$, {$ y_{\text{rand}}=0.99$}, $\mu_d = \mu_{\alpha}=0$, $\sigma_d = 1/3$, and $\sigma_{\alpha} = \pi/108$. The safe distance is $R=0.005$. {Considering that the obstacles are polygonal, the geodesic paths can be constructed using the visibility graph~\cite{van2000computational}.}

\subsection{Correctness and optimality using unbiased sampling}\label{sec:first}
In the first simulation, we test the correctness and optimality of the proposed algorithm with unbiased sampling. The side length $s$ is 0.15. {We consider a single robot that  is initially located at $\x{0} = (0.8,0.1)$.} The assigned task is:
\begin{align}
  \phi_{\text{spec}} & =  \Diamond(\pi_1^{\ell_1}\wedge \Diamond\pi_1^{\ell_3})  \wedge (\neg\pi_1^{\ell_1} \ccalU\pi_1^{\ell_2})  \,\wedge \nonumber \\
  &  \Diamond (\pi_1^{\ell_5} \wedge \Diamond (\pi_1^{\ell_6} \wedge \Diamond \pi_1^{\ell_4})) \wedge  (\neg\pi_1^{\ell_4} \ccalU\pi_1^{\ell_5}) \wedge \square \neg \ccalO .
\end{align}
In words, $\phi_{\text{spec}}$ requires robot 1: (a) to eventually visit region $\ell_1$ and then $\ell_3$, (b) not visit region $\ell_1$ before visiting $\ell_2$,  (c) to eventually visit region $\ell_5$ and next $\ell_6$ and then $\ell_4$, (d) not visit region $\ell_4$ before visiting  $\ell_5$ and (e) avoid obstacles.
The  LTL formula $\phi_{\text{spec}}$ corresponds to an NBA with {$|\mathcal{Q}_B|=28$ states, $|\ccalQ_B^0|=1$, and $|\ccalQ_B^F|=2$.} {which was constructed  using the tool  in \cite{gastin2001fast}.} The term $\Diamond(\pi_1^{\ell_1}\wedge \Diamond\pi_1^{\ell_3})$ can capture sequential surveillance and data gathering tasks by visiting specific regions of interest, the term  $\neg\pi_1^{\ell_1} \ccalU\pi_1^{\ell_2}$ can assign certain priority or designate  the order between different subtasks.


  \begin{table}[t]
\renewcommand{\arraystretch}{0.8}
\caption{Performance versus the number of iterations}
\label{tab:evolvCost}
\centering
 {\begin{tabular}{ccccc}
       \toprule
    $n_{\text{max}}^{\text{pre}}$ & $T(s)$    &  $J(\tau)$ &   $|\ccalP|$  \\
    \midrule
    581.9$\pm$274.1 & 47.0$\pm$37.0 & 0.619$\pm$0.046 & 1$\pm$0\\
    412.4$\pm$186.0 & 38.7$\pm$34.4 &  0.611$\pm$0.050 &1$\pm$0\\
    \midrule
    \multirow{2}{*}{600}   & 47.7$\pm$7.6 (6) & 0.572$\pm$0.034 & 3.9$\pm$2.6\\
    &  115.2$\pm$27.3 (2) & 0.543$\pm$0.027 & 4.5$\pm$2.0\\
    \midrule
    \multirow{2}{*}{800} &  95.5$\pm$10.9 (6) & 0.541$\pm$0.023 & 5.8$\pm$3.1\\
    & 282.6$\pm$68.7 & 0.514$\pm$0.020 & 8.6$\pm$3.5 \\
    \midrule
    \multirow{2}{*}{1000} & 163.2$\pm$30.7 & 0.525$\pm$0.030 & 7.9$\pm$3.6\\
    & 580.6$\pm$101.6 & 0.507$\pm$0.020 & 10.8$\pm$3.0    \\
    \bottomrule
  \end{tabular}}
\justify
  \end{table}
\begin{figure}[t]
  \centering
  \subfigure[$n_{\text{max}}^{\text{pre}}=600$]{
    \label{fig:path1}
    \includegraphics[width=0.46\linewidth]{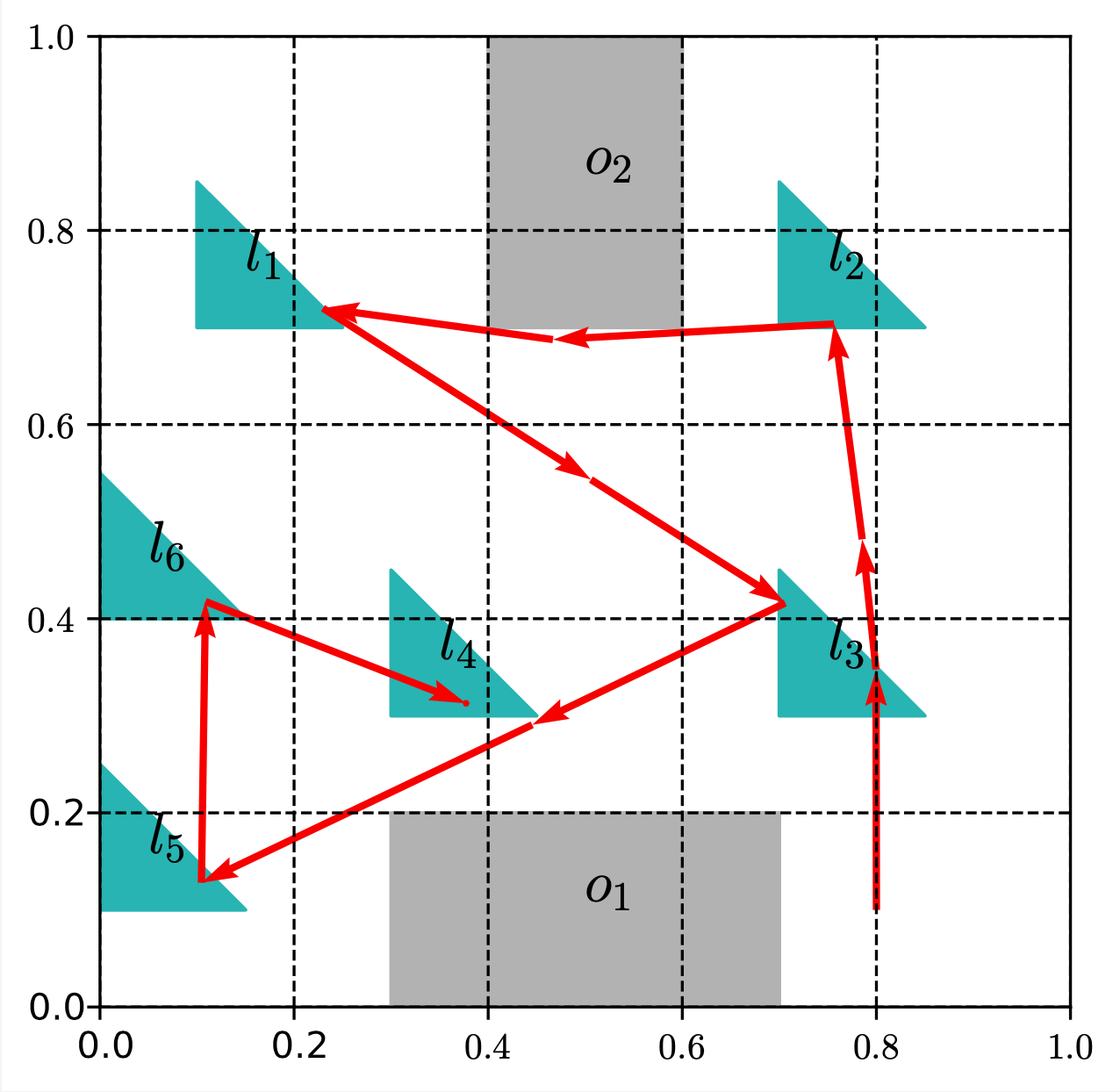}}
   \subfigure[$n_{\text{max}}^{\text{pre}}=1000$]{
    \label{fig:path4}
    \includegraphics[width=0.46\linewidth]{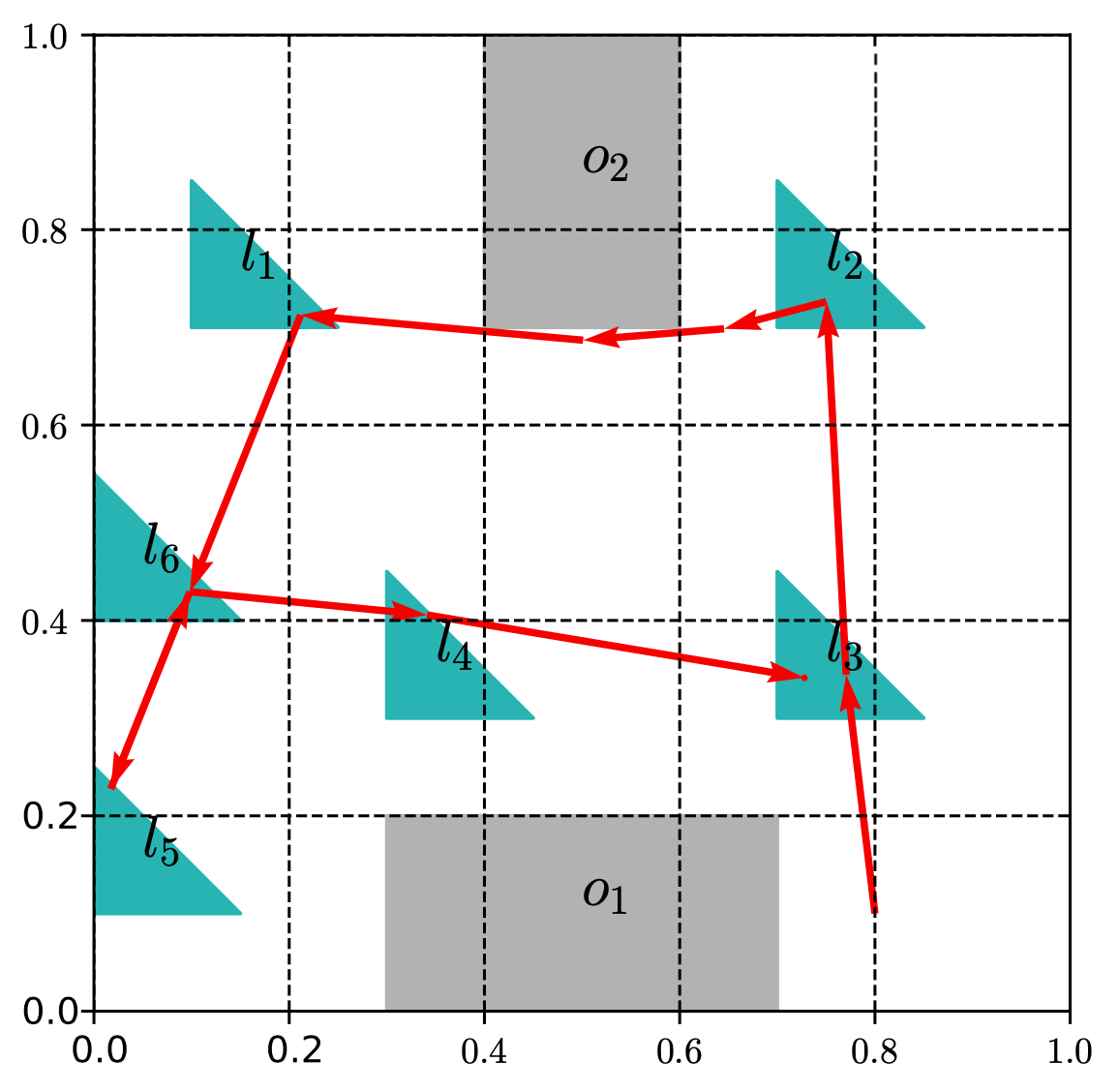}}
  \caption{Simulation Study \RNum{1}: Sample paths.}\label{paths}
\end{figure}

{To determine the connection radius $r(\ccalV_\ccalT)$ in~\eqref{eq:r}, we set $\gamma_{\text{TL-RRT}^*}$ equal to the lower bound in~\eqref{gamma}. However, note that the connection radius in~\eqref{eq:r} can not be computed because it requires knowledge of the optimal plan that is needed in~\eqref{gamma}. Therefore, we also consider the alternative connection radius}
\begingroup\makeatletter\def\f@size{10}\check@mathfonts
\def\maketag@@@#1{\hbox{\m@th\normalsize\normalfont#1}}%
\begin{align}\label{eq:r1}
  r_n(\ccalV_{\ccalT})=\min\Big\{\gamma_{\text{TL-RRT}^*}\Big(\frac{\log\left|[\ccalV_{\ccalT}]_{\sim}\right|}{\left|[\ccalV_{\ccalT}]_{\sim}\right|}\Big)^{\frac{1}{\texttt{dim}}}, \eta\Big\},
\end{align}
\endgroup
    {where $ \gamma_{\text{TL-RRT}^*} = \lceil4\left(\frac{\mu(\ccalW_{\text{free}}^N)}{\zeta_{\texttt{dim}}}\right)^{1/\texttt{dim}} \rceil$, that was proposed in~\cite{karaman2011sampling} for RRT$^*$ and has been tested extensively in practice. The main difference between~\eqref{eq:r} and~\eqref{eq:r1} is in the exponent, which is $1/(\texttt{dim}+1)$ in~\eqref{eq:r} versus $1/\texttt{dim}$ in~\eqref{eq:r1}. In what follows, to determine the lower bound on $\gamma_{\text{TL-RRT}^*}$ needed to compute the connection radius in~\eqref{eq:r}, we first use the connection radius in~\eqref{eq:r1} to find a feasible path and then use the cost of this path as an approximation of the optimal cost in~\eqref{gamma}. Moreover, we set $\theta=0.24$ and $\epsilon=\kappa=0.01$ that appear in~\eqref{gamma}.  Table~\ref{tab:evolvCost} shows statistical results, i.e., mean and standard deviation, on the total runtime, the number $|\ccalP|$ of detected final states of the prefix plan, and the cost $J(\tau)$ of the path for different numbers of iterations $n_{\text{max}}^{\text{pre}}$, averaged over 20 trails. {In the first case, the unbiased TL-RRT$^*$ was terminated when the first feasible path was detected and  in the remaining  cases the algorithm was terminated after a fixed number of iterations $n_{\text{max}}^{\text{pre}}$}. Note that this task is satisfiable by only the prefix plan. The number inside the parentheses is the number of failed trials when no plan was generated.  The first row for each value of $n_{\text{max}}^{\text{pre}}$ corresponds to the connection radius in~\eqref{eq:r1}, whereas the second row corresponds to the connection radius in~\eqref{eq:r}. {The time needed to find a feasible path using the connection radius in~\eqref{eq:r} is not included.}
      Observe that for the same connection radius, as the number of iterations increases, the cost of the computed path decreases, as expected due to Theorem~\ref{prop:opt}. Observe also in Table~\ref{tab:evolvCost} that the connection radius in~\eqref{eq:r} returns paths with lower cost compared to the connection radius in~\eqref{eq:r1}. This is because the connection radius in~\eqref{eq:r} can discover more accepting states in the set $\mathcal{P}$ and, therefore, more paths, compared to using the one in~\eqref{eq:r1}. In our simulations, we observed that the ratio of the connection radius in~\eqref{eq:r} to that in~\eqref{eq:r1} ranges from 0.8 to 6 as the tree grows. This means that the connection radius in~\eqref{eq:r} becomes increasingly larger than that in~\eqref{eq:r1} as the tree grows, and a larger connection radius facilitates the discovery of accepting states. Nevertheless, the difference in the cost of the paths returned by TL-RRT$^*$ using  the two connection radii is not very large, meaning that the connection radius in~\eqref{eq:r1} can still give a good approximation of the optimal path. This is important since~\eqref{eq:r1} is easy to compute as discussed before. Two paths for cases $n_{\text{max}}^{\text{pre}}=600, 1000$ are depicted in Fig.~\ref{paths}. {In practice, we can inflate obstacles and shrink regions to make paths more robust to  perturbations, as in~\cite{janson2018monte}.}}

\subsection{Comparison with other methods w.r.t. the size of regions}\label{sec:test2}
\begin{table*}[t]
  \caption{Comparison of runtimes and cost for the first feasible path for the LTL formula~\eqref{eq:ltl2} involving $N=2$ robots}\label{tab:test2}
  \renewcommand{\arraystretch}{1.1}
  \centering

   {\resizebox{\textwidth}{!}{\begin{tabular}{ccccccccccc}

    \toprule
  \multirow{2}{*}{$s$} & \multicolumn{5}{c}{runtime} & \multicolumn{5}{c}{cost} \\
    \cmidrule(r){2-6} \cmidrule(r){7-11}
    & $T^{\text{unbias}}_{\text{TL-RRT}^*}(s)$   & $T^{\text{bias}}_{\text{TL-RRT}^*}(s)$ &$T_{\text{SYN}}(s)$ & $T_{\text{RRG}}(s)$ & $T_{\text{SMC}}(s)$ &  $J^{\text{unbias}}_{\text{TL-RRT}^*}$& $J^{\text{bias}}_{\text{TL-RRT}^*}$ &  $J_{\text{SYN}}$ & $J_{\text{RRG}}$ & $J_{\text{SMC}}$\\
    \midrule
    $0.25$ & 19.0$\pm$34.4  & 0.4$\pm$0.3  & 124.1$\pm$41.1 &  37.9$\pm$59.2 & 15.6$\pm$3.9  & 2.42$\pm$0.41 & 1.75$\pm$0.21 &  3.22$\pm$0.46 & 3.34$\pm$0.81 & 3.19$\pm$0.71 \\
    $0.20$ & 21.4$\pm$21.2  & 0.4$\pm$0.1  &  123.9$\pm$ 63.0 &116.4$\pm$133.9  & 12.9$\pm$2.1  &  2.41$\pm$0.32 &  1.71$\pm$0.20  & 3.40$\pm$0.43 & 3.23$\pm$0.73  & 3.55$\pm$0.20 \\
    $0.15$ & 41.0$\pm$28.8  & 0.6$\pm$0.3 & 446.9$\pm$147.1 & 315.4$\pm$288.4    & 14.6$\pm$2.0    & 2.43$\pm$0.42 & 1.78$\pm$0.11  & 4.46$\pm$1.03 & 3.83$\pm$0.78 & 3.14$\pm$0.04 \\
    $0.10$ & 197.8$\pm$140.3 & 1.0$\pm$0.4  &  378.1$\pm$279.6 &  ---       & 11.6$\pm$1.0 & 2.65$\pm$0.51 & 1.76$\pm$0.12 & 6.00$\pm$1.25 & ---  & 2.70$\pm$0.00 \\
    \bottomrule
  \end{tabular}}}\\
\justify
\end{table*}

\begin{table*}[t]
  \caption{Comparison of runtimes and cost for the first 5 feasible paths for the LTL formula~\eqref{eq:ltl2} involving $N=2$ robots}\label{tab:run5}
  \renewcommand{\arraystretch}{0.8}
  \centering
 {\resizebox{\textwidth}{!}{\begin{tabular}{ccccccccccccccc}
    \toprule
  \multirow{2}{*}{$s$} &  \multirow{2}{*}{$T^{\text{unbias}}_{\text{TL-RRT}^*}(s)$} &  \multirow{2}{*}{$T^{\text{bias}}_{\text{TL-RRT}^*}(s)$} &  \multirow{2}{*}{$T_{\text{SYN}}(s)$} &  \multirow{2}{*}{$T_{\text{RRG}}(s)$} & \multicolumn{3}{c}{$T_{\text{SMC}}(s)$}  &   \multirow{2}{*}{$J^{\text{unbias}}_{\text{TL-RRT}^*}$} &  \multirow{2}{*}{$J^{\text{bias}}_{\text{TL-RRT}^*}$} &  \multirow{2}{*}{$J_{\text{SYN}}$} &  \multirow{2}{*}{$J_{\text{RRG}}$}  & \multicolumn{3}{c}{$J_{\text{SMC}}$} \\
    \cmidrule(r){6-8} \cmidrule(r){13-15}
    & &&&& 10 & 20 & $*$ & & & & &  10 &  20  & $*$  \\
    \midrule
    $0.25$  &  62.3$\pm$54.4  &  1.3$\pm$0.3 & 227.2$\pm$54.6 &  59.0$\pm$88.6 & 17.1$\pm$0.7  & 4.3$\pm$0.7 &  4.4$\pm$0.2  & 2.07$\pm$0.25 & 1.48$\pm$0.12 & 2.88$\pm$0.20 & 2.51$\pm$0.43 & 3.62   &  3.99 & 3.60 \\
    $0.20$ & 78.7$\pm$39.4  & 1.6$\pm$0.7& 250.3$\pm$52.1 & 167.7$\pm$178.1  & 18.1$\pm$0.8 & 4.6$\pm$0.2 &  4.9$\pm$0.4  &  2.17$\pm$0.28 &  1.56$\pm$0.11 & 2.66$\pm$0.18&  2.76$\pm$0.80 &  2.98  & 3.81 & 2.73   \\
    $0.15$  &  291.8$\pm$280.6  & 1.8$\pm$0.5  & 575.6$\pm$214.9&  ---   & 19.5$\pm$2.9 & 5.4$\pm$0.9 & 3.5$\pm$0.1  & 2.18$\pm$0.18 & 1.57$\pm$0.07 &  3.11$\pm$0.29& ---   & 3.22 & 2.99 & 3.00 \\
    $0.10$ & 522.3$\pm$314.3 & 3.4$\pm$1.5 &485.1$\pm$189.9 &--- & 19.8$\pm$2.2 &  5.5$\pm$1.4 & 4.5$\pm$0.3 & 2.27$\pm$0.27 & 1.67$\pm$0.10& 5.34$\pm$0.38 &---  & 2.58 & 3.30 & 2.70 \\
    \bottomrule
  \end{tabular}}}
\end{table*}

In this section,  by decreasing the size of regions, we compare TL-RRT$^*$ to {the synergistic (SYN) method}~\cite{bhatia2010sampling}, the RRG method~\cite{vasile2013sampling}  and the SMC method~\cite{shoukry2017linear}. Specifically, a team of 2 robots are initially located around  $(0.8, 0.1)$ with no collisions between them. The task is:
\begin{align}\label{eq:ltl2}
\phi_{\text{spec}}= \Box\Diamond\pi_1^{\ell_1}\wedge\Box\Diamond \pi_2^{\ell_2} \wedge \square \Diamond (\pi_1^{\ell_4} \wedge \Diamond \pi_2^{\ell_4}),
\end{align}
which requires: (a) robot 1 to visit region $\ell_1$ infinitely often, (b) robot 2 to visit region $l_2$ infinitely often, (c) {robot 1 and 2 to visit region $\ell_4$ infinitely often and robot 2 will eventually visit region $\ell_4$ after each time robot 1 visits this region}. The term $\square \Diamond (\pi_1^{\ell_4} \wedge \Diamond \pi_2^{\ell_4})$ can capture intermittent connectivity tasks that require robots to reach predetermined communication points infinitely often to exchange information but not necessarily concurrently~\cite{guo2017distributed,kantaros2018distributed}. The considered LTL formula corresponds to an NBA with $|\mathcal{Q}_B|=8$ states, $|\ccalQ_B^0|=1$, and $|\ccalQ_B^F|=2$. {We adopt the connection radius in~\eqref{eq:r1} since it is easy to compute and its use does not significantly affect the cost of the path returned by TL-RRT$^*$, as shown in Section~\ref{sec:first}.
}

{The synergistic planning method in~\cite{bhatia2010sampling} consists of a high-level planner that operates in the product space of $N$ discrete transition systems, corresponding to the $N$ robots, and the B$\ddot{\text{u}}$chi state space, and a low-level sampling-based planner that builds trees  in the continuous space guided by the high-level states. In the simulation, we use the same abstraction as that used by the SMC method and new states are sampled uniformly around  the selected high-level state.}

{The RRG method in \cite{vasile2013sampling} maintains a sparse approximation of the workspace such that states are ``far'' away from each other. To this end, it only connects new sampled states to nodes in the graph if their pairwise distance is smaller than $\eta_2(k)$ and larger than $\eta_1(k)$, {where $k$ is the number of different position states in the graph.}} To provide a fair comparison between TL-RRT$^*$ and the RRG method, we set the parameter $\eta_1(k)$ in \cite{vasile2013sampling} to be $\frac{1}{\sqrt{\pi}}\sqrt[\leftroot{-2}\uproot{2}n]{\frac{\mu(\ccalW^N)\Gamma(dN/2+1)}{k}}$ for all $k \geq 1$, where $\mu(\ccalW^N)$ is the total measure (volume) of the configuration space, $dN$ is the dimension of $\ccalW^N$, $\Gamma$ is the gamma function, and $\eta_2(k) = 2\eta_1(k)$ satisfies (i) $\eta_1(k) < \eta_2(k)$ for all $k\geq1$, (ii) $c\,\eta_1(k) > \eta_2(k)$ for some finite $c>1$ and all $k\geq0$.

Finally, in our implementation of the SMC method, we use Z3~\cite{de2008z3} as the SAT solver and CPLEX as the optimization solver. The workspace is triangulated to create a coarse abstraction and the LTL formula is encoded as Boolean constraints using Bounded Model Checking (BMC)~\cite{biere2006linear}, which involves a predetermined parameter specifying the horizon of the plan. However, a feasible initial horizon is not known a priori. Therefore, when the SMC method fails to return a solution, we re-run it after increasing the horizon by 1.

{Table~\ref{tab:test2} compares the runtimes of the different algorithms until the first feasible plan is discovered and the costs of these plans calculated as in~\eqref{eq:cost1}, averaged over 20 trials, for various choices of the side length $s$ of the labeled regions, where the symbol ``---'' means the runtime is larger than $1000s$. In the case of the SMC method, the initial horizon is 15 and the runtime of the algorithm includes the time of all failed attempts preceding the first successful attempt.  Observe first in Table~\ref{tab:test2} that the runtime of  the RRG method in~\cite{vasile2013sampling} is large. This is because, as the graph built by the RRG method grows, the parameter $\eta_1$ goes to zero and the constructed graph loses its sparsity.
Moreover, as the side length $s$ decreases, it becomes increasingly difficult for the RRG method to sample states that belong to the labeled regions, which further increases the runtime of the algorithm. A similar observation also applies to the synergistic method in~\cite{bhatia2010sampling}. Specifically, since the high-level state space is  large, being the product of two discrete transition systems one for each robot, high-level planning is expensive which increases the runtime of the algorithm. Moreover, as with the RRG method, as the side length $s$ decreases, it becomes increasingly difficult also for the synergistic method to sample new states that belong to the labeled regions, which further increases runtime.  On the other hand, TL-RRT$^*$, with or without bias, grows trees so that sparsity of the graph is not an issue as with the RRG method. Additionally, biased TL-RRT$^*$ guides the sampling process based only on the B$\ddot{\text{u}}$chi states and not on the number of robots or the size of a discrete abstraction of the environment as the synergistic method in~\cite{bhatia2010sampling}. As a result, biased TL-RRT$^*$ is much faster compared to both the RRG method and the synergistic method  for any side length $s$ of the labeled regions. Note that, while the runtime of biased TL-RRT$^*$ also increases as the side length $s$ decreases, this increase is small since sampling is strongly biased towards labeled regions no matter how small they are. Unlike the other methods,  the runtime of the SMC  method does not change much with the side length $s$ of the labeled regions since, in this example, the partition of the workspace does not change much as the shapes and locations  of the triangular regions remain mostly unchanged.  Even so, biased TL-RRT$^*$  also outperforms  the SMC method in terms of runtime in all cases. As for the cost $J(\tau)$,  the plan found by the biased TL-RRT$^*$ outperforms all other methods. This is because biasing sampling along shortest paths in the B$\ddot{\text{u}}$chi automaton has the effect that detours in the workspace that increase cost are avoided.}

{Table~\ref{tab:run5} shows comparative results  until 5 feasible paths are found using the sampling-based methods. {For the SMC  method, results correspond to the runtime and cost of the first feasible plan  for initial horizons equal to 10 and 20, where the 0 standard deviation is dropped for simplicity.} The last column in the SMC case, denoted by ``$*$'', shows the results with ``perfect'' initial horizons. A ``perfect'' initial horizon is the smallest horizon needed to obtain  a feasible plan. Here, the first successful trial was obtained for a horizon 17 for all side lengths of labeled regions. The results shown in Table~\ref{tab:run5}, further validate those discussed above in Table~\ref{tab:test2}. For all sampling-based methods, a noticeable reduction in cost can be obtained at the expense of a growing runtime.  As for the SMC method, if initialized with a smaller horizon, such as  10, it suffers more failures until a feasible plan is found, thus increasing its runtime. Since the first feasible plan is found for horizon 17, the runtime for finding a solution with a larger initial horizon 20 varies slightly. Noticeably, biased TL-RRT$^*$ can reduce cost by slightly increasing runtime. Also, here too, it significantly outperforms all other methods.}

\subsection{Scalability w.r.t the complexity of tasks and the number of robots using biased sampling}
Below we demonstrate the scalability of the biased TL-RRT$^*$ at the expense of  optimality.  {As in Section~\ref{sec:first}, we use two connection radii  in the functions $\texttt{Extend}$ and $\texttt{Rewire}$. The first connection radius is determined by \eqref{eq:r1}, and the second one is 0,  which amounts to  connecting $\q{P}{new}$ with $\q{P}{closest}$ directly. Note that the connection radius 0  suffices to find feasible plans and, thus, can fully demonstrate the scalability of the biased TL-RRT$^*$. In both cases, we set the step-size $\eta$ in the function $\texttt{Steer}$ to be large enough so that $\x{new}$ can be directly reached from $\x{closest}$.
}

We consider a team of up to $N=56$ robots  accomplishing a set of tasks. We adopt the representation in \cite{kantaros2018temporal} where the formula $\phi_{\text{spec}}$ is written in the following compact form:
\begin{align}\label{equ:ltl3}
  \phi_{\text{spec}} & = \Box\Diamond\xi_1 \wedge \Box\Diamond\xi_2 \wedge \Box\Diamond\xi_3 \wedge\Box\Diamond(\xi_4 \wedge \Diamond(\xi_5 \wedge \Diamond \xi_6)) \nonumber \\
 & \quad  \wedge \Diamond \xi_{7}  \wedge \square\Diamond\xi_{8} \wedge (!\xi_{7}~ \ccalU~ \xi_{8}).
\end{align}
The LTL formula is satisfied if (i) $\xi_1$ is true infinitely often; (ii) $\xi_2$ is true infinitely often; (iii) $\xi_3$ is true infinitely often; (iv) $\xi_4$, $\xi_5$ and $\xi_6$ are
true in this order infinitely often; (v) $\xi_7$ is true eventually; (vi) $\xi_8$ is true infinitely often; (vii) $\xi_7$ is false until $\xi_8$ becomes true for the first time. The subformula $\xi_e$ takes the Boolean form of $\wedge_{i=1}^m \pi_i^{\ell_j}$ that involves a subteam of robots. For instance, when $m=3$, $\xi_1$ can be $\xi_1= \pi_{1}^{\ell_3} \wedge \pi_{3}^{\ell_5} \wedge \pi_{5}^{\ell_1}$, which is true if (i) robot 1 is in region $\ell_3$; (ii) robot 3 is in region $\ell_5$ and (iii) robot 5 is in region $\ell_1$. All other Boolean formulas $\xi_e$ are defined similarly.  The corresponding NBA has $33$ states and $348$ edges.  Given a team of robots, we  randomly divide it into overlapping robot subteams in a way such that each robot belongs to at least one subteam. Then, we associate each subteam of robots with one subformula $\xi_e$.\footnote{This formulation amounts to a generator of tasks rather than a specific task instance. It provides a systematic approach to testing the scalability by  increasing the number of robots. E.g., an intermittent communication task~\cite{kantaros2018distributed}  can be generated as $\square \Diamond (\pi_{1}^{\ell_1} \wedge \pi_{3}^{\ell_1}) \wedge \square \Diamond (\pi_{2}^{\ell_2} \wedge \pi_{3}^{\ell_2}) \ldots$.}

 Along with pruning edges from the NBA as  discussed in Section~\ref{sec:biased}, we further delete those feasible edges with labels in the form $\wedge_e \xi_e$, provided that $\wedge_e\xi_e$ does not contain the negation of any subformula $\xi_e$, that require more than one subformulas to be true simultaneously. The reason is that as $m$ grows, each subformula will include more robots, thus, it will become harder to satisfy multiple subformulas simultaneously. After deletion, the resulting NBA has 111 edges, a dramatic drop compared to the original size. Note that the problem is still feasible since edges labeled with a single formula are intact and they contain a solution.\footnote{We can safely and automatically delete those edges since there is no conjunction of subformulas $\wedge_e \xi_e$ in the LTL formula~\eqref{equ:ltl3}. If there is a conjunction in~\eqref{equ:ltl3}, e.g., $\xi_1 \wedge \xi_2$, we can define an additional subformula $\xi_+ = \xi_1 \wedge \xi_2$ to replace the conjunction. We conducted simulations with and without such edge deletions, and observed that when these edges are deleted from the NBA, feasible plans can be found faster due to the smaller size of the pruned NBA and the edge labels that can be more easily satisfied.}
Given a team of $N$ robots, 5 different tasks $\phi_{\text{spec}}$ are generated randomly. It takes on average 2 seconds to prune the NBA. For each task, 5 sets of initial locations are randomly generated from $\ccalW_{\text{free}}\setminus \cup_{i=1}^6 \ell_i$ {with no collisions between robots}.  For each set of initial locations, we run the biased TL-RRT$^*$ 5 times, each terminating when a feasible plan is found, and compare with the SMC  method in \cite{shoukry2017linear}.~{We also tested the synergistic method in~\cite{bhatia2010sampling}, which failed to generate a plan within 1000 seconds, which agrees with the results in~\cite{shoukry2017linear}.} For each $N$, we record the runtime and cost of the SMC method, averaged over 25 trials,  if starting with the ``perfect'' initial horizon as well as the average runtime if the initial horizon is 1 step shorter than the ``perfect'' initial horizon.  the results are averaged over $125$ experiments and are reported in Table~\ref{tab:test3}. {For TL-RRT$^*$, the first row in each block shows results for the connection radius~\eqref{eq:r1} and the second row  for the  connection radius 0.
  Observe that, with connection radius in~\eqref{eq:r1},  biased TL-RRT$^*$  achieves  the lowest cost for the first feasible plans. {We found that the connection radius in~\eqref{eq:r1} decreases slowly, and, therefore, more nodes are considered by the $\texttt{Extend}$ and $\texttt{Rewire}$ functions when  adding the new state to the tree and further improving the tree.} While this increases the runtime of TL-RRT$^*$, it is still less than the runtime of the SMC method for most tasks. On the other hand,  biased TL-RRT$^*$ with connection radius 0 outperforms the SMC method with ``perfect'' or ``imperfect'' initial horizons in terms of both runtime and cost for each task.} {{The reason is that not only is  the node $\q{P}{closest}$ the only one that is considered  when adding the new state to the tree}, but also that, for robot $i$, ${\bf x}_i^{\text{new}}$ is the second point along the path from ${\bf x}_{i}^{\text{closest}}$ to $\ccalL(i)$ if the step-size $\eta$ is large enough. Therefore, connecting   $\q{P}{new}$ to $\q{P}{closest}$ ensures that  progress towards $\ccalL(i)$ is made}, and, thus, a successful transition to the next state with B$\ddot{\text{u}}$chi component $\q{B}{succ,2}$ is more likely to be made at each iteration, accelerating the detection of a feasible path. {Due to the use of geodesic shortest paths, the cost achieved by  biased TL-RRT$^*$ with connection radius 0 is close to that achieved using the  connection radius in~\eqref{eq:r1}.}
{In our simulation, we compute geodesic paths sequentially for all robots involved in one edge label. Therefore, the runtime can be further improved if these computations are done in parallel.}

\begin{table}[t]
\caption{Runtimes and cost for tasks with incremental complexity}\label{tab:test3}

  \renewcommand{\arraystretch}{1}
  \centering
    \color{black}{\resizebox{\columnwidth}{!}{\begin{tabular}{ccccccc}
    \toprule
  \multirow{2}{*}{Task} & \multicolumn{2}{c}{TL-RRT$^*$} & \multicolumn{3}{c}{SMC-based} \\
    \cmidrule(r){2-3} \cmidrule(r){4-6}
&  $T_{\text{total}}(s) $ & $ J(\tau)$  & $T_{\text{total}}(s) $ &  $J(\tau)$ & $ T^{(1)}_{\text{total}}(s) $ \\
    \midrule
    \multirow{ 2}{*}{$\phi_8$}     & 5.7$\pm$3.2 & 2.42$\pm$0.74 & \multirow{ 2}{*}{8.4$\pm$2.9} & \multirow{ 2}{*}{3.30$\pm$0.65} & \multirow{ 2}{*}{12.07} \\
    &  3.3$\pm$1.5 & 2.99$\pm$0.79    & & & \\
    \midrule
    \multirow{ 2}{*}{$\phi_{16}$}  & 56.4$\pm$50.2 & 6.81$\pm$1.53 &  \multirow{ 2}{*}{89.8$\pm$16.1} & \multirow{ 2}{*}{8.27$\pm$0.67} & \multirow{ 2}{*}{131.66}\\
    &17.5$\pm$18.1  & 7.73$\pm$1.04   & & & \\
    \midrule
    \multirow{ 2}{*}{$\phi_{24}$}  & 133.2$\pm$132.9 & 6.85$\pm$0.89   &   \multirow{ 2}{*}{170.5$\pm$8.9} &  \multirow{ 2}{*}{9.93$\pm$0.92} &  \multirow{ 2}{*}{251.43}\\
    & 17.4$\pm$10.6 & 8.75$\pm$0.78  & & & \\
    \midrule
    \multirow{ 2}{*}{$\phi_{32}$}  & 288.6$\pm$184.4 & 9.73$\pm$0.90   & \multirow{ 2}{*}{321.3$\pm$70.5} & \multirow{ 2}{*}{11.81$\pm$1.54} & \multirow{ 2}{*}{470.07} \\
    & 75.8$\pm$85.0 & 13.80$\pm$1.35  & & &\\
    \midrule
    \multirow{ 2}{*}{$\phi_{40}$}  & 601.1$\pm$326.4 & 11.51$\pm$1.22 &  \multirow{ 2}{*}{1025.6$\pm$529.0} & \multirow{ 2}{*}{14.16$\pm$1.32} & \multirow{ 2}{*}{1599.50}\\
    &  97.0$\pm$52.8  & 13.45$\pm$1.43   &&&\\
    \midrule
    \multirow{ 2}{*}{$\phi_{48}$}  & 1278.5$\pm$ 567.5 & 14.61$\pm$1.61&   \multirow{ 2}{*}{960.84$\pm$188.7}  &  \multirow{ 2}{*}{17.19$\pm$1.76} &  \multirow{ 2}{*}{1380.63}\\
    &  148.4$\pm$82.0 &  15.91$\pm$1.18  & & & \\
    \midrule
    \multirow{ 2}{*}{$\phi_{56}$} & 2245.3$\pm$419.7 & 16.50$\pm$1.81  &  \multirow{ 2}{*}{1329.53$\pm$354.8} &  \multirow{ 2}{*}{17.53$\pm$3.07} &  \multirow{ 2}{*}{1632.21}\\
    &  374.1$\pm$491.4 & 16.69$\pm$1.68 & & &\\
    \bottomrule
  \end{tabular}}}\\
\justify
{Simulation Study \RNum{3}: $\phi_N$ denotes a task that  involves $N$ robots. For biased TL-RRT$^*$, $T_{\text{total}}$ is the total runtime needed to prune the NBA and find the first feasible prefix and suffix plan. For SMC, $T_{\text{total}}$ represents the total runtime needed  by the SAT solver and the CPLEX solver, with ``perfect'' initial horizons. $ T^{(1)}_{\text{total}}$ is the average total runtime needed  if the initial horizon is 1 step shorter than the smallest horizon that provides a feasible plan.}
      \end{table}

\section{Conclusion}\label{sec:concl}

The majority of existing LTL planning approaches rely on a discrete abstraction of robot mobility to construct a product automaton which is then used to synthesize discrete motion plans. The limitation of these approaches is that both the abstraction process and the control synthesis are computationally expensive and that the resulting discrete plans are only optimal given the discrete abstraction that was used to generate them. In this paper, we proposed a new sampling-based LTL planning algorithm, with unbiased and biased sampling, which does not require any discrete abstraction of robot mobility and avoids the construction of a product automaton. Instead, it builds incrementally a tree that can explore the workspace and the state-space of an NBA that captures a given LTL specification, simultaneously. We showed that our algorithm is probabilistically complete and asymptotically optimal, and provided numerical experiments that showed that our method outperforms relevant temporal planning methods.




%
\bibliographystyle{IEEEtran}
\bibliography{IEEEabrv,YK_bib}
%
%
\appendices
\section{Proof of Theorem \ref{prop:compl}}\label{app:comp}
To prove completeness of TL-RRT$^*$, we cannot directly adopt the proof for RRT\cite{kuffner2000rrt} since in the product state space considered here the transition relation differs from that in the continuous space. Particularly, while in continuous space a transition is valid between two points  if they are connected by an obstacle-free straight line, for LTL task planning a transition is valid if the obstacle-free straight line additionally does not cross any labled region more than once and the corresponding label enables a transition to at least one B$\ddot{\text{u}}$chi state (otherwise the LTL formula is violated). Nevertheless,  we can still draw on the same high-level idea in~\cite{kuffner2000rrt} to show completeness by first proving that the tree can grow arbitrarily close to any state that is one-hop-reachable from the root, and then extending this concept to multi-hop-reachable states.

Before stating our main results, we provide some necessary notations. Let $\ccalR_P(q_P)$ denote the one-hop reachable set of $q_P$ in $\ccalQ_P$, i.e., $ \ccalR_P(q_P) = \{ q'_P \in \ccalQ_P \,|\, q_P \rightarrow_P q'_P\}$. Then, let {$P_r$} denote the subspace of $\mathcal{W}^{N}_{\text{free}} \times \mathcal{Q}_B$, consisting of states that can be reached from the root through a multi-hop path. Furthermore,  $P_r$ can be divided into two sets, {$P_s$} and $P_v$, where $P_s = \{ q_P \in P_r\,|\, \ccalR_P{(q_P)} \not= \varnothing\}$ and $P_v = \{ q_P \in P_r\,|\, \ccalR_P{(q_P)} = \varnothing\}$. Given a state $q_P \in P_s$, let $\q{P}{cp} \in \ccalV_{\ccalT}$ be its ``virtual'' candidate parent if there exists a feasible transition from $\q{P}{cp}$ to $q_P$. Recall that {$\q{P}{near}$ is a candidate parent of $\q{P}{new}$ in the $\texttt{Extend}$ function if both $\x{near}\rightarrow \x{new}$ and $(\q{B}{near}, L(\x{near}), \q{B}{new}) \in \to_B$. Here, we use ``virtual'' since actually $q_P$ is not necessarily sampled but $\q{P}{new}$ is. For simplicity, we use the term candidate parent below.} Furthermore, let $D_k(q_P)$ denote a random variable whose value is the Euclidean distance between $q_P$ and its  nearest candidate parent node $\q{P}{ncp} = (\x{ncp}, \q{B}{ncp})$ in a tree of $k$ nodes. One specific realization of $D_k(q_P)$ is denoted by $d_k(q_P)$. Finally,  we will use the following theorem.

\begin{thm}[\cite{rudin2006real}] \label{thm:ip}
  Suppose $\{a_n\}$ is a sequence of {real} numbers and $0 \leq a_n < 1$. Then
    $\prod_{n=1}^{\infty} (1 - a_n) >0 \iff \sum_{n=1}^{\infty}a_n < \infty$.
\end{thm}

In what follows we first present results that are necessary to show that TL-RRT$^*$ is probabilistically complete.  For simplicity, $D_k$ and $d_k$ denote $D_k(q_P)$ and $d_k(q_P)$, respectively. Lemmas~\ref{thm:step1} to~\ref{thm:step3} describe how $D_k$ evolves as the tree grows. Together they establish Propositions~\ref{thm:cvx} and~\ref{thm:cvx2}, which bear a similarity to Lemma 1 in~\cite{kuffner2000rrt} in that both focus on states that can be reached from the root in one hop. In Lemma 1 in~\cite{kuffner2000rrt}, the  states belong to a convex set whereas Propositions~\ref{thm:cvx} and~\ref{thm:cvx2} consider one-hop-reachable states explicitly due to different definitions of the transition relations.

\begin{figure}[t]
\begin{center}
  \includegraphics[width=0.3\textwidth]{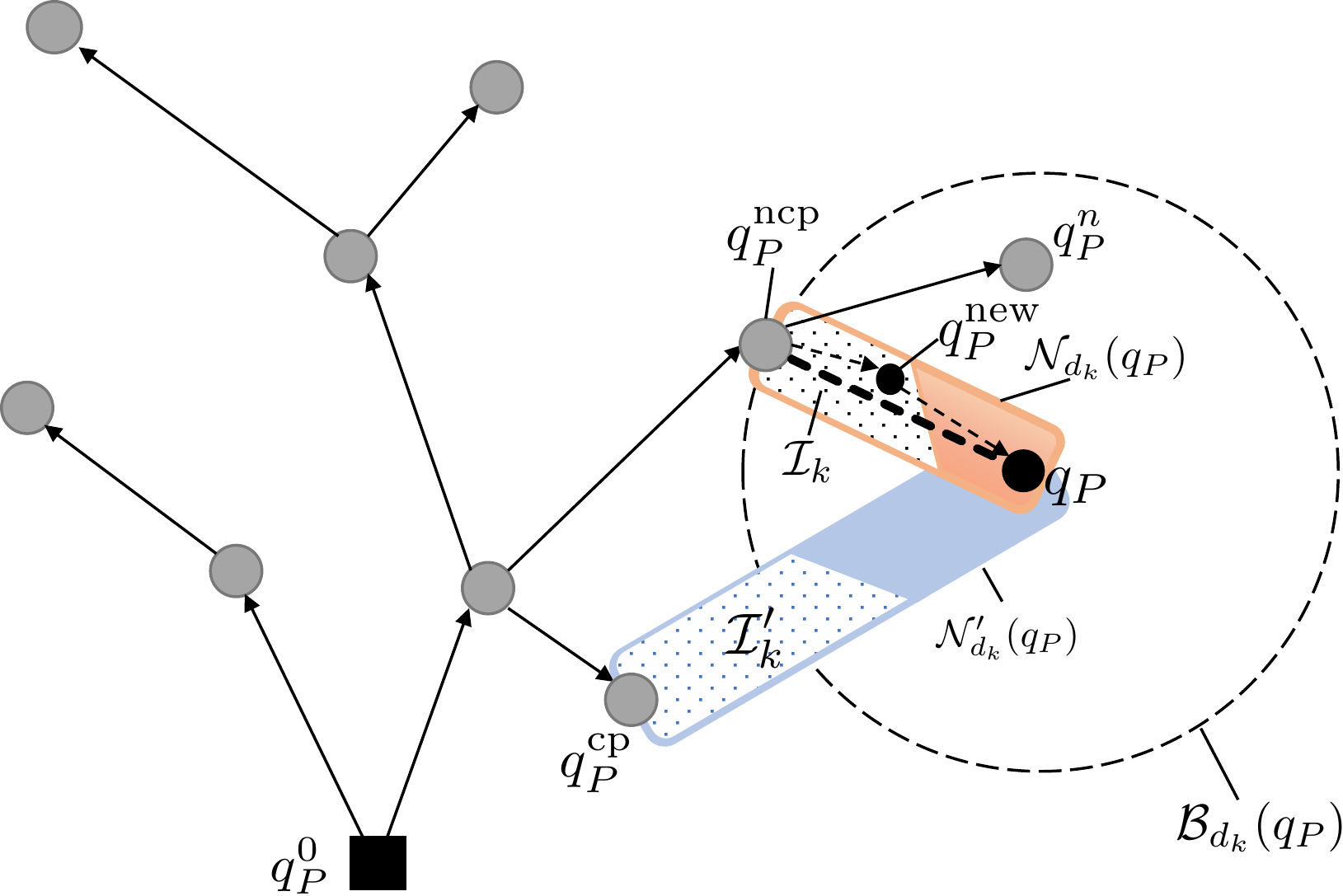}
  \caption{An illustration of the scenario where the distance between $q_P$ and its nearest candidate parent node drops. Although $q_{P}^{n}$ is closer to $q_P$ than $\q{P}{ncp}$, it is not the candidate parent of $q_P$. $\q{P}{ncp}$ is the nearest candidate parent of both $\q{P}{new}$ and $\q{P}{}$. The orange region denotes the neighborhood around the feasible path depicted as the thick dashed line, from $\q{P}{ncp}$ to $q_P$, and the thin dashed line illustrates the alternative of the path in the tree. The dotted area denotes the intersection $\mathcal{I}_k$ of the Voronoi cell $\ccalC(\q{P}{ncp})$ and the neighborhood around the path.   $\q{P}{new}$ is inside $\mathcal{I}_k$. The blue region depicts the case for another candidate parent $\q{P}{cp}$ where $\q{P}{new} \in \mathcal{I}'_k \cap \mathcal{B}_{d_k}(q_P)$.}
  \label{fig:expdrop}
\end{center}
\end{figure}

\begin{lem}\label{thm:step1}
 { Let $q_P = (\vect{x}, q_B) \in P_s$ be any state that can be reached from the root $(\vect{x}_0, q_B^0)$ in a one-hop transition, namely, $q_P \in \ccalR_P(\q{P}{0})$.} Then, $\mathbb{E}(D_{k+1} | D_k = d_k) < d_k$.
\end{lem}
\begin{IEEEproof}
  {Consider any fixed product state $q_p \in \ccalR_P(q_p^0)$. Observe that $d_1$ is equal to the distance between the root and state $q_P$ when the tree only contains the root. Therefore, $d_k < \infty$ for $k\geq1$ since $q_P \in \ccalR_P(q_P^0)$ and the sequence $\{d_k\}$ is non-increasing. Given the realization $d_k$, if the new node $\q{P}{new}$ is not a candidate parent of $q_P$, we have $D_{k+1}= d_k$. {} Otherwise, there are two cases. First, the distance between  $\q{P}{new}$ and $q_P$ may be larger than or equal to $d_k$. In this case, the nearest candidate parent of $q_P$ remains the same, so that $D_{k+1} = d_k$. Second, the distance between $\q{P}{new}$ and $q_P$ is smaller than $d_k$. In this case, $D_{k+1}<d_k$. Let $p_{k}$ denote the probability of the event $\{D_{k+1} <d_k\}$, so that the event $\{D_{k+1}=d_k\}$ occurs with probability $1-p_k$. Then, the expectation $\mathbb{E}(D_{k+1})$ can be written as
\begin{align}
  \mathbb{E}(D_{k+1}| D_k = d_k) = (1-p_{k})\,d_k + p_{k}\,d \label{equ:dk1}
\end{align}
where $d < d_k$ is a positive number.

In what follows, we show that $p_k$ is strictly positive.
Specifically, we define a ball $\mathcal{B}_{d_k}(q_P)$ with radius $d_k$ that is equal to the distance between $q_P$ and its nearest candidate parent node $\q{P}{ncp}$. If $k=1$, then $\q{P}{ncp} = \q{P}{0}$, because $q_P \in \ccalR_P(q_P^0)$. As the tree grows, it is possible that another node becomes a new $\q{P}{ncp}$ for $q_P$. Since $q_P$ is reachable from $q_P^{\text{ncp}}$, there exists a straight  path between $\bbx$ and $\bbx^{\text{ncp}}$, and the feasible transition $ \q{P}{ncp} \rightarrow_{P} q_P  $ can be potentially an edge/path added to the tree. By Assumption~\ref{asmp:nhb}, for every point $(\bbx', q'_B)$ on this path, there exists a ball with radius $\delta_{\bbx'}$ in which all states have the same B$\ddot{\text{u}}$chi state component as $(\bbx', q'_B)$. Let $\delta_{\min}$ denote the minimum value of all $\delta_{\bbx'}$, i.e.,
\begin{align}\label{eq:cylinder}
  \delta_{\min} = \inf\{\delta_{\bbx'} \,|\, \bbx' = \lambda\bbx + (1-\lambda)\bbx^{\text{ncp}},0\leq \lambda\leq1 \}.
\end{align}
Then, a tube neighborhood with radius $\delta_{\min}$ along the path connecting  $\q{P}{ncp}$ and  $q_P$ can be constructed, denoted by $\mathcal{N}_{d_k}(q_P)$, which is convex and obstacle-free; see also Fig.~\ref{fig:expdrop}. Moreover, define the Voronoi partition \cite{aurenhammer1991voronoi} of the space based on the nodes of the tree and let $\ccalC(\q{P}{ncp})$ denote the Voronoi cell around state $\q{P}{ncp}$. Since the set $\mathcal{N}_{d_k}(q_P)$ has positive measure, i.e., $\mu(\mathcal{N}_{d_k}(q_P)) > 0$, we have that the set $\mathcal{I}_k = \ccalC(\q{P}{ncp})\cap \ccalN_{d_k}(q_P)$ also has positive measure. {Note that we will use the notations $\ccalN_{d_k}(q_P),  \ccalC(\q{P}{ncp}), \mathcal{I}_k$ and their variations throughout the rest of this paper.} Since the unbiased sampling function is supported in the free workspace, the  probability that $\x{new}$  lies within $\mathcal{I}_k$ is non-zero. If this happens, since $\ccalI_k$ is obstacle-free, the condition $\x{ncp}\rightarrow \x{new}$ is satisfied, which infers that the path $(\x{ncp}, \q{B}{ncp}) \rightarrow_P (\x{new}, \q{B}{new}) \rightarrow_P (\vect{x}, q_B)$ is valid, where $(\bbx, q_B) = q_P$ and $\q{B}{new}$ is either $\q{B}{ncp}$ or $q_B$. {Note that node $(\x{ncp}, \q{B}{ncp})$ is in the set $\ccalQ_{P}^{\text{nearest}}$ with respect to  $\q{P}{new}$, because $\x{new}$ is in $\ccalC(\q{P}{ncp})$}. Therefore, the state $\q{P}{new}$ will be added to the tree, and it will become the new nearest candidate parent of $q_P$. Hence, $p_k > 0$ and \eqref{equ:dk1} becomes:
\begin{align}\label{equ:inequality}
  \mathbb{E}(D_{k+1}| D_k = d_k) &= (1-p_{k})\, d_k + p_{k}\, d   \nonumber \\
  & < (1-p_{k})\, d_k + p_{k}\, d_k  = d_k
\end{align}
which completes the proof.}
\end{IEEEproof}

{\begin{rem}
In the proof of Lemma~\ref{thm:step1}, we only consider when $\q{P}{new}$ is inside the intersection of $\mathcal{I}_{k}$ and $\mathcal{B}_{d_{k}}(q_P)$. However, $\mathcal{I}_{k}$ involves $q_P$ and the nearest candidate parent $\q{P}{ncp}$. Indeed,  the true value of $p_{k}$ is larger than the one in Lemma~\ref{thm:step1}. For example, consider when there is another candidate parent $\q{P}{cp}$ of $q_P$ which is further from $q_P$ than $\q{P}{ncp}$ is. If $\q{P}{new}$ is located inside the intersection of $\mathcal{I}'_{k} = \ccalC(\q{P}{cp})\cap \ccalN'_{d_k}(q_P)$  and $\mathcal{B}_{d_{k}}(q_P)$, where $ \ccalN'_{d_k}(q_P)$ is generated by $\q{P}{cp}$ and $\q{P}{}$, we can also have that  $D_{k+1} < d_{k}$; see Fig.~\ref{fig:expdrop}.
\end{rem}}

Let $\{\mathbb{E}(D_k)\}$ denote the sequence $\mathbb{E}(D_1), \mathbb{E}(D_2), \ldots$. We  show that $\{\mathbb{E}(D_k)\}$ is strictly decreasing, which serves as an intermediate step to prove this sequence converges to 0.
\begin{lem}\label{thm:step2}
  $\mathbb{E}(D_{k+1}) < \mathbb{E}(D_k)$ holds.
\end{lem}

\begin{IEEEproof}
 Let $f(d_k)$ denote the probability density function of the random variable $D_k$. Multiplying both sides of (\ref{equ:inequality}) by $f(d_{k})$ and integrating over the support of  $D_{k}$, we have
\begin{align}
  \int_{D_k}\mathbb{E}(D_{k+1} | D_k = d_k) f( d_k) \diff{d_k} & < \int_{D_k} d_k f(d_k) \diff{d_k}. \label{equ:ineq}
\end{align}
By the law of total expectation $\mathbb{E}(X) = \mathbb{E}(\mathbb{E}(X|Y))$, where $X$ and $Y$ are two random variables, the left-hand side in \eqref{equ:ineq} is equivalent to $\mathbb{E}(D_{k+1})$.
By definition, the right-hand side in (\ref{equ:ineq}) is $\mathbb{E}(D_k)$. Thus, $\mathbb{E}(D_{k+1}) <   \mathbb{E}(D_k)$.
\end{IEEEproof}
Note that the sequence $\{\mathbb{E}(D_k)\}$ is lower bounded by 0 {since $q_P \in \ccalR_p(q_P^0)$}. The following lemma shows that the decreasing sequence $\{\mathbb{E}(D_k)\}$ in fact converges to zero.

\begin{lem}\label{thm:step3}
We have that $\mathbb{E}(D_k) \rightarrow 0$ as $k\to\infty$.
\end{lem}
\begin{IEEEproof}
  We will show this proposition by contradiction. Specifically, assume that $\inf\{\mathbb{E}(D_k)\} = b >0$, which means that for any $\epsilon>0$, there exists $K\in \mathbb{N}^+$ so that for all $k >K$, we have $b < \mathbb{E}(D_k)  <b + \epsilon$. Note that $\mathbb{E}(D_k) \neq b$, otherwise $\mathbb{E}(D_{k+1}) < b$ since $\{\mathbb{E}(D_k)\}$ is strictly decreasing by Lemma~\ref{thm:step2}. Moreover, let $k_{\epsilon} = \inf\{k:\mathbb{E}(D_k)  < b+\epsilon\}$ denote the number of nodes in the tree when $\mathbb{E}(D_k) < b+ \epsilon$ holds for the first time. Hence, $\mathbb{E}(D_{k_{\epsilon}}) < b + \epsilon$.

Define  the ball $\mathcal{B}_{k_{\epsilon}/2}(q_P)$ with radius $d_{k_{\epsilon}}/2$, centered at $q_P$, where $d_{k_{\epsilon}}$ is the realization of $D_{k_{\epsilon}}$; see Fig.~\ref{fig:dkn}. Recall that $\mathcal{I}_{k_{\epsilon}}$ denotes the intersection of the Voronoi cell around $\q{P}{ncp}$ and the neighborhood of the feasible path from $\q{P}{ncp}$ to $\q{P}{}$; see the proof of Lemma~\ref{thm:step1}. If $\q{P}{new}$ lies within $ \mathcal{I}_{k_{\epsilon}}$, then we have $D_{k_{\epsilon}+1} < d_{k_{\epsilon}}$. Let $\gamma_{k_{\epsilon}}  = \sup\{ \|q'_P|_{\ccalW^N_{\text{free}}}  - \x{ncp}\|: q'_P \in \mathcal{I}_{k_{\epsilon}} \cap \texttt{line}(\q{P}{ncp}, q_P)\}$, where $\texttt{line}(\q{P}{ncp}, q_P)$ is the line segment connecting $\q{P}{ncp}$ and $q_P$. The quantity $\gamma_{k_{\epsilon}}$ is the maximum possible decrease from $d_{k_{\epsilon}}$ if $\q{P}{new} \in \mathcal{I}_{k_{\epsilon}} \cap \texttt{line}(\q{P}{ncp}, q_P)$. Since $\mu(\mathcal{I}_{k_{\epsilon}}) > 0$, we have that $\gamma_{k_{\epsilon}} > 0$. Therefore, we can draw a ball of radius $d_{k_{\epsilon}}-\gamma_{k_{\epsilon}}/2$ centered at $q_P$ that is contained in the ball $\ccalB_{d_{k_\epsilon}}(q_P)$. If $\q{P}{new}$ is inside the intersection of $\mathcal{I}_{k_{\epsilon}}$ and $\mathcal{B}_{(d_{k_{\epsilon}}-\gamma_{k_{\epsilon}}/2)}(q_P)$, denoted by $\mathcal{IB}_{k_{\epsilon}}$, we have that $D_{k_{\epsilon}+1} < d_{k_{\epsilon}} - \gamma_{k_{\epsilon}}/2$, whose probability  satisfies $p_{k_{\epsilon}+1}>0$.
\begin{figure}[t]
\begin{center}
  \includegraphics[width=0.25\textwidth]{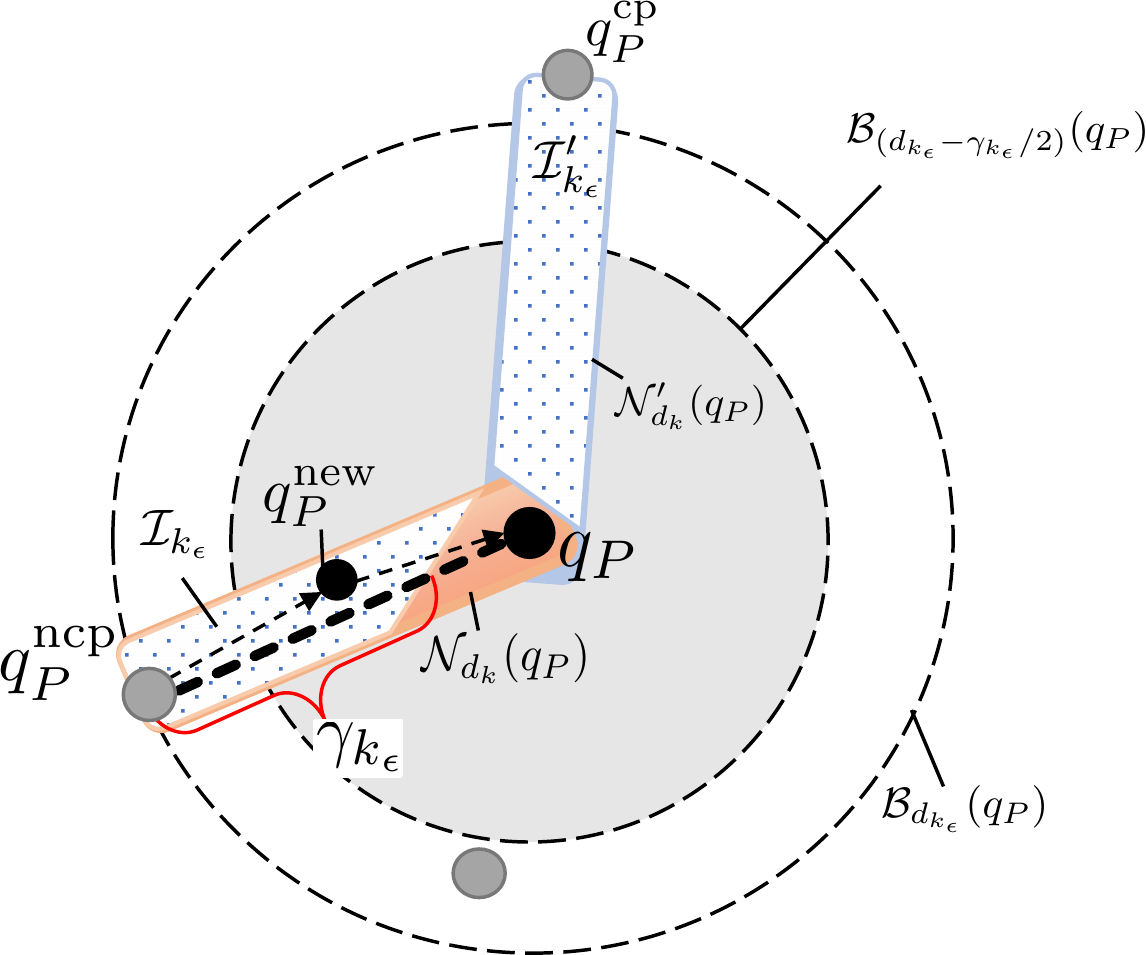}
  \caption{An illustration of the scenario where $D_{k_{\epsilon}+1}$ drops below $d_{k_{\epsilon}} - \gamma_{k_{\epsilon}}/2$. The brown region illustrates the neighborhood around the feasible path depicted as the thick dashed line, from $\q{P}{ncp}$ to $q_P$, and the thin dashed line illustrates alternative of the path in the tree. The dotted area denotes the intersection $\mathcal{I}_{k_{\epsilon}}$ of the Voronoi cell around $\q{P}{ncp}$ and the neighborhood around the path. $\q{P}{new}$ is inside $\mathcal{IB}_{k_{\epsilon}}$. The blue region depicts the case for another candidate parent $\q{P}{cp}$ where $\q{P}{new} \in \mathcal{I}'_{k_{\epsilon}} \cap \mathcal{B}_{(d_{k_{\epsilon}}-\gamma_{k_{\epsilon}}/2)}(q_P)$.}
  \label{fig:dkn}
\end{center}
\end{figure}

Now let $\{D_{k_{\epsilon}+m}\}$ denote a sequence of random variables starting from $D_{k_{\epsilon}}$ for {any} $m \in \mathbb{N}^+$. Note that as the tree grows, more candidate parents of $q_P$ will appear. 
For this, it suffices that $\q{P}{new}$ lies in the intersection of the Voronoi cell around one of the candidate parents $\q{P}{cp}$, the neighborhood $\ccalN_{d_{k_\epsilon}}(q_P) $ of the feasible path connecting $\q{P}{cp}$ and $q_P$, and the ball $\mathcal{B}_{(d_{k_{\epsilon}}-\gamma_{k_{\epsilon}}/2)}(q_P)$. Let $\ccalQ_{k_{\epsilon}+i}^{\text{cp}}$ denote the set of all candidate parents of $q_P$ 
for $i \in \{0,\ldots, m-1\}$, and $\cup\,\mathcal{IB}_{k_{\epsilon}+i}$ denote the union of the intersections $\mathcal{IB}^1_{k_{\epsilon+i}}, \ldots, \mathcal{IB}^{|Q_{k_{\epsilon}+i}^{\text{cp}}|}_{k_{\epsilon}+i}$, associated with every candidate parent node of $q_P$ in $\ccalQ_{k_{\epsilon}+i}^{\text{cp}}$. Following a similar argument as above, we can show that, if $\q{P}{new} \in \cup\,\mathcal{IB}_{k_\epsilon +i}$, then $D_{k_{\epsilon}+i+1} < d_{k_{\epsilon}} - \gamma_{k_{\epsilon}}/2$, which occurs with probability $p_{k_\epsilon+i+1}$. 
Note that $\{D_{k_{\epsilon}+m} < d_{k_{\epsilon}} - \gamma_{k_{\epsilon}}/2\}$ will occur if there exists $i \in  \{0,\ldots, m-1\}$ such that $D_{k_{\epsilon}+i+1} < d_{k_{\epsilon}} - \gamma_{k_{\epsilon}}/2$. Then, conditioned on $D_{k_{\epsilon}} = d_{k_{\epsilon}}$, the event $\{d_{k_{\epsilon}} - \gamma_{k_{\epsilon}}/2 < D_{k_{\epsilon}+m} \leq d_{k_{\epsilon}}\}$ occurs with probability $\prod_{i=0}^{m-1} (1 - p_{k_{\epsilon}+i+1})$. Moreover, with probability $1 - \prod_{i=0}^{m-1} (1 - p_{k_{\epsilon}+i+1})$, the event $\{D_{k_{\epsilon}+m} \leq d_{k_{\epsilon}} -\gamma_{k_{\epsilon}}/2\}$ occurs. Therefore, $\mathbb{E}(D_{k_{\epsilon}+m}| D_{k_{\epsilon}} = d_{k_{\epsilon}}) $ can be bounded as:
\begingroup\makeatletter\def\f@size{10}\check@mathfonts
  \def\maketag@@@#1{\hbox{\m@th\normalsize\normalfont#1}}%
\allowdisplaybreaks
\begin{align}
   \mathbb{E}(D_{k_{\epsilon}+m}| D_{k_{\epsilon}} & = d_{k_{\epsilon}})
  \leq  d_{k_{\epsilon}}  \prod\nolimits_{i=0}^{m-1} (1 - p_{k_{\epsilon}+i+1})  \label{equ:expinq} \\
  & \quad+  \left(d_{k_{\epsilon}} - \gamma_{k_{\epsilon}}/2\right) \Big[1- \prod\nolimits_{i=0}^{m-1} (1 - p_{k_{\epsilon}+i+1})\Big]. \, \nonumber
\end{align}
\endgroup
Multiplying both sides of (\ref{equ:expinq}) by $f(d_{k_{\epsilon}})$, integrating over the support of $D_{k_{\epsilon}}$, and denoting by $\Delta = \frac{1}{2} \int_{D_{k_{\epsilon}}} \gamma_{k_{\epsilon}} f(d_{k_{\epsilon}}) \diff{d_{k_{\epsilon}}} > 0$, we obtain
\begingroup\makeatletter\def\f@size{8}\check@mathfonts
  \def\maketag@@@#1{\hbox{\m@th\normalsize\normalfont#1}}%
  \allowdisplaybreaks
  \begin{align}
    \mathbb{E}(D_{k_{\epsilon}+m}) & \leq   \mathbb{E}(D_{k_{\epsilon}}) \prod\nolimits_{i=0}^{m-1} (1 - p_{k_{\epsilon}+i+1}) \nonumber \\
    & \quad +  \left(\mathbb{E}(D_{k_{\epsilon}}) - \Delta \right) \Big[1- \prod\nolimits_{i=0}^{m-1} (1 - p_{k_{\epsilon}+i+1})\Big]  \nonumber \\
    & < (b+\epsilon)  \prod\nolimits_{i=0}^{m-1} (1 - p_{k_{\epsilon}+i+1})  \nonumber \\
    & \quad+  (b+\epsilon - \Delta)  \Big[1- \prod\nolimits_{i=0}^{m-1} (1 - p_{k_{\epsilon}+i+1})\Big] \nonumber \\
    & = (b + \epsilon) - \Big[1- \prod\nolimits_{i=0}^{m-1} (1 - p_{k_{\epsilon}+i+1})\Big] \Delta. \label{equ:exp}
  \end{align}
  \endgroup
{where the strict inequality comes from the fact $\mathbb{E}(D_{k_{\epsilon}}) < b + \epsilon$.}

By Assumption~\ref{asmp:nhb}, the sequence $\{p_{k_\epsilon+i+1}\} _{i=0}^{m-1}$ is bounded away from zero for all $i\in\{0, \ldots, m-1\}$ for any $m\in\mathbb{N}^+$. It does not converge to 0 since the ball $\ccalB_{d_{k_{\epsilon}}}(q_P)$ is fixed given $k_{\epsilon}$ and increasingly more candidate parents of $q_P$ will appear as the tree grows, {their position components forming a dense subset of $\ccalW^N$. By Assumption~\ref{asmp:nhb}$, \ccalI\ccalB_{k_{\epsilon}+i}$ covers a subset of $\ccalW^N$ with positive measure.} Therefore $\sum_{i=0}^{\infty} p_{k_\epsilon+i+1} = \infty$. Moreover, since $0<p_{k_\epsilon+i+1}<1$, by Theorem~\ref{thm:ip}, we can conclude that $\lim_{m\rightarrow\infty}\prod_{i=0}^{m-1} (1 - p_{k_{\epsilon}+i+1}) = 0 $. Hence, $\lim_{m\rightarrow\infty}1 - \prod_{i=0}^{m-1} (1 - p_{k_{\epsilon}+i+1}) = 1$. 
Then, as $m \rightarrow \infty$, \eqref{equ:exp} becomes
   $\mathbb{E}(D_{k_{\epsilon}+m})  < (b + \epsilon) - \Delta$.
Since $\epsilon$ can be any arbitrary small non-negative number, we can select $\epsilon = \Delta/2$. Then $\mathbb{E}(D_{k_{\epsilon}+m}) < b-\Delta/2 < b$ as $m\rightarrow\infty$, which contradicts the assumption that $b$ is the infimum. Hence, 0 is the infimum of the sequence $\{\mathbb{E}(D_k)\}$. By the monotone convergence theorem, we get that $\mathbb{E}(D_k) \rightarrow 0 $.
\end{IEEEproof}


Using the previous lemmas for unbiased sampling, the following proposition asserts that the tree built by TL-RRT$^*$ rooted at $q_P^0=(\bbx^0,q_B^0)$ will grow arbitrarily close to any state $q_P = (\vect{x}, q_B) \in P_s$ that is  one-hop-reachable from the root.

\begin{prop}
  \label{thm:cvx}
  For any state $q_P = (\vect{x}, q_B) \in P_s \cap \ccalR_{P}(q_P^0)$ in the tree and for any positive constant $ \delta > 0$, we have that
  $\lim_{k\rightarrow\infty} \mathbb{P} (D_k (q_P) \leq \delta) = 1.$
\end{prop}
\begin{IEEEproof}
  According to Lemma~\ref{thm:step3}, $\mathbb{E}(D_k)\to0$ as $k \to \infty$. Following the same logic, we can get that $\mathbb{E}(D_k^2) \rightarrow 0 $. Hence, the variance $\mathrm{Var}(D_k) = \mathbb{E}(D_k^2) - [\mathbb{E}(D_k)]^2 \rightarrow 0$. Since $\mathbb{E}(D_k) \rightarrow 0 $ and $D_k \geq 0 $, for any small number $\delta>0$ there exists a sufficiently large $k \in \mathbb{N}^+$, so that
$\mathbb{E}(D_k) < \delta/2$.
  Therefore, we obtain
 $ \mathbb{P}( D_k  > \delta) \leq \mathbb{P}( | D_k - \mathbb{E}(D_k)| > \delta/2 )  \leq 4\,\mathrm{Var}(D_k)/\delta^2$,
where the first inequality holds because (i) $D_k \leq |D_k - \mathbb{E}(D_k)| + \mathbb{E}(D_k)$ by the triangle inequality, and (ii) $\mathbb{E}(D_k) < \delta/2$, and the second is obtained by Chebyshev inequality. Since $\mathrm{Var}(D_k) \rightarrow 0$, we get that $\mathbb{P}(| D_k | > \delta) \rightarrow 0$ and further that $D_k \rightarrow0$ in probability, namely, for any $\delta>0$, it holds that
  $\lim_{k\rightarrow\infty} \mathbb{P}(D_k \leq \delta) = 1$.
\end{IEEEproof}

    Next, we show that the nodes in the tree that are  arbitrarily close to $q_P$ share the same B$\ddot{\text{u}}$chi state as $q_P$.
\begin{prop}\label{thm:cvx2}
  Given any state $q_P = (\vect{x}, q_B) \in P_s \cap \ccalR_{P}(q_P^0)$, the probability that there exists a node in the tree constructed by TL-RRT$^*$, which shares the same B$\ddot{\text{u}}$chi state as $q_P$ and lies within distance $\delta>0$ from $q_P$ satisfies
  \begingroup\makeatletter\def\f@size{10}\check@mathfonts
  \def\maketag@@@#1{\hbox{\m@th\normalsize\normalfont#1}}%
  \begin{align*}
    \lim_{k \rightarrow \infty} \mathbb{P}\left( \left \{\exists\, q'_P = (\vect{x}', \qb) \in \mathcal{V}_k^{\text{TL-RRT}^*}: \| \vect{x} - \vect{x'}\| \leq \delta\right \}\right) = 1.
  \end{align*}
  \endgroup
  where $\mathcal{V}_k^{\text{TL-RRT}^*}$ is the set of nodes in the tree constructed by TL-RRT$^*$ when the number of nodes is $k$.
\end{prop}
\begin{IEEEproof}
  To see this, recall from the proof of the Lemma~\ref{thm:step3} that with probability $1 - \prod_{i=0}^{m-1} (1 - p_{k_{\epsilon}+i+1})$, the event $\{D_{k_{\epsilon}+m} \leq d_{k_{\epsilon}} - \gamma_{\epsilon/2}\}$ occurs. Next, we show this is also the probability that there is a node inside $\mathcal{B}_{(d_{k_{\epsilon}}-\gamma_{k_{\epsilon}}/2)}(q_P)$ which has the same B$\ddot{\text{u}}$chi state as $q_P$. The reason is that if the new point $\x{new}$ lies within the set $\cup\,\mathcal{IB}_{k_{\epsilon}+i}$, {{defined in the proof of Lemma~\ref{thm:step3},}} then $q_B$ can be paired with $\x{new}$ to create a new node, because (i) $\x{cp} \rightarrow \x{new}$ holds since $\x{new}$ is in $\ccalI_{k_{\epsilon}+i}$, and (ii) $\q{B}{cp} \xrightarrow{L(\x{cp})} q_B$ since $\q{P}{cp}\rightarrow_P q_P$. When $m\rightarrow\infty$, this probability goes to 1,
  \begingroup\makeatletter\def\f@size{10}\check@mathfonts
  \def\maketag@@@#1{\hbox{\m@th\normalsize\normalfont#1}}%
  \begin{align*}
    \lim_{k \rightarrow \infty} \mathbb{P}\left( \left \{\exists\, q'_P = (\vect{x}', \qb) \in \mathcal{V}_k^{\text{TL-RRT}^*}: \| \vect{x} - \vect{x'}\| < \delta\right \}\right) = 1,
  \end{align*}
  \endgroup
  which completes the proof.
\end{IEEEproof}

Propositions~\ref{thm:cvx} and~\ref{thm:cvx2} consider the states in $P_s$ that can be reached from the root in one hop. Next we extend Propositions~\ref{thm:cvx} and~\ref{thm:cvx2} to any reachable state in $P_s$, which is similar to Lemma 2 in~\cite{kuffner2000rrt}.
\begin{prop}
  \label{thm:ncv}
  Let Assumptions~\ref{asmp:nonr} and~\ref{asmp:nhb} hold and further assume that sampling of the free workspace is unbiased. For any state $q_P = (\vect{x}, q_B) \in P_s$ and any positive number $ \delta > 0$, we have that $\lim_{k\rightarrow\infty} \mathbb{P} (D_k (q_P) < \delta) = 1$.  Also, the probability that there exists a node in the tree that shares the same B$\ddot{\text{u}}$chi state as $q_P$ and lies within a ball of radius $\delta$ centered at $q_P$, goes to 1 as $k \to \infty$.
\end{prop}
\begin{IEEEproof}
 We prove this inductively, using Propositions~\ref{thm:cvx} and~\ref{thm:cvx2}. Specifically, let $q^0_P$ denote the root of the tree and assume there exists a feasible path between $q_P$ and the root. Then, there exists a sequence of states $\{\q{P}{0}, \q{P}{1}, \ldots, q_P^l\}$, where $q^i_P = (\vect{x}^i, q_B^i)$, $q_P^l=q_P$, and $q_P^i $ can make one-hop transition to $q_P^{i+1}$ $\forall i\in \{0,\ldots,l-1\}$. 

  By Proposition~\ref{thm:cvx2}, since $q_P^1$ is reachable from $q_P^0$ through a one-hop transition, we have that $\lim_{k \rightarrow \infty} \mathbb{P}\left(\{\exists\, (\vect{x}'_1, q_B^1) \in \mathcal{V}_k^{\text{TL-RRT}^*}:\; \| \vect{x}_1 - \vect{x}'_1\| < \delta\} \right)  = 1$ for any $\delta > 0$. In other words, as $k \to \infty$, the tree will contain a node arbitrarily close to $q_P^1$ with the same B$\ddot{\text{u}}$chi state as $\q{P}{1}$. 
  Now, consider any pair of nodes in the sequence $(\vect{x}_i,q_B^i)$ and $(\vect{x}_{i+1}, q_B^{i+1})$ such that $(\bbx_{i+1}, q_B^{i+1} )\in \ccalR_P((\bbx_{i}, q_B^i))$. Viewing $(\vect{x}'_i, q_B^i)$ as a new initial state, which is very close to $(\bbx_i, q_B^i)$, we can apply Propositions~\ref{thm:cvx} and~\ref{thm:cvx2} to get that $(\bbx_i',q_B^i)$ can reach a node $(\vect{x}'_{i+1}, q_B^{i+1})$, possibly through multiple edges in the tree, that is arbitrarily close to $(\bbx_{i+1},q_B^{i+1})$. Since $q_P = q_P^l$, we conclude that $\lim _{k\rightarrow \infty} \mathbb{P}(D_k (q_P) < \delta) = 1$, and {\footnotesize $\lim_{k \rightarrow \infty} \mathbb{P}\left( \left \{\exists\, q'_P = (\vect{x}', \qb) \in \mathcal{V}_k^{\text{TL-RRT}^*}: \| \vect{x} - \vect{x'}\| < \delta  \right \}\right) = 1$}.
\end{IEEEproof}

{Using Proposition~\ref{thm:ncv}, we can show the probabilistic completeness of TL-RRT$^*$ with unbiased sampling in Theorem~\ref{prop:compl}.}
Specifically, if feasible plans exist, then accepting states exist for both the prefix and suffix plans. For the construction of the prefix plan, the accepting states should be in the set $P_s$, satisfying the LTL specification and reachable from the root. By Proposition~\ref{thm:ncv}, the tree will grow arbitrarily close to an accepting state and it will contain a node with the same B$\ddot{\text{u}}$chi state as the accepting state. This completes the construction of the prefix plan. Moreover, after the predetermined number of iterations, the set $\ccalP$ is returned. Then, for the construction of the suffix plans, the same analysis can be applied assuming that the root becomes one certain element of $\ccalP$.  Therefore a feasible plan will be found with probability 1.

\section{Proof of Corollary~\ref{prop:biased}}\label{app:biased}

The proof is based on a similar analysis as in Lemmas~\ref{thm:step1}-~\ref{thm:step3} that are used to derive Propositions~\ref{thm:cvx} and~\ref{thm:cvx2}. As in Appendix~\ref{app:comp}, we exclude $\texttt{Extend}$ and $\texttt{Rewire}$ functions and connect $\q{P}{new}$ to $\q{P}{closest}$ directly. The key idea in Lemmas~\ref{thm:step1} to \ref{thm:step3}, is that the probability that $\x{new}$ lies within the ball $\ccalB_{d_{k}}(q_P)$ and is close enough to one candidate parent $\q{P}{cp}$ of $q_P$ is bounded away from 0 at each iteration. As a result, $\q{P}{new} = (\x{new}, \q{B}{cp})$ or $\q{P}{new} = (\x{new}, q_B)$ is the new nearest candidate parent of $q_P$. Using biased sampling, this argument still holds. Recall that the biased TL-RRT$^*$ differs from the unbiased one in the use of the biased sampling and the way to detect a cycle around the root. Since RRT$^*$ is used to detect the cycle and it is probabilistically complete, it suffices to focus on the use of the biased sampling. First, since $p_{\text{closest}} \in (0.5, 1)$, {following the distribution in~\eqref{eq:fcl},} any node has a non-zero probability to be selected as $\q{P}{closest}$. Thus, increasingly more candidate parents appear in the tree, as in the case of unbiased sampling. These candidate parents form a dense subset of $\ccalW^N$, so that the probability that any one of these candidate parents $\q{P}{cp}$ is selected as $\q{P}{closest}$ is bounded away from 0. {In the biased sampling, the selection of $\q{P}{closest}$ comes first and then the sampling of $\x{new}$ follows. Thus,  we need to ensure first that a candidate parent is selected as $\q{P}{closest}$.} Second, recall that $\ccalN_{d_k}(q_P)$ denotes the neighborhood around the path connecting $\q{P}{cp}$ and $q_P$. Since $p_{\text{rand}} \in (0.5,1)$ and the collection of $\ccalN_{d_k}(q_P)$ covers a subset of $\ccalW_N$, the probability that $\x{new}$ lies within the neighborhood region $\ccalN_{d_k}(q_P)$ is bounded away from 0, as well. Here, unlike the unbiased case, we  consider $\ccalN_{d_k}(q_P)$ rather than $\ccalI_k$, the intersection of the Voronoi cell $\ccalC(\q{P}{cp})$ and $\ccalN_{d_k}(q_P)$. This is because once $\q{P}{cp}$ is selected as $\q{P}{closest}$, those samples lying within $\ccalN_{d_k}(q_P)$ will produce a new nearest candidate parent of $q_P$, while in the unbiased case, a stricter constraint is imposed which requires the candidate parent to be the nearest node to $\x{new}$, which means $\x{new}$ should lie within $\ccalC(\q{P}{cp})$ first. For these two reasons, the same arguments as in the unbiased case can be made here too, and counterparts of Propositions~\ref{thm:cvx} and~\ref{thm:cvx2} for biased sampling can be obtained. The proof of Corollary~\ref{prop:biased} can be completed by inheriting the proofs of Proposition~\ref{thm:ncv} and Theorem~\ref{prop:compl}.

\section{Proof of Theorem~\ref{prop:opt} }\label{app:opt}
{The proof for the asymptotic optimality of TL-RRT$^*$ is based on that  for RRT$^*$~\cite{karaman2011sampling,solovey2020revisiting}, where~\cite{solovey2020revisiting} fixes a logical gap in the proof in~\cite{karaman2011sampling}.  Our proof differs  from~\cite{solovey2020revisiting} mainly in that we consider here a product state space and, thus, the definitions of transition relations differ.}

{\subsubsection{Construction of  the product path $p_{\epsilon}$}\label{sec:plans}

This step differs from~\cite{karaman2011sampling,solovey2020revisiting} in that here we build  a  product path that lives in the combined continuous and discrete state space. Specifically, let ${\tau}^*$ be the optimal continuous path that optimizes~\eqref{eq:cost2}. {Since Problem~\ref{pr:problem} is robustly feasible, following~\cite{solovey2020revisiting} there exists a robust path ${\tau}_{\epsilon}$ such that $J({\tau}_{\epsilon}) \leq (1 + \epsilon/4) J({\tau}^*)$, where $\epsilon\in(0,1)$. Next, as discussed in Section~\ref{sec:opt}, we can construct a product path $p_{\epsilon}$ from ${\tau}_{\epsilon}$.}

\subsubsection{Construction of the sequence of sets of balls $\{\mathcal{B}_n\}_{n\in\mathbb{N}}$}\label{sec:ball}
This step differs from ~\cite{karaman2011sampling,solovey2020revisiting} in that here we construct a set of balls along the product path $p_{\epsilon}$ instead of the continuous path ${\tau}_{\epsilon}$. Specifically, given the product path $p_{\epsilon}$, we define a set of $M_n$ balls $\ccalB_n = \{\mathcal{B}_{n, 1}, \ldots,\mathcal{B}_{n, M_n}\}$ that cover the whole product path $p_{\epsilon}$, each with radius $q_n  = r_n(\ccalV_\ccalT)/(2+\theta)$ and center on $p_{\epsilon}$, where $\theta\in(0, 1/4)$. The centers of two consecutive balls are $l_n = \theta q_n$ apart. The center of ball $\ccalB_{n,1}$ is the initial state of the product path, and the center of ball $\ccalB_{n,M_n}$ is the accepting state; see also Fig.~\ref{fig:balls}. The construction of the balls here is the same, in a geometric sense, as that in \cite{karaman2011sampling,solovey2020revisiting} where a set of balls is constructed along ${\tau}_{\epsilon}$ rather than $p_{\epsilon}$. Thus, we can  adapt Lemmas 53 and 54 in \cite{karaman2011sampling} to obtain that for any two points, $(\vect{x}_m,\cdot) \in \mathcal{B}_{n,m}$ and $(\vect{x}_{m+1}, \cdot) \in \mathcal{B}_{n,m+1}$ in any two consecutive balls, the following properties hold: (i) the Euclidean distance between $\vect{x}_m$ and $\vect{x}_{m+1}$ is no larger  than the radius $r_n(\ccalV_{\ccalT})$ used in Alg.~\ref{alg:extend} and~\ref{alg:rewire}; (ii) the line  $\overline{\vect{x}_m\vect{x}_{m+1}}$ lies entirely within $\ccalW_{\text{free}}^N$.

Note that the radii of the balls in the set $\mathcal{B}_n$ decrease to 0 as $r_n\to0$, and the distance between the centers  of any two consecutive balls in  $\ccalB_n$ also goes to 0. By Assumption~\ref{asmp:nhb}, for every point on the subpath connecting any two consecutive centers, $(\vect{x}_i, \qb^i) \in \mathcal{B}_{n,i}$ and $(\vect{x}_{i+1}, \qb^{i+1}) \in \mathcal{B}_{n,i+1}$, $\forall i \in [M_n-1]$, there exists a ball in which all states have the same B$\ddot{\text{u}}$chi state. Similar to the definition of the neighborhood $\ccalN_{d_k}(q_P)$ in~\eqref{eq:cylinder}, which is shown in Fig.~\ref{fig:expdrop}, we denote by $\ccalN_{n,i}$ the neighborhood along the line connecting centers $(\bbx_i, q_B^i)$ and $(\bbx_{i+1}, q_B^{i+1})$. We can conclude that there exists a large $N_0 \in \mathbb{N}$, such that $\mathcal{B}_{n,i} \subset \ccalN_{n,i}$ and $\mathcal{B}_{n,i+1} \subset \ccalN_{n,i}$ for any iteration $n>N_0$ and any two consecutive balls, because both the radii of $\mathcal{B}_{n,i}$ and $\mathcal{B}_{n,i+1}$ go to 0 and the distance between the two centers also goes to 0. Furthermore, since the balls in $\ccalB_n$ are constructed along $p_{\epsilon}$, when $n>N_0$, the center $(\bbx_i, q_B^i)$ can transition to the center  $(\bbx_{i+1}, q_B^{i+1})$, i.e. (i) $\vect{x}_i \rightarrow\vect{x}_{i+1}$, and (ii) $\qb^i \xrightarrow{L(\vect{x}_i)} \qb^{i+1}$. In what follows, we  assume $n>N_0$.
}

{Moreover, we define a tree $\mathcal{G}_n$ that contains an edge starting from  $(\vect{x}, q_B)$ and ending at  $(\vect{x}', q'_B)$ if {\it (R1)\/}: $\|\vect{x} - \vect{x}'\| \leq r_n(\ccalV_{\ccalT})$; {\it (R2)\/}: $\vect{x}$ is sampled before $\vect{x}'$; {\it (R3)\/}: $\bbx \rightarrow \bbx'$; and {\it (R4)\/}: $\qb \xrightarrow{L(\vect{x})} \qb'$. Given the same sequence of samples, the cost of the best path in $\mathcal{G}_n$ from the root to any node is not less than that generated by TL-RRT$^*$. {This is because the $\texttt{Extend}$ function in TL-RRT$^*$ will connect the new node $(\bbx',q_B')$ to the node $(\bbx, q_B)$ in the tree that incurs the least cost  as long as rules {\it (R1)-(R4)} are satisfied, and the $\texttt{Rewire}$ function will further decrease the cost of the nodes in the tree, again as long as these rules are satisfied.} Thus, if the algorithm generating $\mathcal{G}_n$ is asymptotically  optimal, so is TL-RRT$^*$.

\subsubsection{Connecting nodes in consecutive balls in $\{\mathcal{B}_n\}_{n\in\mathbb{N}}$}\label{sec:node}

To prove the optimality of $\ccalG_n$, we show that a path exists in ${\ccalG}_n$ that is arbitrarily close to $p_{\epsilon}$. To do so, we need to show that eventually every ball in $\ccalB_n$ contains at least one node of $\ccalG_n$.

Given a sequence of uniformly sampled position states, we can reason about the existence of samples inside the projections of the balls $\{\ccalB_{n}\}_{n\in \mathbb{N}}$ onto the continuous space and the order in which they were sampled. We denote by $\mathbf{X}_j$ the sample point in continuous space at the $j$-th iteration and by $V = \{\mathbf{X}_1, \ldots, \mathbf{X}_n\}$ the sample set up to the $n$-th iteration. Next, we partition the set $V$ into $M_n +1$ subsets $V_0, \ldots, V_{M_n}$ respecting the order in which the points in $V$ were  sampled. Specifically, $V_0$ contains the first $n'$ samples, where $n'=\kappa n$, $\kappa \in (0,1)$,  and the subsequent subsets $V_i$  contain $z  = \lfloor(n-n')/M_n\rfloor$ samples each, except for the last subset which contains the remaining points in $V$, that is, $V_i = \left \{\mathbf{X}_j \, |\, j \in \left \{ n' + (i-1) \cdot z, \ldots, n' + i\cdot z \right \}\right\}$, for $i\in\{1,\ldots, M_n-1\}$.
{Then, Lemma 2 in~\cite{solovey2020revisiting} proves that with probability one there is an ascending subsequence $V'$ of samples in $V$, one sample per $V_i$, that correspond one-to-one to the balls in $\{\ccalB_n\}$ in terms of order. Next we define conditions under which rules {\it (R1)-(R4)} hold so that $\ccalG_n$ can  connect the samples in $V'$, along with their B$\ddot{\text{u}}$chi states, to generate a path  that approximates $p_\epsilon$. These conditions are different than those required in~\cite{solovey2020revisiting}.}}

\begin{figure}[t]
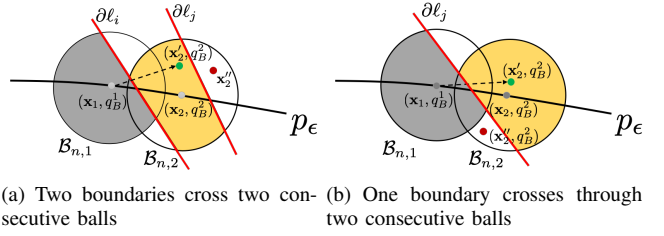

\centering
     \subfigure[Two boundaries cross two consecutive balls]{
    \label{fig:opt_2bd}
  \includegraphics[width=0.46\linewidth,  trim = {0cm 0cm 0cm 0cm}, clip]{opt_2bd.pdf}}
  \subfigure[One boundary crosses through two consecutive balls]{
    \label{fig:opt_2b}
    \includegraphics[width=0.46\linewidth]{opt_2b.pdf}}
  \caption{Graphical depiction of cases where boundaries of regions cross balls. In Fig.~\ref{fig:opt_2bd}, two different boundaries $\partial \ell_i$ and $\partial\ell_j$ cross the first two consecutive balls. In Fig.~\ref{fig:opt_2b}, a boundary crosses through two consecutive balls, the shaded parts illustrate where the centers belong.}
 \end{figure}

{Specifically, consider  the product path in its discretized form $p_{\epsilon} = (\vect{x}_1, \qb^1),(\vect{x}_2,\qb^2),\ldots, (\vect{x}_{M_n}, \qb^{M_n})$, where each point is the center of a ball constructed in~\hyperref[sec:ball]{{\it 2)}}. For simplicity, consider the first two balls. Assume the center $(\vect{x}_1, \qb^1)$ of the first ball  can take a one-hop transition to the second center $(\vect{x}_{2}, \qb^{2})$, therefore, $\vect{x}_1 \rightarrow \vect{x}_{2}$ holds. Consider also a point $\vect{x}'_{2}$ located inside $\mathcal{B}_{n,2}$ that is sampled after $\bbx_1$. Then, $\bbx_1\to \bbx_2'$ does not necessarily hold, even though $\mathcal{B}_{n,1} \subset \ccalN_{n,1}$, $\mathcal{B}_{n,2} \subset \ccalN_{n,1}$. This is because more than one boundaries $\partial \ell_j $ can cross the balls $\ccalB_{n,1}$ and $\ccalB_{n,2}$, in which case the straight line $\overline{\bbx_1\bbx'_2}$ will also cross these boundaries; see Fig.~\ref{fig:opt_2bd}. As a result, rule {\it (R3)} is not satisfied and the graph $\ccalG_n$ can not connect $\bbx_1$ to $\bbx_2'$.}
{However, as $n$ grows to infinity, the radii of the balls in the set $\ccalB_n$ go to 0, and, therefore, the probability that more than one boundaries cross the balls $\ccalB_{n,1}$ and $\ccalB_{n,2}$ goes to 0. Thus, as $n$ grows, at most one boundary $\partial\ell_j$ of a labeled region $\ell_j$ can cross one or both balls $\ccalB_{n,1}$ and $\ccalB_{n,2}$; see Fig.~\ref{fig:opt_2b}. {In this case, rule {\it (R3)} is satisfied and $\bbx_1$ can be connected to $\bbx'_2$ in $\ccalB_{n,2}$. In what follows, we show that the exact location of $\bbx_2'$ in $\ccalB_{n,2}$ is important in order to further satisfy rule {\it (R4)}.} Specifically, when $n$ is large, this boundary can be locally approximated by a hyperplane, which divides $\mathcal{B}_{n,2}$ into two parts. Suppose $\bbx''_2$ and $\bbx_2$ are in different parts of the ball $\mathcal{B}_{n,2}$. Then, the B$\ddot{\text{u}}$chi state $q_B^2$ can be paired with $\bbx''_2$, because in Alg.~\ref{alg:tree}, the B$\ddot{\text{u}}$chi state space is searched exhaustively. But since  $L(\bbx_2) \neq L(\bbx''_2)$, it is possible that $q_B^2$ can not transition at all or can transition to a state other than $q_B^3$, i.e., the B$\ddot{\text{u}}$chi state of the center of $\ccalB_{n,3}$. In this case,   the plan under construction will not proceed along $p_{\epsilon}$. To avoid this situation, $\vect{x}'_{2}$ should lie within that part of $\ccalB_{n,2}$ that contains its center, as opposed to~\cite{solovey2020revisiting} where to follow  ${\tau}_{\epsilon}$, it suffices that $\vect{x}'_2$ lies anywhere within $\ccalB_{n,2}$. {Following this logic, to satisfy rule {\it (R4)}, the samples in  $V'$ should lie in those parts of the balls that contain their centers. Let $E_n^1$ be the event that the projection of that part of ball $\ccalB_{n,i}$ that contains its center on the continuous space contains at least one sample in $V_i$,  i.e., $E^1_n  = \{\forall 1 \leq i \leq M_n, \ccalB'_{n,i}\, \cap \, V_i \not= \varnothing \}$, where  $\ccalB_{n,i}'$ is the part of the ball that contains its center so that $\ccalB'_{n,i} = \ccalB_{n,i}$ if $\ccalB_{n,i}$ is not crossed by a boundary, and $\ccalB'_{n,i} \subset \ccalB_{n,i}$ otherwise, and the intersection $\ccalB'_{n,i} \, \cap \,V_i$ is defined on the projection of $\ccalB'_{n,i}$ on the continuous space. Assuming that the event $E^1_n$ occurs, next we show that if  rules {\it (R1)-(R4)}  are satisfied, then indeed $\ccalG_n$ can connect samples in the subsequence $V'$, augmented by B$\ddot{\text{u}}$chi states, to get a path that approximates $p_{\epsilon}$.}}

 {Let $n$ be large enough so that at most one  boundary crosses the balls in the set $\{\ccalB_n\}$, and consider the node $(\bbx'_3, q_B^3) \in \ccalB'_{n,3}$ with $\bbx'_3$ drawn after $\bbx'_2$.  Then, since  $\bbx'_2$ and $\bbx'_3$ are located in two consecutive balls and we have proved in step~\hyperref[sec:ball]{{\it 2)}} that the distance between any states in any two consecutive balls is no more than $r_n(\ccalV_{\ccalT})$, {\it (R1)} is satisfied. Furthermore, since $\bbx'_2$ is sampled before $\bbx'_{3}$, {\it (R2)} holds. Recall from step~\hyperref[sec:ball]{{\it 2)}} that the straight line connecting any two states in any two consecutive balls is entirely inside $\ccalW_{\text{free}}^N$, and that those balls are shrinking, so eventually there is at most one boundary crossing through two consecutive balls. Therefore, rule {\it (R3)} is also satisfied. {As for rule {\it (R4)}, note that  $q_B^2 \xrightarrow{L(\bbx'_2)} q_B^3$ since $\bbx'_2$ belongs to the same part of the ball that contains its center and the center $(\bbx_2, q_B^2)$ can transition to the center $(\bbx_3, q_B^3)$, that is, $q_B^2 \xrightarrow{L(\bbx_2)} q_B^3$. Therefore, {\it (R4)} is also satisfied and $(\bbx_3',q_B^3)\in \ccalB_{n,3}'$ will be added to the tree $\ccalG_n$. Iterating over the sequence of balls, we conclude that the tree $\ccalG_n$ will have nodes in each one of the parts $\ccalB'_{n,i}$ that contain their centers. Since every $\ccalB_{n,i}'$ contains one point in $V'$, we finally conclude that $\ccalG_n$ can connect the samples in the subsequence $V'$.}

   In what follows, we prove that the probability of event $E_n^1$ approaches 1. First, let $E_n^2$ denote the event that $\x{new}=\x{rand}$ at the $n$-th iteration. Letting $n\to \infty$ and using the fact  that $\lim_{n\to\infty} \mathbb{P}(E_n^2) = 1 $ shown in~\cite{solovey2020revisiting}, we have that }
   \begingroup\makeatletter\def\f@size{8}\check@mathfonts
\def\maketag@@@#1{\hbox{\m@th\normalsize\normalfont#1}}%
\allowdisplaybreaks
 \begin{align}
  \mathbb{P}\left(E_n^1\right) & = \mathbb{P}\big(E_n^1 | E_n^2\big) \mathbb{P}\left(E_n^2\right) +  \mathbb{P} (E_n^1 | \overline{E_n^2}) \mathbb{P}\big(\overline{E_n^2}\big) \nonumber \\
  &  \geq \mathbb{P} \big(E_n^1 | E_n^2 \big) \mathbb{P}\big(E_n^2\big) \nonumber  \\
    & = \big ( 1 -  \mathbb{P}\big(\overline{E_n^1} \big| E_n^2\big) \big) \big( 1 - \mathbb{P}\big(\overline{{E}_n^2}\big) \big) \nonumber \\
  & = 1 - \mathbb{P}\big(\overline{{E}_n^2}\big) -  \mathbb{P}\big(\overline{E_n^1}\big| {{E}}_n^2\big)  +  \mathbb{P}\big(\overline{E_n^1}\big| {{E}}_n^2\big)  \mathbb{P}\big(\overline{{{E}}_n^2}\big) \nonumber \\
 & =  1 -  \mathbb{P}\big(\overline{E_n^1}\big| {{E}}_n^2\big), \label{eq:event2}
 \end{align}
 \endgroup
{where $\overline{E_n^2}$ is the complement of the event $E_n^2$. Next, let $c_n$ denote the number of balls that are crossed by a boundary. Then,  we have that}
\begingroup\makeatletter\def\f@size{8}\check@mathfonts
\def\maketag@@@#1{\hbox{\m@th\normalsize\normalfont#1}}%
\allowdisplaybreaks
\begin{align}
  & \mathbb{P}\big (\overline{E_n^1}\big| {E}_n^2 \big)  = \mathbb{P}(\exists 1 \leq i \leq M_n, \ccalB'_{n,i} \, \cap \, V_i = \varnothing ) \label{eq:not2}\\
   & \leq \sum\nolimits_{i=1}^{M_n} \mathbb{P}(\ccalB'_{n,i}  \, \cap \, V_i = \varnothing) = \sum\nolimits_{i=1}^{M_n} \Big(1 - \frac{\mu(\ccalB'_{n,i})}{\ccalW_{\text{free}}^N}\Big)^{|V_i|} \label{eq:uniform} \\
  & \leq (M_n- c_n)  \Big(1 - \frac{\mu(\ccalB_{n,i})}{\ccalW_{\text{free}}^N}\Big)^{|V_i|} + c_n  \Big(1 - \frac{\mu(\ccalB_{n,i})}{2\ccalW_{\text{free}}^N}\Big)^{|V_i|}
  \label{eq:split}\\
  & \leq M_n  \Big(1 - \frac{\mu(\ccalB_{n,i})}{\ccalW_{\text{free}}^N}\Big)^{|V_i|} + c_n  \Big(1 - \frac{\mu(\ccalB_{n,i})}{2\ccalW_{\text{free}}^N} \Big)^{|V_i|}, \label{eq:mncn}
\end{align}
\endgroup
{where $|V_i|=(n-n')/M_n$,~\eqref{eq:uniform} is obtained assuming that $\x{new}$ is sampled uniformly which holds for large $n$ since $\x{new} = \x{rand}$, and to derive~\eqref{eq:split}, {we divide all balls into two classes based on whether they are crossed by a boundary or not. For those balls crossed by a boundary, we have $\mu(\ccalB'_{n,i}) \geq \mu(\ccalB_{n,i})/2$; otherwise, $\mu(\ccalB'_{n,i}) = \mu(\ccalB_{n,i})$.} According to~\cite{solovey2020revisiting}, the first term in~\eqref{eq:mncn} approaches 0 when $\gamma_{\text{TL-RRT}^*}\geq (2 + \theta) ( \frac{(1+\epsilon/4)J(\tau^*)}{(\texttt{dim}+1)\theta (1-\kappa)} \cdot   \frac{\mu(\ccalW_{\text{free}}^N)}{\zeta_{\texttt{dim}}})^{\frac{1}{\texttt{dim}+1}}$, and also $ (1 - \frac{\mu(\ccalB_{n,i})}{\ccalW_{\text{free}}^N})^{|V_i|} \leq \exp\{-\xi \gamma_{\text{TL-RRT}^*}^{\texttt{dim}+1}\log{n}\}$, where $\xi:= \frac{\theta \zeta_{\text{dim}} (1-\kappa)}{J({\tau}^{\epsilon})(2+\theta)^{\text{dim}+1} \mu(\ccalW_{\text{free}}^{N})}$. {Following the same logic, we can show that the fraction $\frac{1}{2}$ in the second term in~\eqref{eq:mncn} {can be absorbed into } $\xi$} and $ (1 - \frac{\mu(\ccalB_{n,i})}{2\ccalW_{\text{free}}^N})^{|V_i|} \leq \exp\{-\frac{1}{2}\xi \gamma_{\text{TL-RRT}^*}^{\texttt{dim}+1}\log{n}\}$. Since $c_n$ is a small number (because there are only a finite number of boundaries), we have that the second term in~\eqref{eq:mncn}  approaches 0. Thus, $\mathbb{P} (\overline{E_n^1}| {E}_n^2  ) $ approaches 0. Substituting in~\eqref{eq:event2} we have that $\mathbb{P}(E_n^1)$  converges to 1 as $n\to \infty$.  This guarantees that $\ccalG_n$ will contain a path that approximates $p_{\epsilon}$.

  The rest of the proof is similar to that in~\cite{solovey2020revisiting}. {Essentially, we show that the probability of the event $E_{n}^3$ that at most $\alpha M_n$ of the balls $\ccalB_{n,i}^{\beta}$ do not contain any samples from $V_i$ approaches 1, where $\alpha \in (0, \theta\epsilon/16), \beta \in (0, \theta \epsilon/16)$, and $\ccalB_{n,i}^{\beta}$ shares the same center with $\ccalB_{n,i}$ but its radius is a fraction $\beta$ of that of $\ccalB_{n,i}$. In this way,  we exclude the case where the tree $\ccalG_n$ has a path approximating   $p_\epsilon$ but zig-zaging around it. We omit the details due to space limitations. {Recall from step~\hyperref[sec:plans]{\it 1)} that the cost of the path $\tau_\epsilon$ is smaller than $ (1 + \epsilon/4) J({\tau}^*) $.}  Then, when events $E_n^1$ and $E_n^3$ occur, {as shown in~\cite{solovey2020revisiting},} the cost of the path returned by $\ccalG_n$ is less than $(1+\epsilon) J({\tau}^*)$, which is slightly larger than $ (1 + \epsilon/4) J({\tau}^*) $, since this path is an approximation of $p_{\epsilon}$.}
}

\section{Proof of Theorem~\ref{prop:opt1}}\label{app:biasedopt}
{The key step to obtain $\gamma_{\text{TL-RRT}^*}^{\text{unbiased}}$ for uniform sampling lies in upper-bounding the probability {$\mathbb{P} \big(\overline{E_n^1}\big| {{E}}_n^2 \big)$} in~\eqref{eq:not2} by a quantity that converges to 0. Although the biased samples are guided towards given labeled regions, according to~\eqref{eq:fnew}, with probability $(1-y_{\text{rand}})^N$ the sample $\x{rand}$ still follows the uniform distribution. Thus, within the whole sequence of $n$ biased samples, there is a subsequence that contains $(1-y_{\text{rand}})^N \cdot n$ samples on average that were sampled uniformly. We focus only on this subsequence of uniform samples and proceed in the same way as in the proof for the unbiased sampling. Specifically, for this  subsequence, $|V_i|$ in~\eqref{eq:mncn} will become $(1-y_{\text{rand}})^N (n-n')/M_n$. Moreover, as before,  the radius of the balls $\{\ccalB_{n}\}$ in Fig.~\ref{fig:balls} will be proportional to the connection radius $r_n(\ccalV_\ccalT)$ that depends on the length $n$ of the whole sequence. {Therefore, in~\eqref{eq:mncn},   $|V_i|$ will change, and so will $\mu(\ccalB_{n,i})$ and $M_n$ since they are related to $\gamma^{\text{biased}}_{\text{TL-RRT}^*}$.} Following a similar derivation as in~\cite{solovey2020revisiting} that shows that the first term in~\eqref{eq:mncn} approaches 0,  we can show that the extra term $(1-y_{\text{rand}})^N$ in $|V_i|$ will enter the denominator of $\gamma^{\text{unbiased}}_{\text{TL-RRT}^*}$, and the upper bound on $\mathbb{P} \big(\overline{E_n^1}\big| {{E}}_n^2 \big)$ will still converge  to 0. Note that the upper bound on $\mathbb{P} \big(\overline{E_n^1}\big| {{E}}_n^2 \big)$ obtained using the whole sequence of biased samples will be  smaller than that obtained using  the subsequence of uniform samples. This is because the samples that are biased towards accepting states can also lie in the balls $\{\ccalB_{n}\}$, {further reducing the chance that a ball $\ccalB_{n,i}$ does not contain a sample in $V_i$.
    Since this bound will be smaller, the probability of $E_n^1$ will converge faster to 1, which explains why biased TL-RRT$^*$ is faster than the unbiased version.}}
%




\end{document}